\documentclass[acmsmall]{acmart}
\AtBeginDocument{%
  }

\setcopyright{acmlicensed}
\acmJournal{PACMMOD}
\acmYear{2025} \acmVolume{3} \acmNumber{1 (SIGMOD)} \acmArticle{53} \acmMonth{2}\acmDOI{10.1145/3709703}
\received{July 2024}
\received[revised]{September 2024}
\received[accepted]{November 2024}

\acmISBN{978-1-4503-XXXX-X/2018/06}

\newcommand{\name}{\textsc{Memo}\xspace}
\newcommand{\add}[1]{{#1}}

\usepackage{subfigure}
\usepackage{xspace}
\usepackage{soul}
\usepackage{xcolor}
\usepackage{multirow}
\usepackage{multicol}
\usepackage{array}
\usepackage{booktabs}
\usepackage{makecell}
\usepackage{enumitem}
\usepackage{bbding}
\usepackage{colortbl}
\usepackage{hhline}
\begin{document}

\title{\name: Fine-grained Tensor Management For Ultra-long Context LLM Training}


\author{Pinxue Zhao}
\authornote{Pinxue Zhao, Hailin Zhang, Fangcheng Fu, Xiaonen Nie and Bin Cui are with the School of Computer Science \& Key Lab of High Confidence Software Technologies (MOE), Peking University. Bin Cui is also with the Institute of Computational Social Science, Peking University (Qingdao).}
\affiliation{%
 \institution{Peking University}
 \country{China}
}
\email{pinxue.zhao@pku.edu.cn}

\author{Hailin Zhang}
\authornotemark[1]
\affiliation{%
 \institution{Peking University}
 \country{China}
}
\email{z.hl@pku.edu.cn}

\author{Fangcheng Fu}
\authornotemark[1]
\affiliation{%
\institution{Peking University}
\country{China}}
\email{ccchengff@pku.edu.cn}

\author{Xiaonan Nie}
\authornotemark[1]
\affiliation{%
 \institution{Peking University}
 \country{China}}
\email{xiaonan.nie@pku.edu.cn}

\author{Qibin Liu}
\affiliation{%
 \institution{Tencent Inc.}
 \country{China}}
\email{brendenliu@tencent.com}

\author{Fang Yang}
\affiliation{%
 \institution{Tencent Inc.}
 \country{China}}
\email{youngfyang@tencent.com}

\author{Yuanbo Peng}
\affiliation{%
 \institution{Tencent Inc.}
 \country{China}}
\email{yuanbopeng@tencent.com}

\author{Dian Jiao}
\affiliation{%
 \institution{Tencent Inc.}
 \country{China}}
\email{focusjiao@tencent.com}

\author{Shuaipeng Li}
\affiliation{%
 \institution{Tencent Inc.}
 \country{China}}
\email{shuaipengli@tencent.com}

\author{Jinbao Xue}
\affiliation{%
 \institution{Tencent Inc.}
 \country{China}}
\email{jinbaoxue@tencent.com}

\author{Yangyu Tao}
\affiliation{%
 \institution{Tencent Inc.}
 \country{China}}
\email{brucetao@tencent.com}

\author{Bin Cui}
\authornotemark[1]
\affiliation{%
 \institution{Peking University}
 \country{China}}
\email{bin.cui@pku.edu.cn}
\renewcommand{\shortauthors}{Pinxue Zhao et al.}

\begin{abstract}
Nowadays, Large Language Models (LLMs) have been trained using extended context lengths to foster more creative applications.
However, long context training poses great challenges considering the constraint of GPU memory.
It not only leads to substantial activation memory consumption during training, but also incurs considerable memory fragmentation.
To facilitate long context training, existing frameworks have adopted strategies such as recomputation and various forms of parallelisms.
Nevertheless, these techniques rely on redundant computation or extensive communication, resulting in low Model FLOPS Utilization (MFU).
In this paper, we propose \name, a novel LLM training framework designed for fine-grained activation memory management.
Given the quadratic scaling of computation and linear scaling of memory with sequence lengths when using FlashAttention, we offload memory-consuming activations to CPU memory after each layer’s forward pass and fetch them during the backward pass.
To maximize the swapping of activations without hindering computation, and to avoid exhausting limited CPU memory, we implement a token-wise activation recomputation and swapping mechanism.
Furthermore, we tackle the memory fragmentation issue by employing a bi-level Mixed Integer Programming (MIP) approach, optimizing memory reuse across transformer layers.
Empirical results demonstrate that \name achieves an average of 
$1.97\times$
and 
$1.80\times$
MFU compared to Megatron-LM and DeepSpeed, respectively.
This improvement is attributed to \name's ability to minimize memory fragmentation, reduce recomputation and intensive communication, and circumvent the delays associated with the memory reorganization process due to fragmentation. 
By leveraging fine-grained activation memory management, \name facilitates efficient training of 7B LLM with 1 million sequence length on just 8 A800 GPUs, achieving an MFU of 52.30\%. 
\end{abstract}

\ccsdesc[500]{Computing methodologies~Distributed computing methodologies}

\keywords{Large Language Model, Long Context Training, Tensor Management, Data Rematerialization}



\maketitle

\section{Introduction}
\label{sec:intro}

Since the advent of ChatGPT~\cite{DBLP:journals/corr/abs-2303-08774}, Large Language Models (LLMs) have demonstrated remarkable proficiency in comprehending and generating natural language texts.
Besides revolutionizing the field of language processing, which encompasses translation~\cite{DBLP:journals/corr/abs-2304-04675}, coding~\cite{DBLP:journals/corr/abs-2308-12950,DBLP:journals/corr/abs-2401-14196, scis2}, etc., transformer-based LLMs have also found applications in multi-modal scenarios, such as image processing~\cite{DBLP:conf/iclr/DosovitskiyB0WZ21,DBLP:conf/iccv/PeeblesX23}, video stream analysis~\cite{DBLP:journals/aiopen/RuanJ22}, and AI for science~\cite{DBLP:journals/nature/BiXZCG023,DBLP:journals/corr/abs-2311-07361}.
To accommodate novel applications that require lengthy contexts~\cite{rag_survey}, LLMs have developed to support long context input, from 2K-4K~\cite{DBLP:journals/corr/abs-2307-09288,taori2023alpaca} to 32K~\cite{DBLP:journals/corr/abs-2310-06825,llama32k}, 128K~\cite{DBLP:journals/corr/abs-2303-08774,DBLP:journals/corr/abs-2402-10171}, or even millions of tokens~\cite{kimichat,tongyiqianwen,DBLP:journals/corr/abs-2402-08268}.
Considering the extrapolation problem~\cite{DBLP:conf/iclr/PressSL22,DBLP:journals/corr/abs-2310-05209}, which refers to the decline in LLM performance when input sequences exceed the training length, it is necessary to conduct long context training~\cite{DBLP:journals/corr/abs-2309-14509,DBLP:journals/corr/abs-2401-09149,DBLP:journals/corr/abs-2405-07719} or fine-tuning~\cite{DBLP:journals/corr/abs-2309-00071,DBLP:journals/corr/abs-2402-13753} to facilitate long sequence inference.
Beyond natural language processing, increasing the context length is also essential across diverse domains, including video processing~\cite{opensora}, protein properties prediction~\cite{chandra2023transformer}, weather forecasting~\cite{DBLP:conf/icml/NguyenBKGG23}, and health care~\cite{DBLP:journals/corr/abs-2201-11838}.

\add{Maximizing system performance with limited memory is a common and significant challenge in the data management community.
Within this context, training LLMs with long sequence lengths poses difficulties due to restricted GPU memory.
}
During training, a large amount of activations\footnote{\add{In neural network training, the outputs of operators are referred to as activations. Some of these, termed ``skeletal activations'' in this work, must be stored for gradient computation during the backward pass, while others are termed ``transient activations''.  More details are provided in Section~\ref{sec:anatomy}.}}
must be stored for gradient computation during the backward pass, resulting in substantial memory consumption. 
Typically, it is well known that the self-attention module in the transformer architecture has a quadratic computation and memory complexity w.r.t. the sequence length. 
%
FlashAttention~\cite{flash1, flash2}, now a standard technique for attention computation in LLM training, accelerates computation and shrinks the memory complexity to be linear w.r.t. the sequence length by scheduling memory I/O and recomputing necessary components during the backward pass.
Except for attention, the remaining activation memory also scales linearly with the sequence length, which can become quite large in long context scenarios.
For instance, training a GPT model with 7B parameters on a sequence length of 1 million can lead to an activation memory of 4096GB, far exceeding the memory capacity of commonly used accelerators (e.g. 80GB for an NVIDIA H100/A100 GPU).

\begin{figure}[!t]
    \centering
    \subfigure{
    \scalebox{0.4}{
    \includegraphics[width=\linewidth]{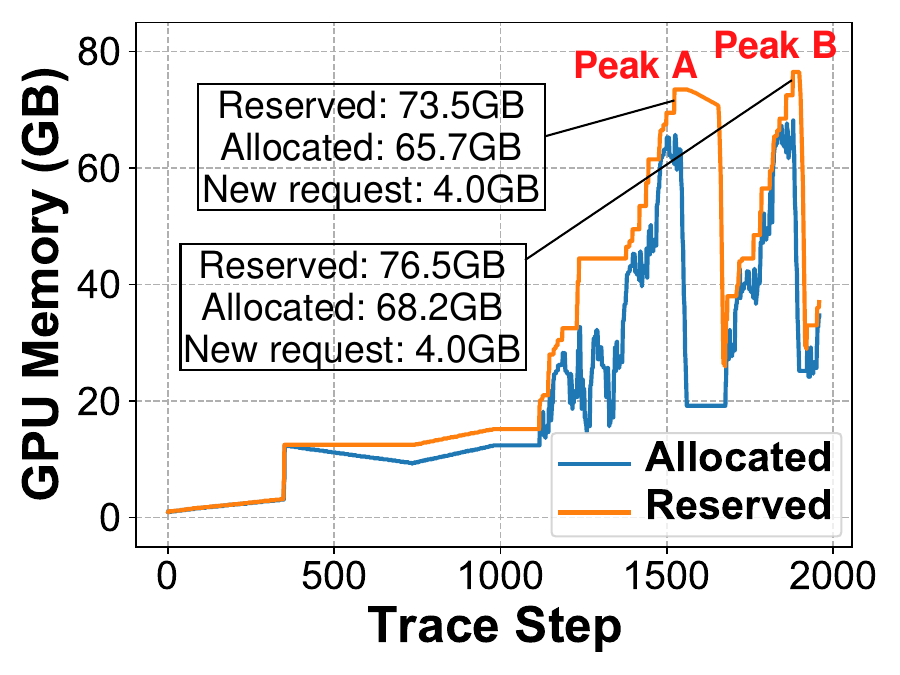}
    }\label{fig:fragments}
    }
    \subfigure{
    \scalebox{0.4}{
    \includegraphics[width=\linewidth]{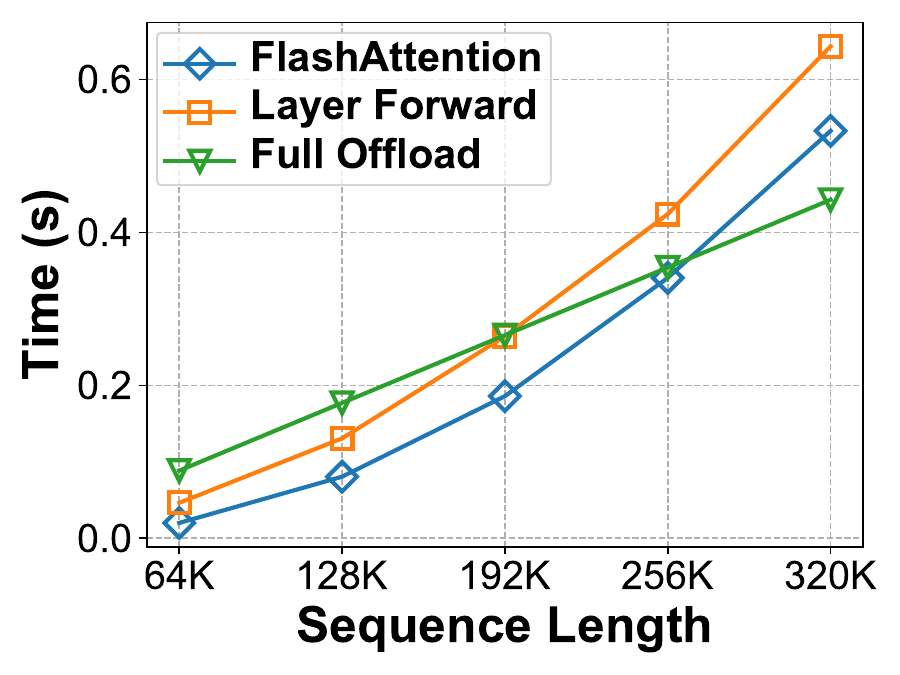}
    }\label{fig:intro:time}
    }
    \caption{The left figure, generated using PyTorch's snapshot API~\cite{torch_mem_utils}, shows the allocated and reserved GPU memory of PyTorch when training a 7B GPT model with sequence length 512K. 
    The right figure shows the time consumption of FlashAttention computation, one transformer layer forward computation, and one-layer full activation offloading when training a 7B GPT on 8 A800 GPUs with a TP size of 8.}
\end{figure}

\add{Moreover, dynamic memory allocators inherently face the issue of memory fragmentation due to frequent allocation and deallocation, which complicates efficient data management.}
%
%
Besides storing the \textit{skeletal activations} for the backward pass, there are also tremendous \textit{transient activations} that are temporarily generated during computation (we will formally categorize the two kinds of activations in Section~\ref{sec:anatomy}). Such transient activations have distinct data life cycles and usually lead to frequent allocation and deallocation of GPU memory. 
Currently, most LLM training systems are built on top of PyTorch~\cite{DBLP:conf/nips/PaszkeGMLBCKLGA19}, including Megatron-LM~\cite{DBLP:journals/corr/abs-1909-08053} and DeepSpeed~\cite{DBLP:conf/kdd/RasleyRRH20}.
%
PyTorch employs a caching memory allocator designed to reduce the costly ``cudaMalloc'' and ``cudaFree'' operations by caching and reusing allocated memory blocks. 
However, the frequent memory (de)allocation requests in the caching allocator result in significant memory fragmentation~\cite{DBLP:conf/asplos/0003ZXLLHGWZZZ24}.
This issue becomes more severe in long context training, considering the fact that the (de)allocated memory blocks are significantly larger than those in normal tasks.
Memory fragmentation not only leads to Out-Of-Memory (OOM) error but also significantly hinders training efficiency because of the frequent triggering of the PyTorch memory reorganization process, which involves calls to ``cudaFree'' and ``cudaMalloc'' to release cached blocks and reclaim GPU memory.
Figure~\ref{fig:fragments} illustrates an example of GPU memory fragmentation. 
At the peaks of the curves, there is more than 4GB memory reserved but not allocated.
However, when the training task tries to allocates 4GB memory, the allocator fails to find a continuous memory space to fulfill the allocation request. 
Consequently, it necessitates invoking a series of ``cudaFree'' and ``cudaMalloc'' to reorganize memory, which blocks GPU computation.
%

In this paper, we aim to tackle the memory challenges encountered during long context LLM training.
Specifically, we propose and implement an LLM training framework \name  to address the activation data management problem.
There are several key observations that inspire our design.

\add{
In the data management domain, when high-bandwidth memory is limited, rematerialization is a typical technique to free up memory by releasing data that is not immediately needed and reconstructing data structures on demand.
For instance, Spark~\cite{spark} supports rematerializing Resilient Distributed Datasets (RDDs)~\cite{rdd} through recomputation or swapping\footnote{Swapping refers to moving data between different levels of memory hierarchy. More details are provided in Section~\ref{sec:pre_memory_reduction}.} between CPU memory and disk storage when RAM is insufficient.
Faiss~\cite{faiss}, a widely used vector database, utilizes a CPU-GPU memory hierarchy to accommodate the high memory requirement. 
Additionally, the dynamic view materialization solution~\cite{dynamic} employs an LRU cache to manage materialized views, and rematerializes cache-missed views upon needed.
These methods leverage recomputation and swapping strategies to efficiently manage data structures within the memory hierarchy.
}
%
Similarly, in deep learning training, to reduce the peak memory consumption caused by skeletal activations, activation recomputation ~\cite{DBLP:journals/corr/ChenXZG16,DBLP:conf/iclr/KirisameLHBHRCT21,DBLP:journals/corr/abs-2205-05198} and swapping~\cite{vdnn,transformerengine} are also widely adopted.
\footnote{Parallelism techniques like sequence parallelism~\cite{DBLP:journals/corr/abs-2205-05198,DBLP:journals/corr/abs-2309-14509} and context parallelism~\cite{DBLP:journals/corr/abs-2310-01889,DBLP:journals/corr/abs-2310-03294,contextparallel} are also compelling approaches to reduce memory at the price of extra communication overhead. Our work is compatible with these parallelism techniques.} 
Typically, both of them reduce memory consumption at the price of extra time cost. 
%
The activation recomputation technique discards some activations in the forward pass and later recomputes them in the backward pass, leading to extra computation cost. 
%
The swapping technique offloads the activations to CPU memory in the forward pass to relieve the GPU memory pressure, and later fetches them back to GPU memory in the backward pass, incurring the overhead of data transmission between CPU and GPU memory.

\textbf{\textit{Observation 1: Opportunity for activation swapping.}}
\label{intro:observ:1}
Contemporary mainstream LLM training frameworks such as Megatron-LM and DeepSpeed prefer activation recomputation to swapping, which is due to the fact that the GPU computing ability has a far more rapid growth than the connectivity between CPU and GPU memory in the past few years (see Section~\ref{sec:pre_memory_reduction} for details). 
%
However, we find that the situation is a bit different in long context training of LLMs. 
Denote $s$ as the sequence length. 
The computation complexity of one transformer layer is $O(s^2)$, while the activation memory complexity is $O(s)$ thanks to FlashAttention. 
During GPU computation, we can leverage the idle CPU-GPU bandwidth, offloading activations to CPU memory during the forward pass, and fetching the activations during the backward pass. 
As the sequence length increases, there is greater potential for overlapping computation and communication, given that their time requirements scale quadratically and linearly with the sequence length, respectively.
As shown in Figure~\ref{fig:intro:time}, eventually, after reaching a specific sequence length (192K in this case), the transmission of  activations can be fully overlapped with GPU computation.

%

\begin{figure}[!t]
  \centering
  \includegraphics[width=0.8\linewidth]{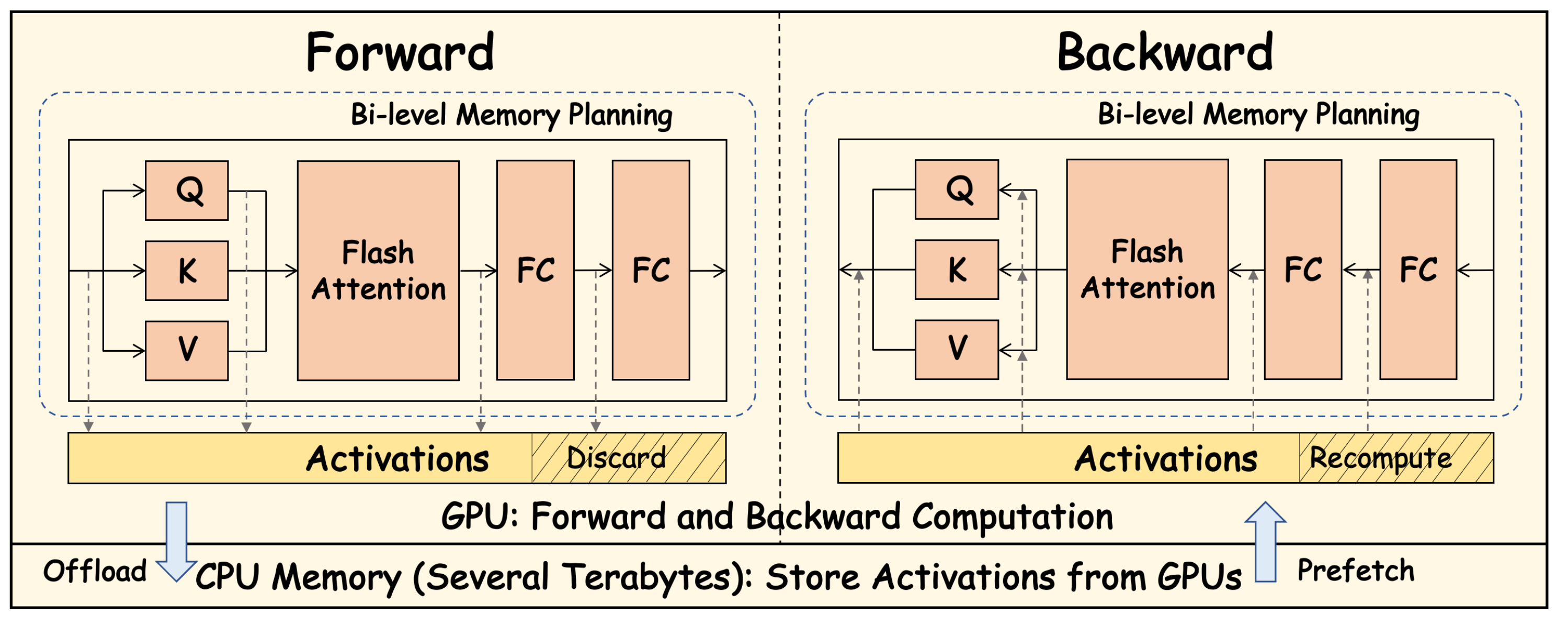}
  \caption{An overview of \name. We devise a fine-grained recomputation and swapping mechanism to manage the skeletal activations for backward propagation, and leverage a bi-level memory planning method to reuse the memory space of transient activations across the transformer layers.}
    \Description{Memo Overview}
  \label{fig:method:memo_overview}
\end{figure}

However, in practice, there is limited chance to swap all activations.
On the one hand, extremely long training data is rare, and most of the time we need to train on data that doesn't fully overlap the activation transmission and the computation.
On the other hand, offloading all activations may cause CPU OOM issues --- the CPU memory is responsible for storing all activations from all GPUs on the same machine, but the current CPU memory is typically several terabytes, which is insufficient for very long sequence lengths. 
%
Considering the above challenges, we introduce \ul{\textbf{a fine-grained activation recomputation and swapping mechanism}} to manage the skeletal activations. 
We consider both tensor-level and token-level activation management.
For each layer, following previous works~\cite{transformerengine,DBLP:journals/corr/abs-2310-03294}, we consistently offload two activation tensors, the input of each transformer layer and the output of FlashAttention, to CPU memory.
For other activation tensors, we only offload a fraction (denoted as $\alpha$) of tokens, and recompute the rest part during the backward pass.
We model the time cost of activation recomputation and transmission and determine the fraction $\alpha$ through a well-formulated linear programming problem, which aims to maximize offloading activations without impeding GPU computation or causing CPU OOM issues. 
%
During the backward pass, prefetching activations can also overlap with GPU computation, because the backward computation is typically twice as much as the forward computation.
With both tensor-level and token-level activation management, we make full use of the idle bandwidth and minimize the recomputation overhead to improve the overall efficiency.

\add{Regarding memory fragmentation, research has utilized the characteristics of targeted workloads to analyze and resolve the issue, such as experimentally analyzing the impact of memory allocation on high-performance query engines~\cite{defragment1} and addressing persistent memory fragmentation with efficient defragmentation algorithms~\cite{defragment2}. 
In the same vein, for long-sequence LLM training, we aim to leverage the specific characteristics of LLM training to address memory fragmentation.
}

\textbf{\textit{Observation 2: Deterministic memory (de)allocation pattern across iterations and layers.}}
%
In long sequence LLM training, the memory fragmentation mainly comes from frequent and irregular memory (de)allocation requests.
However, we observe that, typical LLM training adheres to a deterministic computation process across iterations and layers.
All transformer layers in an LLM are identical, and each training iteration involves the same computation.
While the general-purpose caching allocator is designed for dynamic computation routines, training LLMs can be conceptualized as static computation graphs~\cite{DBLP:conf/asplos/AnselYHGJVBBBBC24}, which have identical structures across layers.
This provides an opportunity to design static planning for each layer and reuse the allocated memory of each layer, thereby mitigating memory fragmentation.

To enhance memory utilization while minimizing fragmentation, we leverage a \ul{\textbf{hierarchical Mixed Integer Programming (MIP)}} technique to tackle the memory planning problem.
Before training, we profile the memory (de)allocation requests of one training iteration, then use MIP to solve an optimized memory plan for a single transformer layer. 
Since the memory requests of the transformer layers are identical,  the entire memory block for one layer can be directly reused for the subsequent identical layer.
Considering each transformer layer's memory block as a single memory allocation request, we further solve another MIP problem that plans memory allocation for the entire LLM training, including the initial embedding layer, all transformer layers, and the final classifier layer.
We only need to solve the problem once before the actual training, since all iterations can utilize the same memory plan.
The near-optimal memory plan eliminates the fragmentation issue and avoids PyTorch's time-consuming memory reorganization mechanism.

Putting them together, in response to the activation memory challenge in long context training, we propose \ul{\textbf{\name}}, an LLM training framework with fine-grained tensor memory management.
%
\add{We consider the challenge as an activation data management problem, and draw inspiration from the data rematerialization and memory defragmentation techniques to address the challenge.
Figure~\ref{fig:method:memo_overview} presents an overview of \name.}
To make full use of the idle CPU-GPU bandwidth during training with different sequence lengths, we introduce a token-wise fine-grained activation recomputation and swapping strategy.
We employ a bi-level hierarchical MIP technique to solve the memory planning problem and eliminate memory fragmentation.
To the best of our knowledge, this is the first training framework that enables efficient training of a 7B LLM on 8 GPUs with a sequence length of 1 million.

\begin{table}[!t]
 \footnotesize
    \caption{Commonly used notations in this work.} \label{tab:notation}
    \begin{tabular}{c|l}
    \toprule[1pt]
    \textbf{Notation} & \textbf{Explanation} \\
    \midrule[0.8pt]
    $b$ & Batch size \\
    $s$ & Context length \\
    $n$ & Number of transformer layers \\
    $h$ & Hidden size \\
    $P$ & Number of model parameters \\
    $\alpha$ & The fraction of swapping \\
    \bottomrule[1pt]
    \end{tabular}
\end{table}

We summarize our contributions as follows:
\begin{itemize}[leftmargin=*,parsep=0pt,itemsep=0pt,topsep=2pt,partopsep=2pt]
    \item We propose and implement an LLM training framework \name to address the activation data management problem in long context LLM training.
    \item We introduce a fine-grained activation recomputation and swapping mechanism to fully utilize the idle CPU-GPU communication bandwidth during time-consuming GPU computation. 
    \item We employ a bi-level MIP technique to solve the memory planning problem and significantly mitigate memory fragmentation.
    \item We evaluate \name through extensive experiments, and demonstrate an average of \add{$1.97\times$} and \add{$1.80\times$} \add{improvement in terms of MFU\footnote{\add{MFU (Model FLOPs Utilization) is a widely used efficiency metric that evaluates how well the accelerators are utilized in model training. It is calculated as the ratio of the observed throughput to the theoretical throughput which assumes the hardware operates at peak FLoating-point Operations Per Second (FLOPS)~\cite{mfu} . More details are provided in Section~\ref{exp:metric}.}}} compared to Megatron-LM and DeepSpeed, respectively. Additionally, \name is the first framework that enables the efficient training of 7B LLM with 1 million context length on only 8 A800 GPUs.
   
\end{itemize}


\section{Preliminary}

In this section, we present an overview of the architecture and training process of LLMs, along with memory reduction strategies and distributed training techniques.
Commonly used notations are listed in Table~\ref{tab:notation}.



\begin{figure}[tb]
    \centering
    \subfigure[The architecture of a typical LLM.]{
    \scalebox{0.47}{
    \includegraphics[width=\linewidth]{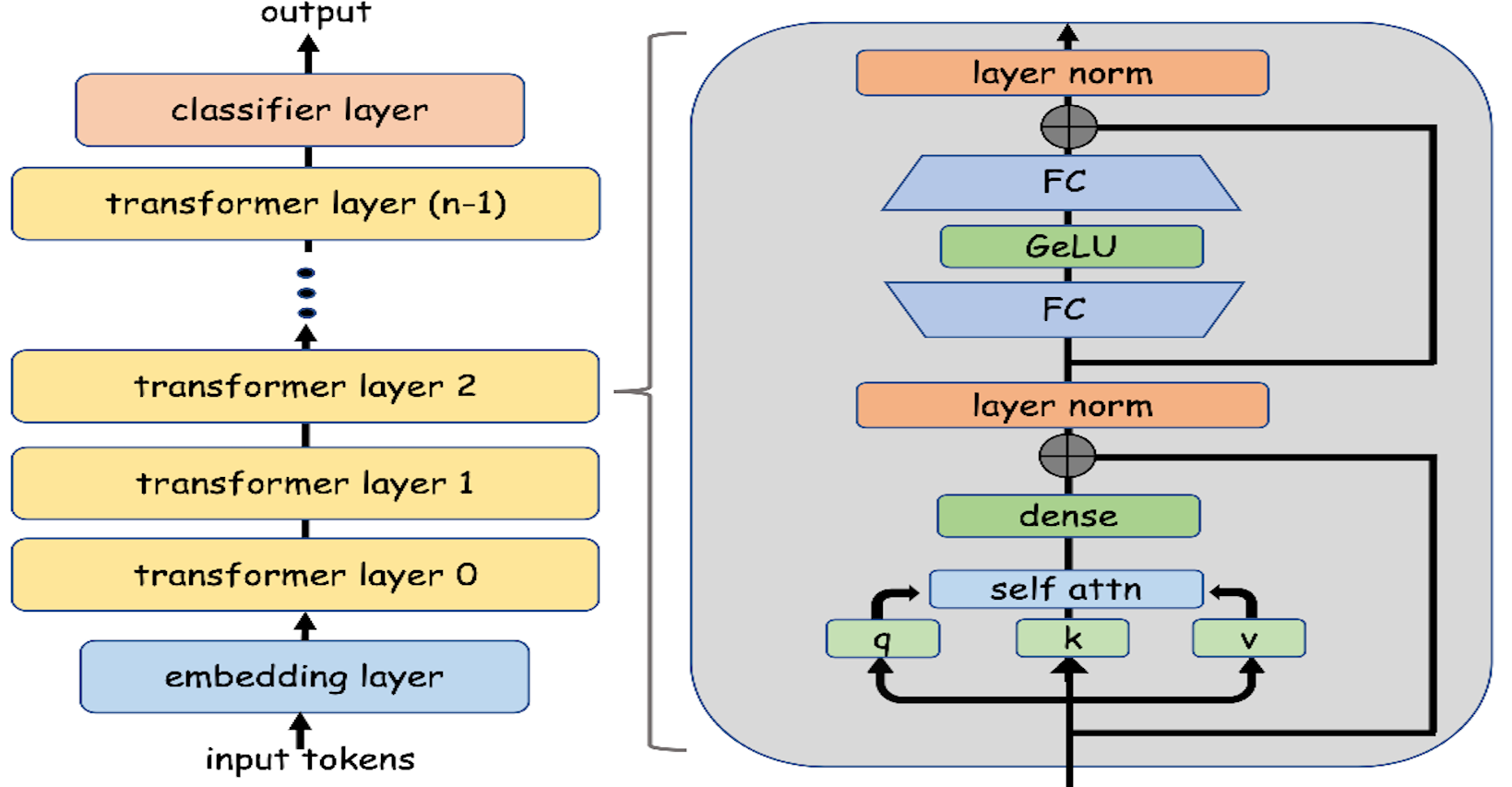}
    }\label{fig:method:arch}
    }
    \subfigure[An example memory request sequence.]{
    \scalebox{0.47}{
    \includegraphics[width=\linewidth]{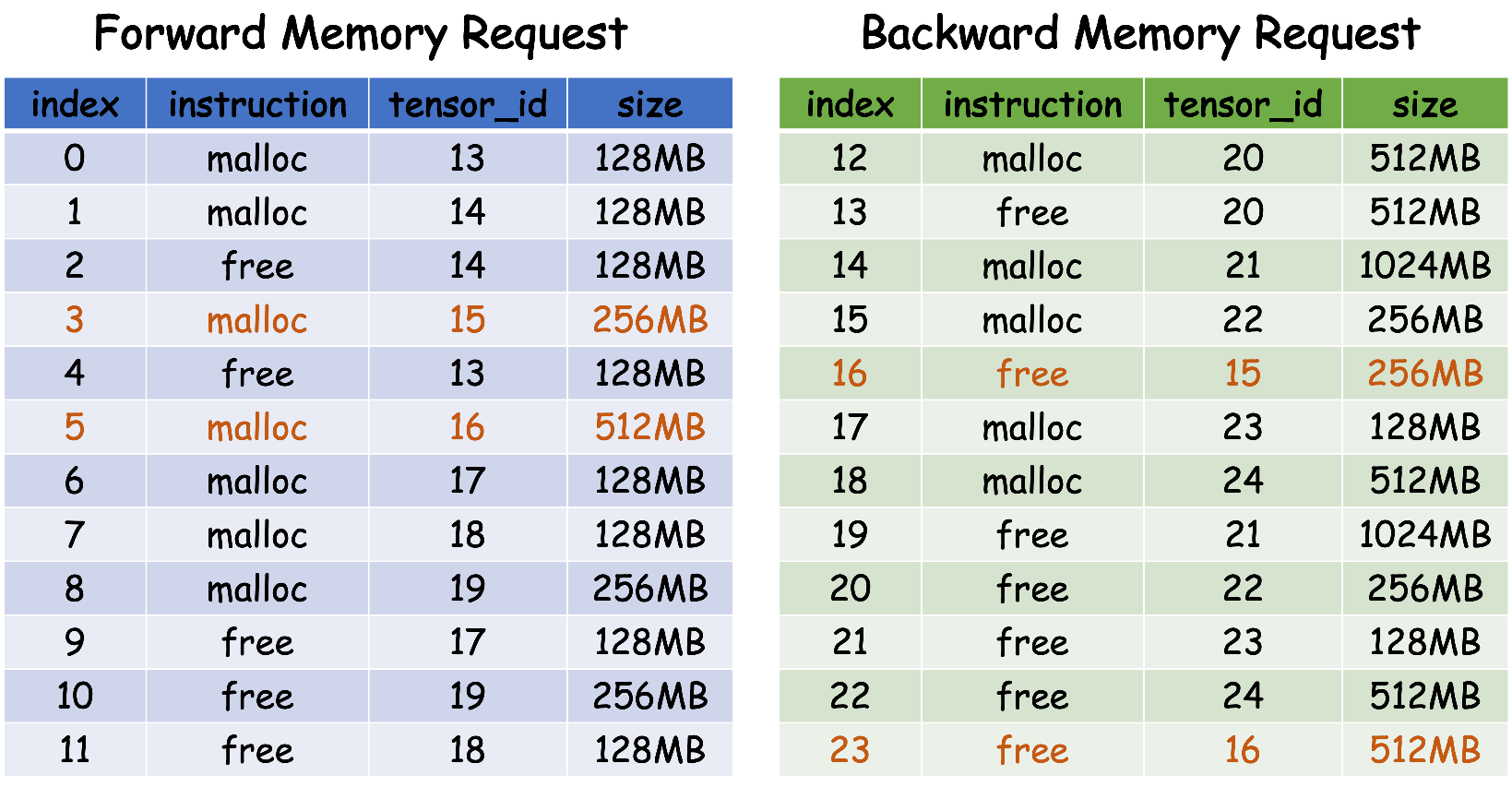}
   }\label{activation-class}
    }
    \caption{(a): The architecture of a typical LLM. (b): An example memory request sequence of a transformer layer's forward and backward pass. Tensors 15 and 16 are skeletal tensors, while the others are transient tensors. }
\end{figure}

\subsection{Large Language Models}
\subsubsection{\textbf{Architecture}}\label{preliminary:llm_arch}

As shown in Figure~\ref{fig:method:arch}, the architecture of an LLM comprises an input embedding layer, multiple transformer layers, and a final classifier layer. 
The embedding layer converts input tokens into continuous representations. 
Each decoder-only transformer layer constitutes a multi-head self-attention module with causal mask, and an Feed-Forward Network (FFN) module containing Fully-Connected (FC) networks. 
The classifier layer takes the hidden states produced by the transformer layers as input, and generates a probability distribution over the vocabulary.


\subsubsection{\textbf{The Training Process}} 

The training process of LLM involves two phases: the forward pass and the backward pass. 
During the forward pass, the model processes the input data through its layers, and finally generates predictions. 
The output tensors of the operators in forward pass are called activation tensors, some of which are stored for backward pass computation according to gradient-based learning.

The backward pass, on the other hand, computes the gradients with regard to the model parameters.
These gradients are used to update the model's parameters.
Following the chain rule in gradient computation, the backward pass relies on the activation tensors from the forward pass to compute gradients.

\subsubsection{\textbf{The Challenge of Huge Memory Requirement in Long Context Training}} 

%
%
%

%
%
%
%
%
%

\add{Self-attention is the most critical module in LLMs. 
It facilitates information interaction between tokens: the input tensor is first projected into query ($Q$), key ($K$), and value ($V$), each with the shape ($b, s, h$); then the output is given by $O = \text{softmax}({QK^{T}}/{\sqrt{d}})\cdot V$, which incurs $O(s^2)$ time and space complexity due to the ($b, s, s$) matrices.
FlashAttention~\cite{flash1, flash2}, the de-facto attention implementation in nowadays LLM computation, processes the computation in tiles, discards intermediate results, and maintains compact states to generate the final output $O$.
This method avoids storing the $O(s^2)$ matrices.
During the backward pass, FlashAttention re-computes these intermediate results in a tiled manner for gradient calculation.
Thanks to this design, FlashAttention significantly reduces memory requirements to just $O(s)$ complexity.
%
%
}
\add{Although several alternative approaches like sparsification~\cite{efficient_transformer} (e.g., BigBird~\cite{zaheer2021bigbirdtransformerslonger}, Longformer~\cite{beltagy2020longformerlongdocumenttransformer}), kernelization (e.g., LinearAttn~\cite{transformersarernn}, CosFormer~\cite{qin2022cosformerrethinkingsoftmaxattention}), and low-rank approximation (e.g., Linformer~\cite{wang2020linformerselfattentionlinearcomplexity}, Nystr\"omformer~\cite{xiong2021nystromformernystrombasedalgorithmapproximating}) also aim to reduce the quadratic memory demands of self-attention, these methods modify the attention mechanism and can potentially compromise accuracy.
In this paper, we adopt FlashAttention as the default method, considering its prevalent use in practice.
} 

Although FlashAttention has reduced the memory complexity of LLM training from $O(s^2)$ to $O(s)$, the linearly scaling activation memory remains the primary challenge in long context training.
For example, as we will elaborate in Section~\ref{sec:anatomy}, when training a 7B GPT model with 32 layers and a hidden size of 4096, using a single 1 million length sequence, the forward activation tensors required by the backward pass consume 4096GB (when using half-precision numbers), whereas the typical memory capacity of a GPU is much smaller.
%
To cope with this issue, there are two lines of efforts, which are the memory reduction techniques and distributed parallelism strategies. 
In the rest of this section, we will introduce these two lines respectively. 
It is worth noting that although our work primarily concentrates on the memory reduction techniques, the proposed \name framework is compatible with a wide range of parallelism strategies.

\subsection{Memory Reduction Techniques}
\label{sec:pre_memory_reduction}

\add{
%
As mentioned in Section~\ref{sec:intro}, when existing query processing engines face the challenge of limited high-bandwidth memory, a common strategy to mitigate memory pressure involves discarding certain data structures and rematerializing them as needed~\cite{spark,faiss,dynamic}. 
There are two prevalent techniques: (1) recomputing the results and (2) swapping data to lower-tier memory and retrieving it when necessary. 
Both methods are also widely-used in neural network training to rematerialize activation tensors.
}


Activation recomputation~\cite{DBLP:journals/corr/ChenXZG16,DBLP:conf/iclr/KirisameLHBHRCT21,DBLP:journals/corr/abs-2205-05198} (a.k.a. activation checkpointing) selectively stores the inputs of certain layers rather than all intermediate activations. 
During the backward pass, the required activations are recomputed on-the-fly. 
While this approach reduces the activation memory footprint required for LLM training, it introduces additional computation, which impacts efficiency.
Swapping~\cite{vdnn,transformerengine,zero-offload}, also known as CPU offloading, aims to relieve the GPU memory pressure by offloading GPU tensors to CPU memory, and fetch them back to GPU when needed. 
\add{Through careful scheduling, the data transmission overhead can be overlapped with GPU computation, a technique also popular in GPU databases~\cite{faiss, gpudatabaseoverlap1, gpudatabaseoverlap2}. 
However, if data transmission is too time-consuming to overlap, swapping can significantly slow down training.
}
%
In general, both memory reduction techniques release the memory of activations in the forward pass, but need to rematerialize them in the backward pass, at the price of extra computation or data transmission overhead, respectively. 

In the past few years, GPU computing capabilities have improved over 100$\times$ (e.g., the half-precision performance of P100 and H100 are 18.7 and 1979 TFLOPS, respectively), while the improvement of CPU-GPU bandwidth is only 4$\times$ (from PCIe 3.0 to PCIe 5.0). 
As a consequence, mainstream LLM training frameworks favor the activation recomputation technique.\footnote{Both Megatron-LM and DeepSpeed have supported activation recomputation for long. Nevertheless, Megatron-LM does not support swapping until the release of TransformerEngine v1.3 in Feb 2024. Besides, DeepSpeed primarily focuses on swapping of model states, encompassing model parameters, gradients and optimizer states~\cite{zero-offload}, as they constitutes the most significant portion of memory footprint in short context training tasks. However, in long context training scenarios, the memory consumption of activations has surpassed that of model states.}
In practice, when training LLMs with long context input, full activation recomputation is often employed, which involves storing only the input tensor of each transformer layer and recomputing the required activations during backward propagation.


\subsection{Distributed Parallelism Strategies}
Distributed training is essential for efficiently training LLMs, especially in scenarios of long context training. 
To facilitate the training of large-scale data and model, several distributed parallelism strategies have been proposed.

\underline{\textit{Data Parallelism}} (DP)~\cite{DBLP:conf/nips/DeanCMCDLMRSTYN12,DBLP:conf/nips/ZinkevichWSL10,limu} duplicates model parameters and distributes the input data across multiple devices. 
Each device holds a complete copy of the model and processes its input data independently. 
After backward propagation, the devices synchronize parameter gradients to ensure consistency across the model copies. 

\underline{\textit{Zero-Redundancy Optimizer}} (ZeRO)~\cite{zero} is a series of variants built upon DP, aiming to alleviate memory pressure. 
%
Naive DP replicates model parameters, gradients and optimizer states among all devices.
ZeRO is designed in three stages to reduce these memory requirements respectively. 
First, ZeRO-1 partitions the optimizer states among all DP workers. 
Next, ZeRO-2 extends ZeRO-1 by also partitioning gradients, further reducing memory footprint. 
Finally, ZeRO-3, based on ZeRO-2, partitions model parameters among DP workers, further mitigating memory pressure but introducing additional communication to gather parameters during training.

\underline{\textit{Tensor Parallelism}} (TP)~\cite{DBLP:journals/corr/abs-1909-08053} partitions the self-attention and feed-forward modules of transformer layers across multiple devices along either the column or row dimension. 
It addresses the problem that LLMs can not fit into the memory of a single device. 
It involves extra collective communication operations (i.e. AllReduce) to synchronize the intermediate results. 
Therefore, TP is usually applied within a computing node, where intra-node GPUs are connected via high-bandwidth NVLink.

\underline{\textit{Pipeline Parallelism}} (PP)~\cite{DBLP:conf/nips/HuangCBFCCLNLWC19,DBLP:conf/icml/NarayananPSCZ21, jcst1} is also proposed to address the problem that LLMs cannot be fit into a single device. 
Different from TP, PP partitions model layers into several stages, then distributes the stages to different devices. 
The input data is processed through these stages in a pipeline fashion. 
Given the peer-to-peer communication style, the PP stages are often distributed across nodes.
However, PP introduces a phenomenon known as ``bubble'', which corresponds to GPU idle time.
The issue becomes more severe when the number of micro-batches is small.
%

To facilitate efficient long context training, several novel parallelism strategies have been proposed recently.

\underline{\textit{Sequence Parallelism}} (SP)~\cite{DBLP:journals/corr/abs-2205-05198} is built upon TP to further reduce activation memory overhead. 
It splits the sequence dimension in the part of the model that does not apply TP.
The original AllReduce communication now transitions to AllGather and ReduceScatter.

\underline{\textit{DeepSpeed-Ulysses}}~\cite{DBLP:journals/corr/abs-2309-14509}, built upon ZeRO, is another form of sequence parallelism.
During self-attention computation, it splits the head dimension, whereas in other model components, it partitions the sequence dimension.
For transitioning between modules, it utilizes AllToAll communications, theoretically reducing communication overhead compared to SP. 
However, its SP degree is limited by the number of heads in self-attention.
To further relieve the memory pressure, DeepSpeed-Ulysses leverages ZeRO to distribute model parameters.

\underline{\textit{Context Parallelism}} (CP)~\cite{DBLP:journals/corr/abs-2310-01889,DBLP:journals/corr/abs-2310-03294,contextparallel} shards the query, key, and value matrices within the attention module along the sequence dimension across different devices.
During attention computation, necessary communications are involved to ensure consistent results, which can be overlapped with computation by careful scheduling. 

In practice, these parallelism strategies and memory reduction techniques can be integrated and employed simultaneously to facilitate efficient training of LLMs.


\section{Anatomy and System Desiderata}
\label{sec:anatomy}

Managing fragmented massive data storage is a critical data management issue, particularly when workloads are constrained by limited high-bandwidth memory.
Long-context LLM training demonstrates such a challenge, where huge fragmented activation memory significantly impedes efficient training within constrained GPU memory resources.
%

%
In this section, we first provide an in-depth anatomy of the key characteristics of activation data storage in long-context LLM training. Based on this analysis, we present the design desiderata that motivates the development of \name.

\subsection{Categorization of Activation Tensors}

In long-context LLM training, the primary memory consumption originates from activation tensors, which are the outputs of computing operators in LLMs.
According to their life cycles, we can categorize activations generated during the forward propagation into two classes, which are the \ul{\textit{skeletal activations}} and the \ul{\textit{transient activations}}, where the former is necessary for the backward propagation while the latter is not.

For illustration, in Figure~\ref{activation-class}, tensors 13, 14, 17, 18, and 19 are produced during the forward pass of a transformer layer, and are discarded before the completion of this layer's forward pass.
Similarly, tensors 20, 21, 22, 23, and 24 are generated during the backward pass of this layer, and are discarded after corresponding computation. 
We term them ``transient tensors'' because they are created and discarded within a single layer's forward or backward pass. Transient tensors usually serve as temporary results.
Conversely, tensors 15 and 16 are generated during the forward propagation and are needed for backward propagation, so they are discarded in this layer's backward pass. 
We refer to these tensors as ``skeletal tensors'' because they are produced during the forward pass, and are essential for the gradient calculation during the backward pass.

\subsection{Analysis of Skeletal Activations}

\begin{figure}[!t]
  \centering
  \includegraphics[width=0.5\linewidth]{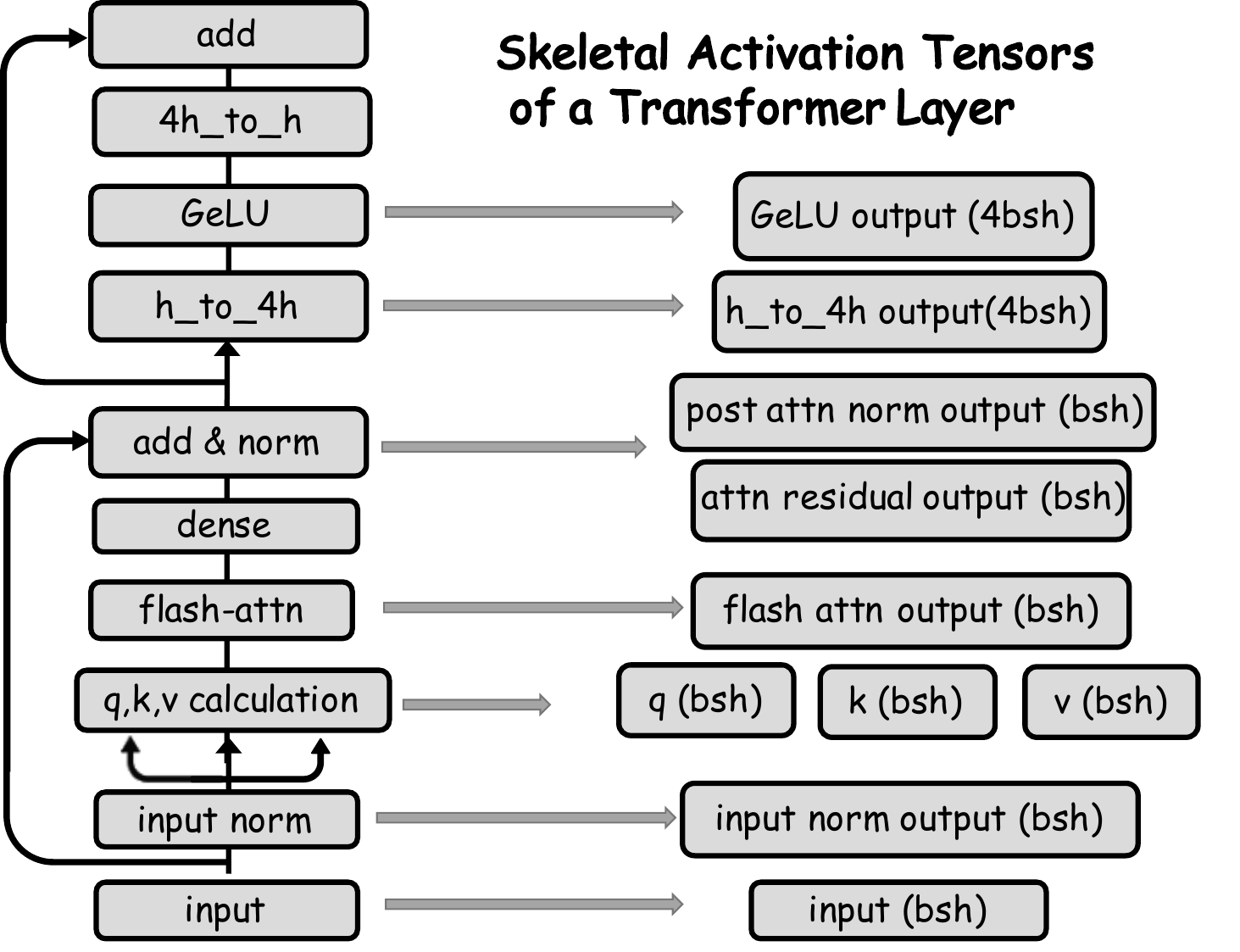}
  \caption{Illustration of the transformer layer architecture. The sizes of skeletal activations are provided in the brackets.}
  \Description{Time consumption}
  \label{fig:method:transformer_arch_and_skeletal_tensors}
\end{figure}

\begin{figure*}[!t]
  \centering
  \includegraphics[width=0.9\linewidth]{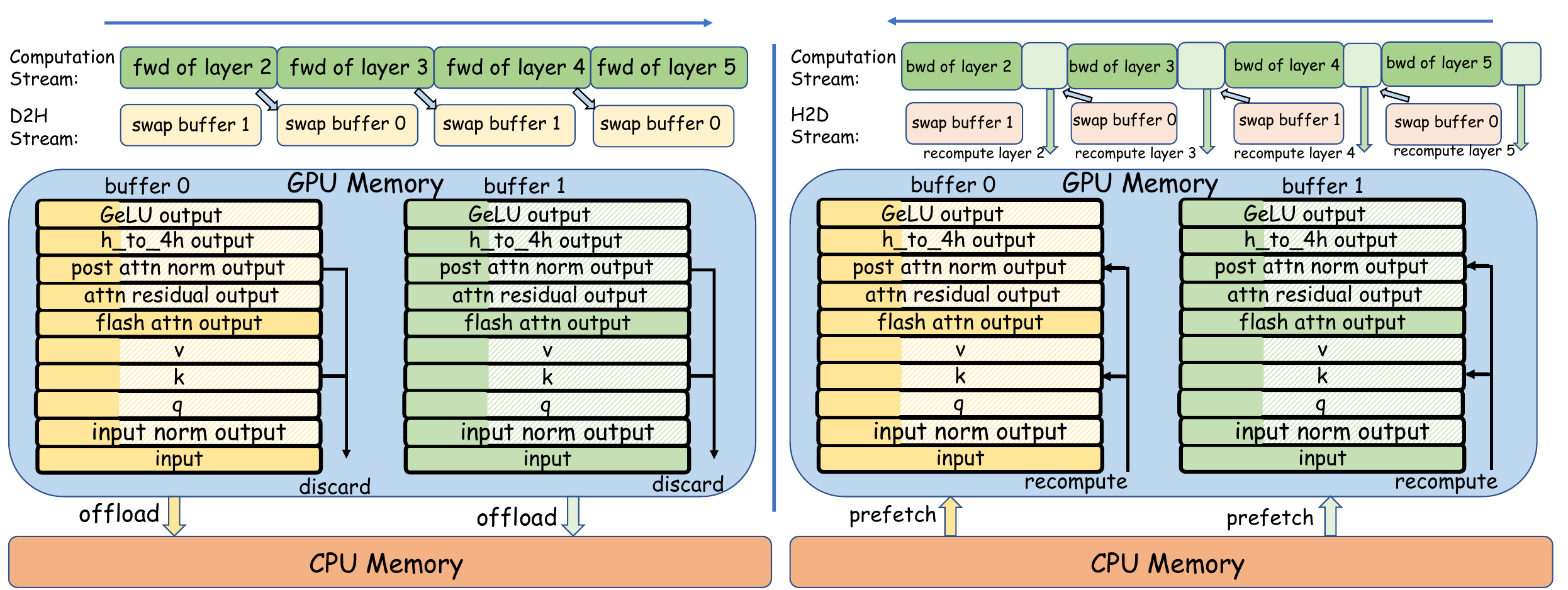}
  \caption{Forward and backward propagation with rounding buffers for token-wise recomputation/swapping. During forward propagation, the darker part in the rounding buffers is offloaded to CPU, while the lighter part is discarded; during backward propagation, the darker part in the rounding buffers is prefetched from CPU, while the lighter part is recomputed.}
  \Description{Rounding buffers with token-wise recomputation}
  \label{fig:method:token_wise_fwd_bwd}
\end{figure*}

%
%
%
%

%

Figure~\ref{fig:method:transformer_arch_and_skeletal_tensors} presents all skeletal tensors generated within a transformer layer's forward propagation, along with their sizes.
We can see that the total size of all skeletal activations in a single transformer layer amounts to $16bsh$.
To exemplify, when training the GPT-7B model ($h = 4096$, 32 layers) with a sequence length ($s$) of 1 million, if we store the skeletal activations in half-precision floating numbers, it would take 4096 GB for only one sequence ($b = 1$), exceeding the memory capacity of even 50 A100/H100 GPUs. 

An important characteristic of skeletal activations is that they are needed by backward computation, so they must reside in GPU at least before the backward propagation of the corresponding transformer layer begins. 
However, maintaining all skeletal activations for backward propagation is infeasible. 
To this end, memory-saving techniques like recomputation and swapping, which are also prevalent in traditional data management problems~\cite{spark,faiss,dynamic}, become necessary for long context training.
In LLM training, these techniques first release the skeletal activations of a transformer layer in the forward propagation, and later rematerialize them before the corresponding backward propagation. 

Unfortunately, na\"ively applying activation recomputation or swapping  is insufficient to tackle the challenge of managing large-scale skeletal activations.
Both techniques trade time for memory --- the activation recomputation technique incurs extra computation overhead while the swapping technique necessitates transmitting the activations from CPU memory to GPU memory. 
Using recomputation alone incurs significant additional computation overhead, while employing swapping alone can lead to CPU OOM error (when the sequence is too long) or block GPU computation (when the swapping time cannot be fully overlapped).
%
%
%
As a result, we desiderate a meticulous orchestration of the two memory-saving techniques to manage the skeletal activations, so that we can minimize the extra overhead while accommodating the huge memory requirement in long context training of LLMs. 
To achieve this, we develop a \textit{token-wise activation recomputation and swapping} mechanism, which will be demonstrated in Section~\ref{subsec:tokenwise}. 



\subsection{Analysis of Transient Activations}
Transient activations are intermediate results generated and discarded during the forward (or backward) pass of a transformer layer. 
Actually, there are more transient activations than skeletal activations in a transformer layer.
Specifically, we observe that the number of transient activations can exceed 5 times that of skeletal activations. 
Without careful management, the frequent allocation and deallocation can lead to memory fragmentation, which degrades system performance and poses a major concern for managing massive data storage.
There have been studies in the data management community focusing on memory defragmentation~\cite{defragment1, defragment2}, leveraging the characteristics of target workloads to devise appropriate and innovative methods.
Following the methodology, we analyze the LLM training process and attempt to defragment the tensor memory.
During training, memory requests are identical across both transformer layers and training iterations, providing an opportunity to manage and reuse these memory regions effectively to minimize fragmentation.
%
%
In particular, the memory addresses of a single transformer layer's transient activation tensors can be reused by all other transformer layer's corresponding transient activation tensors. 
However, in practice, memory reuse is not fulfilled because the PyTorch caching allocator lack prior information of the memory request sequence during training iterations. 
This inspires us to statically plan the memory addresses of each transformer layer's transient tensors, which will be described in detail in Section~\ref{subsec:memoryplanning}.

\section{\name Design}

In this section, we propose \name for fine-grained activation memory management.
Our proposed method leverages fine-grained and structured activation management, akin to concise memos that share vital information. 
The main challenge of long context training is the large activation size which scales linearly w.r.t. sequence length. 
We propose token-wise activation recomputation and swapping, along with a bi-level memory planning to address the issue, which targets skeletal activations and transient activations, respectively.
The overview of \name is depicted in Figure~\ref{fig:method:memo_overview}.

\subsection{Token-wise Recomputation and Swapping}
\label{subsec:tokenwise}

Skeletal tensors, generated during the forward pass of a transformer layer, must reside in GPU memory for the subsequent backward propagation.
%
In practice, as sequence length grows, the size of skeletal activations increases linearly, which can easily exceed the capacity of GPU memory. 
As introduced in Section~\ref{sec:pre_memory_reduction}, currently the most widely-used technique to tackle this issue is activation recomputation, which stores only the input of each transformer layer, and discards the rest skeletal activation tensors of this layer. 
%
Prior to backward propagation of each layer, an additional forward pass of the layer is conducted to reconstruct all skeletal tensors so that the backward computation can be carried out.
However, we note that the vanilla activation recomputation strategy is not an optimal choice to handle the challenge of linearly increasing skeletal activation memory, considering the following two reasons: 
(1) activation recomputation introduces redundant computation, thus diminishing  training efficiency; and (2) the memory overhead of retaining the input tensor of each transformer layer can still be expensive, especially when the sequence length is too long or the number of layers is too large. 
Take the training of GPT-7B with a context length of 1 million as an example again. 
For only one sequence, the input tensors of all 32 transformer layers together consume 128GB. 
Even using a SP degree of 8, it takes 16GB for each GPU to store the input tensors of all 32 transformer layers, which already takes up to 20\% of total GPU memory capacity.

As explained in Observation~\ref{intro:observ:1}, the computation complexity of FlashAttention w.r.t. sequence length is $O(s^2)$, while the size of skeletal activations within a transformer layer scales linearly with sequence length. This provides us with the opportunity to offload skeletal activations to CPU memory, thereby saving GPU memory. We can prefetch them back to GPU before the backward propagation of the corresponding transformer layer. The swapping of skeletal activations can overlap with GPU computations in long context training, since the CPU-GPU data transmission does not consume GPU computation units.

To facilitate the overlapping, we utilize two rounding GPU buffers to store the skeletal activations for all transformer layers. The two rounding buffers are allocated before the actual training iterations begin.
As shown in Figure~\ref{fig:method:token_wise_fwd_bwd}, transformer layers with even layer indices place their skeletal activation tensors in rounding buffer 0, while layers with odd layer indices use rounding buffer 1.

After the computation of transformer layer $i$, rounding buffer $(i\%2)$ will be offloaded to CPU using a separate CUDA stream. This happens simultaneously with the computation of transformer layer $(i+1)$. Before the forward computation of transformer layer $(i+2)$, a CUDA event is employed to ensure the content of rounding buffer $(i\%2)$ has been fully offloaded to CPU memory, thus the transformer layer $(i+2)$ can safely rewrite rounding buffer $(i\%2)$.

For backward propagation, after the backward pass of transformer layer $(i+2)$ ends, the contents within rounding buffer $(i\%2)$ become useless, and we start prefetching the skeletal activations of transformer layer $i$ to rounding buffer $(i\%2)$ using another CUDA stream. The prefetching of transformer layer $i$'s skeletal activations happens simultaneously with the backward propagation of transformer layer $(i+1)$.
When the sequence length is sufficiently long, with careful computation-transmission overlapping and synchronization, CPU swapping can substitute activation recomputation without incurring additional overhead. 

However, there are two constraints that prevent us from offloading all skeletal activations to CPU memory. 
\begin{itemize}[leftmargin=*,parsep=0pt,itemsep=0pt,topsep=2pt,partopsep=2pt]
    \item For sequence lengths that are not sufficiently long, the time required to offload all skeletal activations to CPU memory surpasses the computation time for a single transformer layer. This discrepancy forces the computation of transformer layer $(i+2)$ to be delayed until the offloading of rounding buffer to CPU memory is completed, thereby blocks the normal GPU computation workflow. For instance, as illustrated in Figure~\ref{fig:intro:time}, when training a 7B GPT model on 8 GPUs with a TP degree of 8, ideal overlap between the computation of a transformer layer and the offloading of its skeletal activations occurs only for sequence lengths exceeding 192K.   In practice, the sequence lengths of most training datasets are moderate and may be not sufficient to ensure an ideal overlap between computation and transmission.

    \item In theory, a longer sequence length provides more  opportunities for overlapping swapping with GPU computation. However, in practice, the CPU memory is often limited. For a typical GPU server which has several terabytes CPU memory (e.g. 2TB in our environment), it is insufficient to store all skeletal activations when the sequence length is excessively long or the number of transformer layers is too large. For instance, when training the 7B model on a server equipped with 8 GPUs using a sequence length of 1 million, the skeletal activations amount to a total size of 4096GB, which is double the capacity of CPU memory.
    
\end{itemize}

\begin{figure}[!t]
  \centering
  \includegraphics[width=0.75\linewidth]{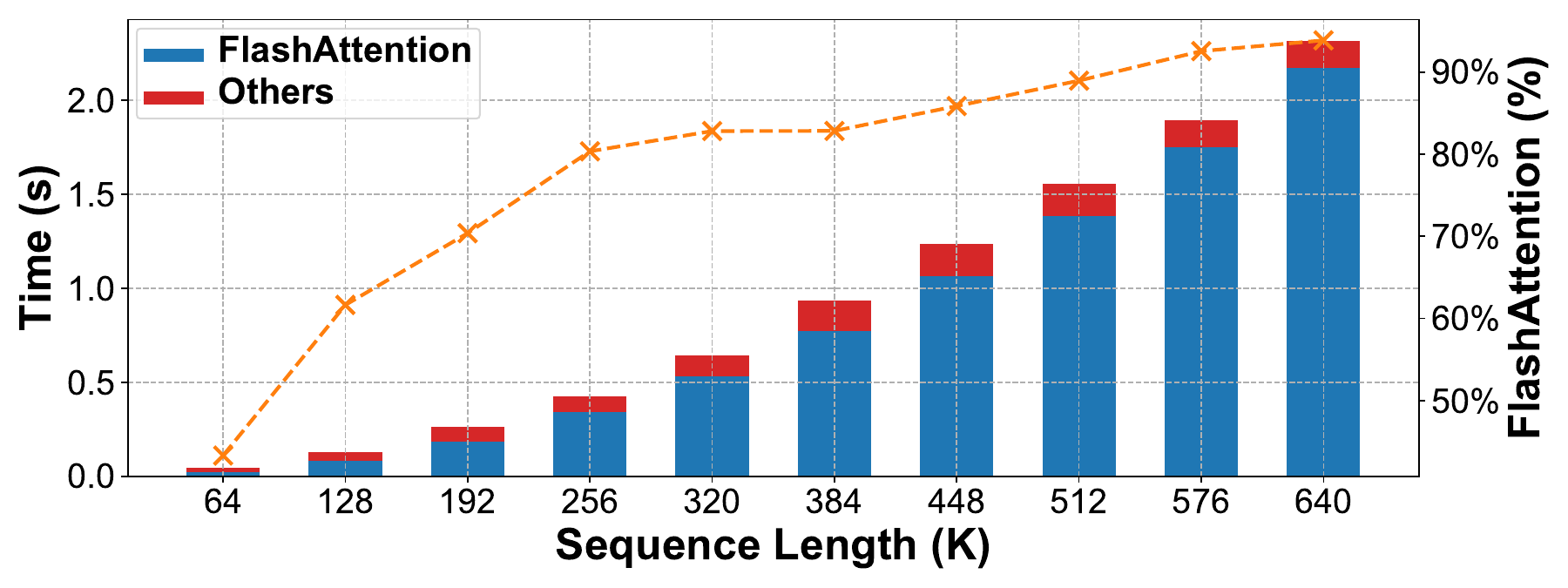}
  \caption{Forward time of FlashAttention and other parts of a transformer layer when training a 7B GPT on 8 GPUs with a TP degree of 8.}
  \Description{Forward time of FlashAttention and other parts of a transformer layer}
  \label{fig:method:percentage}
\end{figure}

Therefore, instead of simply offloading all skeletal activations to CPU memory, we employ selective activation swapping to ensure perfect overlap of computation and transmission for short sequences as well as to avoid depleting CPU memory for extremely long context lengths.
\name determines the selection of swapping at both the tensor and token granularities, as depicted in Figure~\ref{fig:method:token_wise_fwd_bwd}. 

At the tensor granularity, we consider the benefits of leveraging the swapping technique rather than the recomputation technique of different modules. 
As depicted in Figure~\ref{fig:method:percentage}, FlashAttention constitutes the most substantial portion of the forward computation of a transformer layer. Notably, when the sequence length exceeds 576K, FlashAttention accounts for more than 90\% of the computation involved in a single transformer layer.  However, as illustrated in Figure~\ref{fig:method:transformer_arch_and_skeletal_tensors}, the output of FlashAttention only accounts for 6.25\% of total skeletal activation size. 
This inspires us to offload the entire output tensor of FlashAttention to CPU memory since recomputing its output is very time-consuming. 
Besides, since LLMs have a layered structure, in order to reconstruct the ``input\_norm'', ``q'', ``k'', ``v'' tensors, we also store the input of each transformer layer to CPU, following common recomputation strategy~\cite{DBLP:journals/corr/ChenXZG16}.  


At the token granularity, we develop the token-wise activation recomputation and swapping technique to reduce the memory consumption of all skeletal activation tensors other than the output of FlashAttention and the input of each layer. 
To be specific, as shown in Figure~\ref{fig:method:token_wise_fwd_bwd}, for each of these skeletal activation tensors, we offload a fraction (denoted as $\alpha$) to CPU, while the remaining part is discarded, ensuring perfect overlapping and to avoid CPU OOM error. Before the backward pass, the discarded part is rematerialized via recomputation while the offloaded part is prefetched.

To determine the fraction $\alpha$, we solve the following problem:

\begin{align}
\nonumber
\max \quad & \alpha, \\
\nonumber
\text{s.t.} \quad & (S_{input} + S_{attn} + \alpha\cdot S_{others})/B \le T_{layer}, \\
\nonumber
& (n-2)(S_{input} + S_{attn} + \alpha\cdot S_{others}) \le M_{CPU}.
\label{inequality_alpha}
\nonumber
\end{align}

where $S_{input}$, $S_{attn}$, and $S_{other}$ stand for the size of input tensor,  the size of FlashAttention output tensor, the total size of other skeletal activation tensors, respectively; $B$ is the PCIe bandwidth between GPU and CPU, $T_{layer}$ is the forward time of a single transformer layer, $n$ is the total number of transformer layers, and $M_{CPU}$ stands for the capacity of CPU memory.
It is worth noting that, the last two transformer layers can initiate the backward pass immediately after the forward pass, obviating the need for swapping. 
These variables can be easily obtained through profiling before training, so we can determine an appropriate $\alpha$ without much effort.

\subsection{Bi-level Memory Planning}
\label{subsec:memoryplanning}

\begin{figure}[!t]
  \centering
  \includegraphics[width=0.75\linewidth]{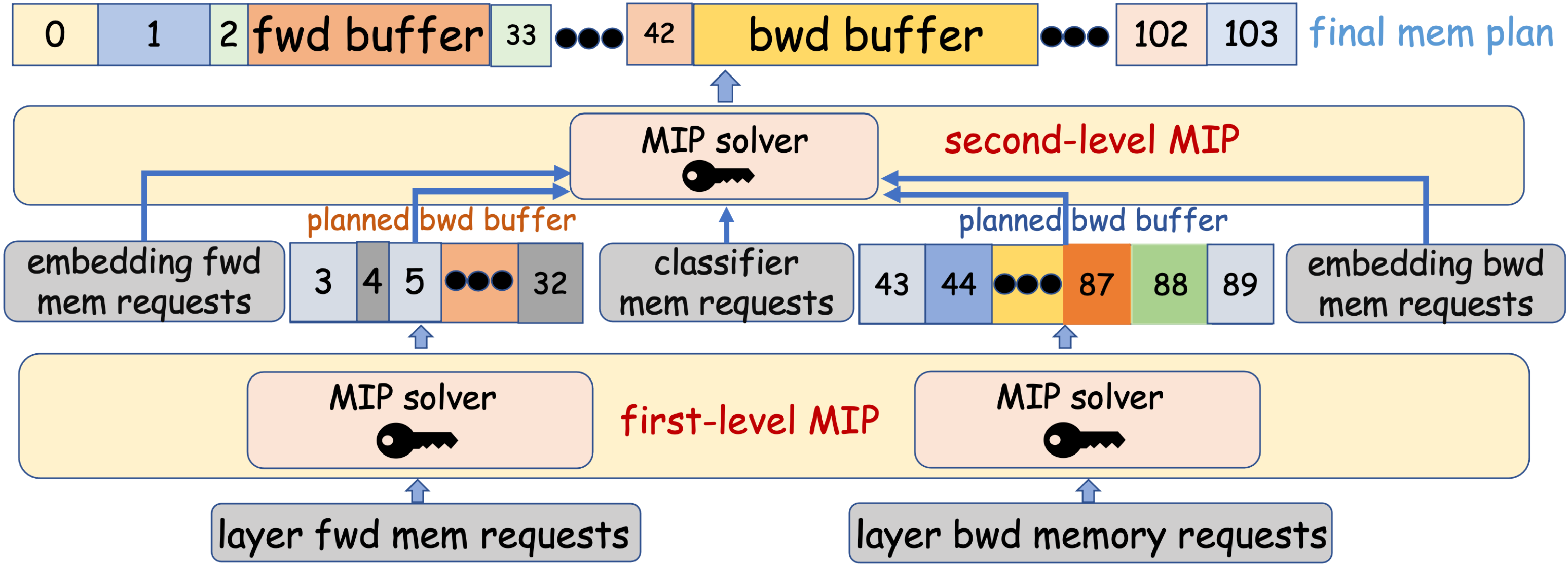}
  \caption{Bi-level MIP algorithm.}
  \Description{bi-level}
  \label{fig:method:bilevel}
\end{figure}

In the previous subsection, we have tackled the management of skeletal activations by the fine-grained recomputation and swapping technique.
However, frequent allocation and deallocation of the transient activation tensors still lead to GPU memory fragmentation, which forces the allocator to frequently reorganize GPU memory using time-consuming ``cudaFree'' and  ``cudaMalloc'' operations.
To address the issue, and to achieve full reuse of GPU memory across all transformer layers, we design a bi-level Mixed Integer Programming (MIP) method.

In practice, our initial step involves profiling the sequence of memory requests during a single training iteration. Given the memory request sequence, the challenge lies in determining the address of each requested tensor while at the same time minimizing the peak memory usage. This task aligns with the well-established offline Dynamic Storage Allocation (DSA) problem~\cite{sekiyama2018profileguidedmemoryoptimizationdeep}, which can be formulated as a Mixed Integer Programming (MIP) problem. 
A concise overview of this formulation is shown as follows.

The offline DSA problem handles a sequence of memory allocations and deallocations, and aims to determine the address of each allocated memory block and at the same time minimizing the peak memory usage. Parameters of offline DSA problem includes:
\begin{itemize}[leftmargin=*,parsep=0pt,itemsep=0pt,topsep=2pt,partopsep=2pt]
    \item  $n$, the number of requested tensors.
    \item  $S_{i}$, the size of requested tensor $i$, for $\forall i\in \{1,2,...,n\}$.
    \item  $E = \{(i,j)| \text{tensor } i,j  \text{ have overlapped lifespan} \}$.
\end{itemize}

And the problem can be written as

\begin{equation*}
\begin{split}
&\, \, \, \min \,\, \, \, M, \\
&s.t.\quad  \left\{\begin{array}{lc}
A_{i} +  S_{i} \leq M, i \in \{1,2,...,n\}, \\
A_{i} +  S_{i} \leq A_{j} + z_{ij} \cdot M_{cap}, (i,j)\in E, \\
A_{j} +  S_{j} \leq A_{i} + (1-z_{ij}) \cdot M_{cap}, (i,j)\in E,  \\
0 \leq M \leq M_{cap}, \\
A_{i} \geq 0, i \in \{1,2,...,n\},
\end{array}\right.
\end{split}
\end{equation*}
where $A_{i}$ stands for the address of requested tensor $i$, $M$ stands for the peak memory usage, $M_{cap}$ is the memory capacity, and $z_{ij}$ is defined as
\begin{equation}
z_{ij}=\left\{
\begin{array}{cl}
0,  & A_{i} + S_{i} \leq A_{j}, (i, j) \in E,  \\
1,  & A_{j} + S_{j} \leq A_{i}, (i, j) \in E.    \\
\end{array} \right. 
\nonumber
\end{equation}

Here the first constraint and the last two constraints define and limit peak memory, while the second and third constraints ensure non-overlapping tensors.
Following this formulation, the solution for each tensor's address is optimal. 
\add{However, modern LLM training involves thousands of allocation and deallocation requests within a single training iteration, which makes this NP-hard MIP problem computationally intractable. 
Consequently, it's infeasible to solve this MIP problem in one pass, given the prohibitively high time cost.}

Fortunately, all transformer layers have identical structures and memory request sequences, which presents repetitive substructures within the MIP problem. 
By leveraging this inherent repetitiveness, we can instead devise a bi-level hierarchical MIP optimization algorithm, which is both computationally feasible and effective.  
\begin{figure}[!t]
  \centering
  \includegraphics[width=0.75\linewidth]{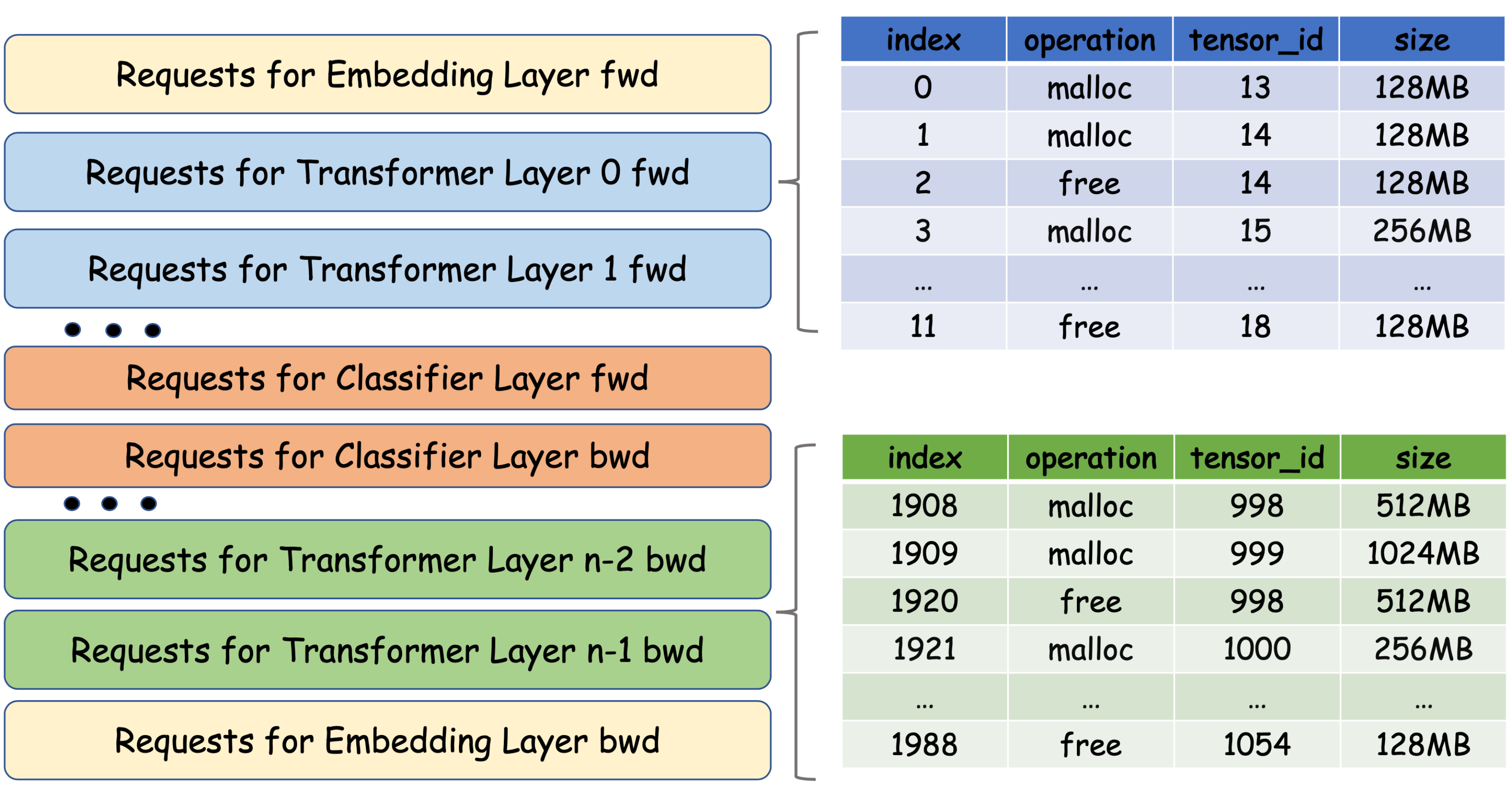}
  \caption{Memory request sequence during training. }
  \Description{transient request sequence}
  \label{fig:method:transient}
\end{figure}

As discussed in Section~\ref{preliminary:llm_arch}, a typical LLM consists of an embedding layer, $n$ consecutive transformer layers, and a final classification layer. 
As shown in Figure~\ref{fig:method:transient}, each layer has forward memory request sequence and backward memory request sequence. The memory request sequence is in the form of a sequence of ``malloc tensor\_id size'' and ``free tensor\_id size''.
%
Since all transformer layers in an LLM are identical, they have the same forward/backward pass memory request sequence. 

As shown in the bottom of Figure~\ref{fig:method:bilevel}, we first solve the offline DSA sub-problem for just one transformer layer's forward (backward) pass, which is called the first-level MIP. \add{The scale of the first-level MIP is small enough to tackle.} This offline DSA problem can be simply solved by any MIP solver \add{(e.g. Gurobi~\cite{gurobi})}.  After this step, the peak memory needed for the forward (backward) propagation of a single transformer layer, as well as the address of each transient tensor within a transformer layer is determined.
After solving the sub-problem for one transformer layer, all other transformer layers can reuse the same memory address for (de)allocation. 

\add{Besides identical transformer layers, LLMs also have other layers that process input tokens and classify output tokens.
Therefore, after solving the first MIP for one-layer DSA problem, we conduct the second MIP for the whole LLM training to generate the peak memory requirement and addresses of all transient activation tensors. }
\add{To simplify the optimization process,} we can replace the original fine-grained memory request sequence of each transformer layer's forward (backward) propagation with a ``pseudo'' large memory request pair, as shown in Figure~\ref{fig:method:bilevel}.
\add{The size of this ``pseudo'' memory block corresponds to the memory usage of each transformer layer as determined by the first-level MIP.}
After the substitution, this reformulated memory request sequence also satisfies the formulation of an offline DSA problem, with a size small enough to be efficiently solved. 
We then leverage the MIP solver again to solve this second-level MIP problem.
After this step, the addresses of all activation tensors, and the peak memory needed for all transient activation tensors can be determined.

\subsection{System Implementation}

\begin{figure}[!t]
  \centering
  \includegraphics[width=0.8\linewidth]{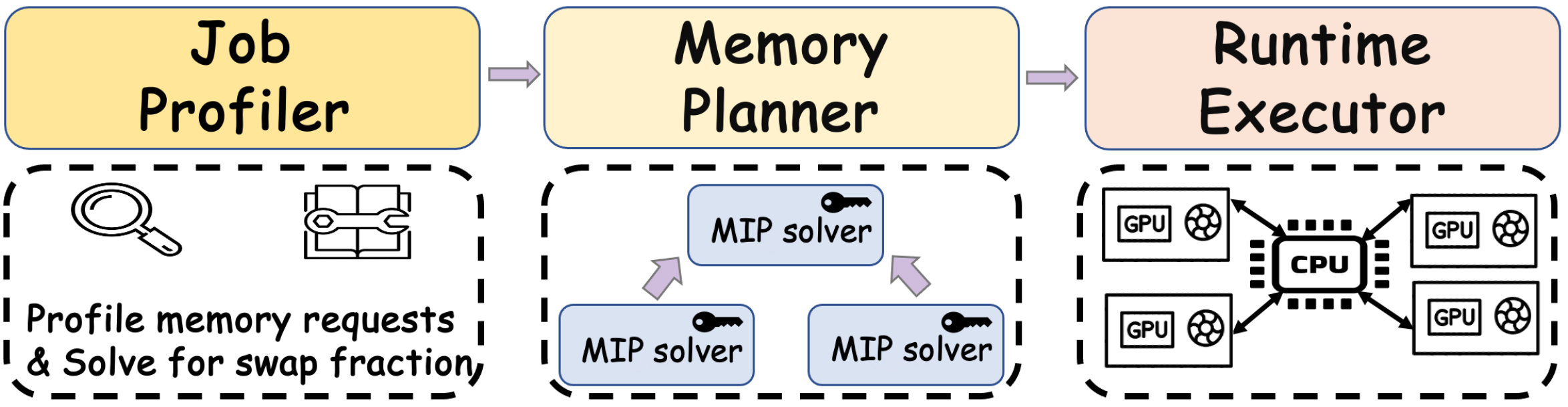}
  \caption{Overall architecture of \name.}
  \Description{Overall architecture of our system}
  \label{system-architecture}
\end{figure}

\subsubsection{\textbf{Overview}}
Figure~\ref{system-architecture} illustrates the overall architecture of \name. 
First, the job profiler takes in the model configuration, then executes a training iteration to profile the memory requests directed to the PyTorch CUDA allocator during the training phase. The job profiler also determines offloading fraction $\alpha$ by solving optimization problem in Section~\ref{subsec:tokenwise}. 
These memory requests comprise a sequence of allocation and deallocation instructions. 
Afterwards, the memory planner receives the memory requests, executes the bi-level MIP optimization algorithm and, generates a memory plan, which constitutes the addresses of all transient activation tensors during one training iteration. 
Finally, the runtime executor reads the memory plan and conducts the training process.

\subsubsection{\textbf{Job Profiler}}
The job profiler is designed to profile the memory request sequence during a training iteration. To implement the module, we have extended the PyTorch CUDA allocator with extra interfaces that log each memory request it receives, in the format of ``malloc tenosr\_id size'' and ``free tensor\_id size''.  

However, naively recording all memory requests may lead to OOM error.
For example, directly profiling a GPT-7B model with a sequence length of 512K on 8 GPUs can result in OOM error. 
Fortunately, all transformer layers have identical memory footprint. We leverage this property by only profiling one transformer layer's memory footprint and then applying it to all transformer layers.

When the sequence is too long, we cannot even profile one single transformer layer.
In such extreme cases, we turn to the CUDA Unified Memory feature, which enables the swapping between GPU memory and CPU memory under the hook, creating an illusion of unlimited GPU memory. 
By integrating CUDA Unified Memory support into the PyTorch CUDA allocator, we have successfully managed to profile the training of extremely long context lengths.

The profiler also gathers the basic information to determine $\alpha$ in Section~\ref{subsec:tokenwise}, including the size of each skeletal activation tensor, and the forward time of a one layer.  Subsequently, it solves for the optimal $\alpha$ to maximize the overlapping of computation and transmission as well as to avoid CPU OOM error. 

\subsubsection{\textbf{Memory Planner}}
Given the memory request sequence generated by the job profiler, memory planner executes the bi-level MIP optimization algorithm as introduced in Section~\ref{subsec:memoryplanning} to generate a memory plan, which includes the address of each transient activation tensor and the peak memory usage needed during training. \add{\name uses the Gurobi~\cite{gurobi} optimizer to solve the MIP problems.}  In all our experiments, memory planning takes less than 5 minutes, which is negligible compared to the training time of LLMs.

\begin{figure}[!t]
  \centering
  \includegraphics[width=0.7\linewidth]{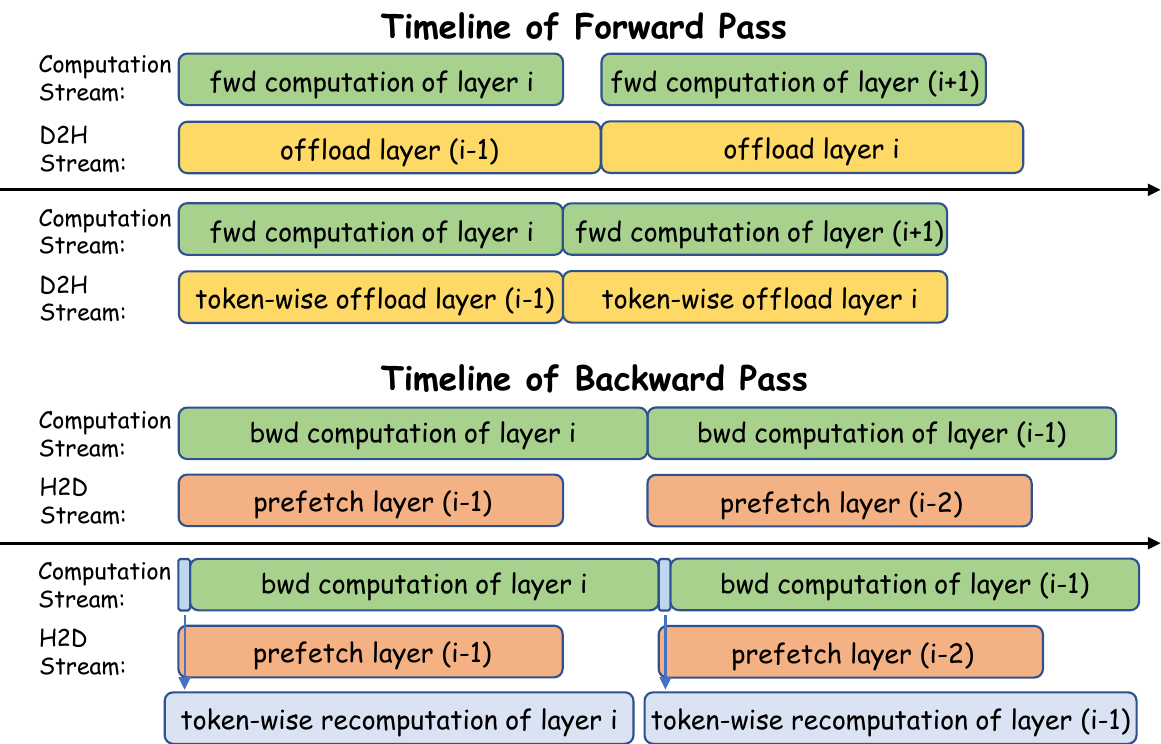}
  \caption{Scheduling of computation, offloading and prefetching w/ and w/o token-wise recomputation. Given the superior computing ability of modern GPUs, the recomputation part is faster than the offloading part that blocks forward computation.}
  \label{fig:method:scheduling}
  \Description{scheduling of computation, offloading and prefetching during training, without token-wise recomputation.}
\end{figure}

\subsubsection{\textbf{Runtime Executor}}
The runtime executor takes the memory plan, and executes the training process. 
It is built on the top of Megatron-LM~\cite{DBLP:journals/corr/abs-1909-08053} with TransformerEngine~\cite{transformerengine}, one of the most popular training frameworks. 
The runtime executor utilizes two rounding buffers for the storage of skeletal activations, as introduced in Section~\ref{subsec:tokenwise}.
%
%
Meanwhile, the transient activation tensors are (de)allocated according to the memory plan.

Three CUDA streams are employed for efficient overlapping of data transmission and GPU computation, which are for GPU computation, activation offloading from GPU to CPU, and activation prefetching from CPU to GPU, respectively. 
Figure~\ref{fig:method:scheduling} shows the scheduling of computation and transmission. 
After the computation of one layer's forward pass, the skeletal activations of this layer are scheduled to be transferred to the CPU memory, which can overlap with the computation of the next layer. 
Before the backward computation of one layer, the forward skeletal activations of the previous layer are scheduled to be fetched back to GPU. 
In addition, token-wise tensor recomputation is scheduled before the layer's backward pass.
By hiding the activation swapping with computation and enabling the lightweight, token-wise activation recomputation, \name minimizes the overhead of activation rematerialization at full stretch. 

\begin{table}[!tb]
\footnotesize
\caption{Configurations of the evaluated models.} 
\label{tab:expr:modelsize}

\newcommand{\cwidth}[0]{0.8cm}
\begin{tabular}{c|p{\cwidth}<{\centering}p{\cwidth}<{\centering}p{\cwidth}<{\centering}p{\cwidth}<{\centering}p{\cwidth}<{\centering}}
\toprule[1pt]
  &  \multicolumn{5}{c}{\textbf{Hyper Parameters}}  \\
\multirow{-2}{*}{\textbf{Model Size}} & \textbf{$n_{layers}$} & \textbf{$h$} & \textbf{$h_{ffn}$}  & \textbf{$n_{head}$} &\textbf{$n_{vocab}$}  \\
\midrule[0.8pt]
7B &32 &4096 &16384 &32 &50257  \\ 
13B &40 &5120 &20480 &40 &50257 \\ 
30B &48 &7168 &28672 &56 &50257  \\ 
65B &80 &8192 &32768 &64 &50257  \\ 
\bottomrule[1pt]
\end{tabular}

\end{table}

\section{Experiments}

In this section, we conduct experiments across various model sizes and input sequence lengths to show that \name achieves superior efficiency in long context training of LLMs.

\subsection{Setup}

\noindent\textbf{Hardware:}
Our experiments are conducted on an A800 GPU cluster, with each node equipped with 8 NVIDIA A800 GPUs (80GB). 
The GPUs within each node are interconnected via NVLinks (400GB/s), while the nodes are interconnected through Infiniband (200GB/s). 
Each node has 2TB CPU memory, and the GPU-CPU communication bandwidth is 32GB/s. 
%

\noindent\textbf{Baselines:}
We select two widely-used LLM training frameworks as baselines. 
The first is Megatron-LM (commit id: ccfeda47cb)~\cite{DBLP:journals/corr/abs-1909-08053} in conjunction with TransformerEngine (v1.3)~\cite{transformerengine}. 
Megatron-LM, maintained by NVIDIA, is renowned for its comprehensive support of hybrid parallelisms, including DP, TP, PP, SP, and CP.
The other baseline is Megatron-Deepspeed (commit id 7eb36a11b3) paired with DeepSpeed (v0.14.3)~\cite{DBLP:conf/kdd/RasleyRRH20}, which is recognized for ZeRO optimizers and DeepSpeed-Ulysses~\cite{DBLP:journals/corr/abs-2309-14509}, a novel parallel training strategy designed for long context LLM training. 
%
For all baselines, we employ FlashAttention~\cite{flash1, flash2} and mixed-precision training~\cite{mixed_precision, bf16_mixed_precition}, which are commonly used in LLM training.

\noindent\textbf{Metrics:}
\label{exp:metric}
We use three important evaluation metrics to measure the training efficiency: Model FLOPs Utilization (MFU), Tokens per GPU per Second (TGS), \add{and wall-clock time}.
MFU is defined as the ratio of model FLOPs per second to the theoretical peak FLOPs per second of the GPU (e.g. 312 TFLOPS for NVIDIA A800 GPUs)~\cite{DBLP:journals/corr/abs-2205-05198}. Based on FlashAttention~\cite{flash1,flash2} and considering the causal mask, the exact formula for calculating model FLOPS per sample is:
$$     6 \cdot s \cdot P + 6 \cdot n  \cdot h \cdot s^{2} .$$
%
MFU is a standard metric that measures the training efficiency of how model FLOPs utilize computational resources.
On the other hand, TGS directly measures training throughput, \add{and wall-clock time reflects the latency of training each batch}, providing a clear view of how quickly a model can be trained using a given amount of training samples.
All metrics are crucial for LLM researchers and engineers, enabling comparisons among various training strategies (including distributed parallelisms and activation recomputation).

\noindent
\textbf{Workloads:}
Our experiments cover a wide range of workloads to examine the strength of \name. 
In particular, we consider training the 7B, 13B, 30B and 65B GPT models on 8, 16, 32, and 64 GPUs, respectively, with various sequence lengths ranging from \add{4K} to 1408K. 
The detailed model configurations are shown in Table~\ref{tab:expr:modelsize}. 

\begin{table}[ht]

\setlength{\arrayrulewidth}{1pt}
\newcommand{\rank}[1]{\scriptsize{\uline{#1}}}
\newcommand{\allline}[0]{\hhline{|-|-|-|-|-|-|-|-|-|-|-|-|-|}}
\newcommand{\leaveone}[0]{\hhline{|~|-|-|-|-|-|-|-|-|-|-|-|-|}}
\newcommand{\leavetwo}[0]{}
\newcommand{\cwidth}[0]{0.68cm}
\newcommand{\cwidthfirst}[0]{0.66cm}
\scriptsize
\caption{MFU (Model FLOPS Utilization), TGS (Tokens per GPU per Second), and wall-clock time of different methods on different numbers of GPUs, model sizes, and sequence lengths. ``DS'' stands for ``DeepSpeed'', and ``Mega'' stands for ``Megatron-LM''. \XSolidBrush$_{oom}$ stands for Out Of Memory error when the system runs out of GPU memory, and \XSolidBrush$_{oohm}$ stands for Out Of Host Memory error, which means the CPU memory is depleted. ``h'', ``m'' and ``s'' represent hour, minute and second respectively.} 
\label{tab:expr:mfu}

\renewcommand{\arraystretch}{1.045}

\begin{tabular}{|p{\cwidthfirst}<{\centering}|p{\cwidth}<{\centering}p{\cwidth}<{\centering}p{\cwidth}<{\centering}|p{\cwidth}<{\centering}p{\cwidth}<{\centering}p{\cwidth}<{\centering}|p{\cwidth}<{\centering}p{\cwidth}<{\centering}p{\cwidth}<{\centering}|p{\cwidth}<{\centering}p{\cwidth}<{\centering}p{\cwidth}<{\centering}|}
\allline
 &  \multicolumn{3}{c|}{\textbf{7B GPT on 8 GPUs}} & \multicolumn{3}{c|}{\textbf{13B GPT on 16 GPUs}} & \multicolumn{3}{c|}{\textbf{30B GPT on 32 GPUs}} & \multicolumn{3}{c|}{\textbf{65B GPT on 64 GPUs}} \\
\leaveone
\multirow{-2}{*}{\textbf{SeqLen}} & \textbf{DS} & \textbf{Mega} & \textbf{\name} & \textbf{DS} & \textbf{Mega} & \textbf{\name} & \textbf{DS} & \textbf{Mega} & \textbf{\name} & \textbf{DS} & \textbf{Mega} & \textbf{\name} \\
\allline
 &  44.71\% & \textbf{49.45\%} & \textbf{49.45\%} & 44.08\% & \textbf{46.17\%} & \textbf{46.17\%} & 43.70\% & \textbf{50.67\%} & \textbf{50.67\%} & 43.89\% & \textbf{53.61\%} & \textbf{53.61\%} \\
\leavetwo
 &  3235.64 & \textbf{3578.86} & \textbf{3578.86} & 1675.53 & \textbf{1755.23} & \textbf{1755.23} & 724.44 & \textbf{840.14} & \textbf{840.14} & 337.95 & \textbf{412.90} & \textbf{412.90} \\
\leavetwo
\multirow{-3}{*}{4K} & 2.53s  & \textbf{2.29s} & \textbf{2.29s} & 2.44s & \textbf{2.33s} & \textbf{2.33s} & 2.83s & \textbf{2.44s} & \textbf{2.44s} & 3.03s & \textbf{2.48s} & \textbf{2.48s} \\

\allline
 & 46.28\% & \textbf{51.57\%} & \textbf{51.57\%}  & 38.90\% & \textbf{50.91\%} & \textbf{50.91\%} & 38.78\% & \textbf{49.27\%} & \textbf{49.27\%} & 43.34\% & \textbf{53.27\%} & \textbf{53.27\%} \\
\leavetwo
 & 3116.25 & \textbf{3472.95} & \textbf{3472.95} & 1393.24 & \textbf{1823.52} & \textbf{1823.52} & 615.22 & \textbf{781.68} & \textbf{781.68} & 321.00 & \textbf{394.61} & \textbf{394.61} \\
\leavetwo
\multirow{-3}{*}{8K} & 5.26s  &  \textbf{4.72s}  & \textbf{4.72s} & 5.88s & \textbf{4.50s} & \textbf{4.50s} & 6.66s & \textbf{5.24s} & \textbf{5.24s} & 6.38s & \textbf{5.19s} & \textbf{5.19s} \\

\allline
 & 36.41\% &  \textbf{51.29\%} & \textbf{51.29\%} & 36.95\% & \textbf{51.52\%} & \textbf{51.52\%} & 34.54\% & \textbf{49.84\%} & \textbf{49.84\%} & 40.98\% & \textbf{51.10\%} & \textbf{51.10\%} \\
\leavetwo
 & 2152.65 & \textbf{3032.14} & \textbf{3032.14} & 1186.24 & \textbf{1654.11} & \textbf{1654.11} & 504.64 & \textbf{728.22} & \textbf{728.22} & 282.05 & \textbf{351.58} & \textbf{351.58} \\
\leavetwo
\multirow{-3}{*}{16K} & 15.2s  &  \textbf{10.8s}  & \textbf{10.8s} & 13.8s & \textbf{9.91s} & \textbf{9.91s} & 16.2s & \textbf{11.2s} & \textbf{11.2s} & 14.5s & \textbf{11.6s} & \textbf{11.6s}\\

\allline
 & 35.91\% &  \textbf{52.75\%} & \textbf{52.75\%} & 34.51\% & \textbf{49.79\%} & \textbf{49.79\%} & 35.19\% & \textbf{48.06\%} & \textbf{48.06\%} & 35.73\% & 34.00\% & \textbf{44.79\%} \\
\leavetwo
 & 1706.13 & \textbf{2506.39} & \textbf{2506.39} & 917.96 & \textbf{1324.27} & \textbf{1324.27} & 443.77 & \textbf{606.16} & \textbf{606.16} & 215.32 & 204.83 & \textbf{269.84} \\
\leavetwo
\multirow{-3}{*}{32K} & 38.4s  &  \textbf{26.1s}  & \textbf{26.1s} & 35.7s & \textbf{24.7s} & \textbf{24.7s} & 36.9s & \textbf{27.0s} & \textbf{27.0s} & 38.0s & 39.9s & \textbf{30.4s} \\

\allline
 & 27.95\% & 41.55\% & \textbf{52.34\%} & 27.97\% & 38.51\% & \textbf{52.65\%} & 29.93\% & 35.76\% & \textbf{52.12\%} & 31.05\% & 22.79\% & \textbf{47.80\%} \\
\leavetwo
 & 935.76 & 1417.74 & \textbf{1786.22} & 553.85 & 762.46 & \textbf{1042.50} & 296.41 & 354.20 & \textbf{516.16} & 149.80 & 109.94 & \textbf{230.62} \\
\leavetwo
\multirow{-3}{*}{64K} & 2m17s  & 1m33s  & \textbf{1m13s} & 1m58s & 1m26s & \textbf{1m3s} & 1m51s & 1m33s & \textbf{1m3s} & 1m49s & 2m19s & \textbf{1m11s} \\

\allline
 & 25.46\% & 24.13\% & \textbf{50.96\%} & 25.45\% & 23.02\% & \textbf{50.93\%} & 25.54\% & 14.70\% & \textbf{49.66\%} & 26.13\% & 15.10\% & \textbf{48.61\%} \\
\leavetwo
 & 555.51  & 526.62 & \textbf{1111.99} & 333.51 & 301.67 & \textbf{667.41} & 176.92 & 101.85 & \textbf{344.09} & 90.15 & 52.10 & \textbf{167.69} \\
\leavetwo
\multirow{-3}{*}{128K} & 7m52s  &  8m18s  & \textbf{3m56s} & 6m33s & 7m15s & \textbf{3m16s} & 6m10s & 10m43s & \textbf{3m10s} & 6m4s & 10m30s & \textbf{3m15s} \\

\allline
 & 23.38\% & 29.07\% & \textbf{53.62\%} & 21.98\% & 25.30\% & \textbf{51.22\%} & \XSolidBrush$_{oom}$  & 17.15\% & \textbf{50.00\%} & 22.07\% & 9.57\% & \textbf{49.87\%} \\
\leavetwo
 & 296.48 & 368.65 & \textbf{679.92} & 171.82 & 197.78 & \textbf{400.39} & \XSolidBrush$_{oom}$  & 72.24 & \textbf{216.41} & 48.51 & 21.03 & \textbf{109.58} \\
\leavetwo
\multirow{-3}{*}{256K} & 29m28s  &  23m42s  & \textbf{12m51s} & 25m26s & 22m5s & \textbf{10m55s} & \XSolidBrush$_{oom}$  & 29m30s & \textbf{10m6s} & 22m31s & 51m56s & \textbf{9m58s} \\

\allline
 & \XSolidBrush$_{oom}$ & 27.98\% & \textbf{53.04\%} & \XSolidBrush$_{oom}$  & 22.88\% & \textbf{51.91\%} & \XSolidBrush$_{oom}$  & 23.32\% & \textbf{50.69\%} & 20.40\% & 12.07\% & \textbf{48.85\%} \\
\leavetwo
 & \XSolidBrush$_{oom}$ &  250.07 & \textbf{474.02} & \XSolidBrush$_{oom}$  & 127.40 & \textbf{289.11} & \XSolidBrush$_{oom}$  & 73.37 & \textbf{159.52} & 32.89 & 19.46 & \textbf{78.76} \\
\leavetwo
\multirow{-3}{*}{384K} & \XSolidBrush$_{oom}$  & 52m25s  & \textbf{27m39s} & \XSolidBrush$_{oom}$  & 51m26s & \textbf{22m40s} & \XSolidBrush$_{oom}$  & 44m39s & \textbf{20m32s} & 49m49s & 1h24m & \textbf{20m48s} \\

\allline
 & \XSolidBrush$_{oom}$ & 34.43\% & \textbf{51.84\%} & \XSolidBrush$_{oom}$  & 29.10\% & \textbf{52.40\%} & \XSolidBrush$_{oom}$  & \XSolidBrush$_{oom}$  & \textbf{51.06\%} & 19.83\% & 5.32\% & \textbf{49.71\%} \\
\leavetwo
 & \XSolidBrush$_{oom}$ &  237.56 & \textbf{357.70} & \XSolidBrush$_{oom}$  & 125.86 & \textbf{226.65} & \XSolidBrush$_{oom}$  & \XSolidBrush$_{oom}$  & \textbf{126.23} & 25.24 & 6.77 & \textbf{63.29} \\
\leavetwo
\multirow{-3}{*}{512K} & \XSolidBrush$_{oom}$  & 1h14m  & \textbf{48m51s} & \XSolidBrush$_{oom}$  & 1h10m & \textbf{38m33s} & \XSolidBrush$_{oom}$  & \XSolidBrush$_{oom}$  & \textbf{34m37s} & 1h27m & 5h23m & \textbf{34m31s} \\

\allline
 & \XSolidBrush$_{oom}$ & 30.90\% & \textbf{52.59\%} & \XSolidBrush$_{oom}$  & 19.41\% & \textbf{52.13\%} & \XSolidBrush$_{oom}$  & \XSolidBrush$_{oom}$  & \textbf{51.72\%} & 19.06\% & \XSolidBrush$_{oom}$  &  \textbf{50.05\%} \\
\leavetwo
 & \XSolidBrush$_{oom}$ &  173.63 & \textbf{295.51} & \XSolidBrush$_{oom}$  & 83.94 & \textbf{184.33} & \XSolidBrush$_{oom}$  & \XSolidBrush$_{oom}$  & \textbf{105.28} & 20.04 & \XSolidBrush$_{oom}$  & \textbf{52.65} \\
\leavetwo
\multirow{-3}{*}{640K} & \XSolidBrush$_{oom}$  &  2h6m  & \textbf{1h14m} & \XSolidBrush$_{oom}$  & 2h10s & \textbf{59m15s} & \XSolidBrush$_{oom}$  & \XSolidBrush$_{oom}$  & \textbf{51m52s} & 2h16m & \XSolidBrush$_{oom}$  & \textbf{51m52s} \\

\allline
 & \XSolidBrush$_{oom}$ & \XSolidBrush$_{oom}$ & \textbf{51.89\%} & \XSolidBrush$_{oom}$ & \XSolidBrush$_{oom}$ & \textbf{51.71\%} & \XSolidBrush$_{oom}$ & \XSolidBrush$_{oom}$ & \textbf{51.18\%} & 19.53\% & \XSolidBrush$_{oom}$ & \textbf{51.16\%} \\
\leavetwo
 & \XSolidBrush$_{oom}$ & \XSolidBrush$_{oom}$ & \textbf{245.76} & \XSolidBrush$_{oom}$ & \XSolidBrush$_{oom}$ & \textbf{154.63} & \XSolidBrush$_{oom}$ & \XSolidBrush$_{oom}$ & \textbf{88.55} & 17.50 & \XSolidBrush$_{oom}$ & \textbf{45.84} \\
\leavetwo
\multirow{-3}{*}{768K} & \XSolidBrush$_{oom}$  & \XSolidBrush$_{oom}$  & \textbf{1h47m} & \XSolidBrush$_{oom}$ & \XSolidBrush$_{oom}$ & \textbf{1h25m} & \XSolidBrush$_{oom}$ & \XSolidBrush$_{oom}$ & \textbf{1h14m} & 3h7m & \XSolidBrush$_{oom}$ & \textbf{1h11m} \\

\allline
 & \XSolidBrush$_{oom}$ & \XSolidBrush$_{oom}$ & \textbf{52.71\%} & \XSolidBrush$_{oom}$ & \XSolidBrush$_{oom}$ & \textbf{51.76\%} & \XSolidBrush$_{oom}$ & \XSolidBrush$_{oom}$ & \textbf{51.50\%} & 19.12\% & \XSolidBrush$_{oom}$ & \textbf{51.05\%} \\
\leavetwo
 & \XSolidBrush$_{oom}$ &\XSolidBrush$_{oom}$ & \textbf{251.98} & \XSolidBrush$_{oom}$ & \XSolidBrush$_{oom}$ & \textbf{134.08} & \XSolidBrush$_{oom}$ & \XSolidBrush$_{oom}$ & \textbf{77.48} & 14.93 & \XSolidBrush$_{oom}$ & \textbf{39.85} \\
\leavetwo
\multirow{-3}{*}{896K} & \XSolidBrush$_{oom}$  & \XSolidBrush$_{oom}$  & \textbf{2h22m} & \XSolidBrush$_{oom}$ & \XSolidBrush$_{oom}$ & \textbf{1h54m} & \XSolidBrush$_{oom}$ & \XSolidBrush$_{oom}$ & \textbf{1h39m} & 4h16m & \XSolidBrush$_{oom}$ & \textbf{1h36m} \\

\allline
 & \XSolidBrush$_{oom}$ & \XSolidBrush$_{oom}$ & \textbf{52.30\%} & \XSolidBrush$_{oom}$ & \XSolidBrush$_{oom}$ & \textbf{52.06\%} & \XSolidBrush$_{oom}$ & \XSolidBrush$_{oom}$ & \textbf{51.24\%} & 19.00\% & \XSolidBrush$_{oom}$ & \textbf{51.27\%} \\
\leavetwo
 & \XSolidBrush$_{oom}$ & \XSolidBrush$_{oom}$ & \textbf{188.73} & \XSolidBrush$_{oom}$ & \XSolidBrush$_{oom}$ & \textbf{118.96} & \XSolidBrush$_{oom}$ & \XSolidBrush$_{oom}$ & \textbf{68.20} & 13.14 & \XSolidBrush$_{oom}$ & \textbf{35.45} \\
\leavetwo
\multirow{-3}{*}{1024K} & \XSolidBrush$_{oom}$  & \XSolidBrush$_{oom}$  & \textbf{3h5m} & \XSolidBrush$_{oom}$ & \XSolidBrush$_{oom}$ & \textbf{2h27m} & \XSolidBrush$_{oom}$ & \XSolidBrush$_{oom}$ & \textbf{2h8m} & 5h33m & \XSolidBrush$_{oom}$ & \textbf{2h3m} \\

\allline
 & \XSolidBrush$_{oom}$ & \XSolidBrush$_{oom}$ & \XSolidBrush$_{oom}$ & \XSolidBrush$_{oom}$ & \XSolidBrush$_{oom}$ & \textbf{51.74\%} & \XSolidBrush$_{oom}$ & \XSolidBrush$_{oom}$ & \textbf{51.73\%} & 19.11\% & \XSolidBrush$_{oom}$ & \textbf{51.20\%} \\
\leavetwo
 & \XSolidBrush$_{oom}$ & \XSolidBrush$_{oom}$ & \XSolidBrush$_{oom}$ & \XSolidBrush$_{oom}$ & \XSolidBrush$_{oom}$ & \textbf{105.74} & \XSolidBrush$_{oom}$ & \XSolidBrush$_{oom}$ & \textbf{61.72} & 11.86 & \XSolidBrush$_{oom}$ & \textbf{31.77} \\
\leavetwo
\multirow{-3}{*}{1152K} & \XSolidBrush$_{oom}$  & \XSolidBrush$_{oom}$  & \XSolidBrush$_{oom}$ &  \XSolidBrush$_{oom}$& \XSolidBrush$_{oom}$ &\textbf{3h6m}  & \XSolidBrush$_{oom}$ & \XSolidBrush$_{oom}$ & \textbf{2h39m} & 6h43m & \XSolidBrush$_{oom}$ & \textbf{2h35m} \\

\allline
 & \XSolidBrush$_{oom}$ &  \XSolidBrush$_{oom}$ & \XSolidBrush$_{oom}$ & \XSolidBrush$_{oom}$ & \XSolidBrush$_{oom}$ & \textbf{51.78\%} & \XSolidBrush$_{oom}$ & \XSolidBrush$_{oom}$ & \textbf{51.59\%} & 18.90\% & \XSolidBrush$_{oom}$ & \textbf{51.42\%} \\
\leavetwo
 & \XSolidBrush$_{oom}$ &  \XSolidBrush$_{oom}$ & \XSolidBrush$_{oom}$ & \XSolidBrush$_{oom}$ & \XSolidBrush$_{oom}$ & \textbf{95.72} & \XSolidBrush$_{oom}$ & \XSolidBrush$_{oom}$ & \textbf{55.79} & 10.64 & \XSolidBrush$_{oom}$ & \textbf{28.94}\\
\leavetwo
\multirow{-3}{*}{1280K} & \XSolidBrush$_{oom}$  & \XSolidBrush$_{oom}$  & \XSolidBrush$_{oom}$ &  \XSolidBrush$_{oom}$& \XSolidBrush$_{oom}$ & \textbf{3h48m} &  \XSolidBrush$_{oom}$& \XSolidBrush$_{oom}$ & \textbf{3h16m} & 8h33m & \XSolidBrush$_{oom}$ & \textbf{3h9m}  \\

\allline
 & \XSolidBrush$_{oom}$ &  \XSolidBrush$_{oom}$ &  \XSolidBrush$_{oom}$& \XSolidBrush$_{oom}$ & \XSolidBrush$_{oom}$ & \textbf{52.10\%} & \XSolidBrush$_{oom}$ & \XSolidBrush$_{oom}$ & \XSolidBrush$_{oohm}$ & \XSolidBrush$_{oom}$ & \XSolidBrush$_{oom}$ & \textbf{51.45\%}\\
\leavetwo
 & \XSolidBrush$_{oom}$ &  \XSolidBrush$_{oom}$ &  \XSolidBrush$_{oom}$& \XSolidBrush$_{oom}$ & \XSolidBrush$_{oom}$ & \textbf{87.93} & \XSolidBrush$_{oom}$ & \XSolidBrush$_{oom}$ & \XSolidBrush$_{oohm}$ & \XSolidBrush$_{oom}$ & \XSolidBrush$_{oom}$ & \textbf{26.50}\\
\leavetwo
\multirow{-3}{*}{1408K}  & \XSolidBrush$_{oom}$ &  \XSolidBrush$_{oom}$ &  \XSolidBrush$_{oom}$& \XSolidBrush$_{oom}$ & \XSolidBrush$_{oom}$ & \textbf{4h33m} & \XSolidBrush$_{oom}$ & \XSolidBrush$_{oom}$ & \XSolidBrush$_{oohm}$ & \XSolidBrush$_{oom}$ & \XSolidBrush$_{oom}$ & \textbf{3h47m}\\

\allline
\end {tabular}
\end {table}

\subsection{End-to-end Evaluation}
We compare the end-to-end training efficiency of \name and two baselines. 
%
%
Table~\ref{tab:expr:mfu} shows the MFU, TGS, \add{and wall-clock time} of DeepSpeed-Ulysses, Megatron-LM and \name under different training workloads. 
During evaluation, we manually adjust the distributed parallelism strategies for each system and each workload to achieve optimal training performance for fair comparisons. 
%

%
Overall, \name is capable of training longer sequences than the competitors. 
Across the training of 7B, 13B, 30B, and 65B models on 8, 16, 32, and 64 GPUs, DeepSpeed-Ulysses supports sequence lengths of 256K, 256K, 128K, and 1280K, while Megatron-LM supports sequence lengths of 640K, 640K, 384K, 512K.
In comparison, \name achieves superior performance in all scenarios, enabling training sequence lengths of 1024K, 1408K, 1280K, and 1408K.
Megatron-LM only supports up to 640K sequence length, even if we have leveraged a high model parallel degree and enabled the memory reduction techniques. 
%
This is unsurprising since it overlooks the memory fragmentation issue, leading to OOM for large sequence lengths.
DeepSpeed, thanks to its support of DeepSpeed-Ulysses sequence parallel and ZeRO-3 optimizer, is capable of training 1280K sequence length when training the 65B model on 64 GPUs.
%
When training smaller models, DeepSpeed supports only very small sequence lengths. 
This is because it can only utilize a small SP degree of 8, which either aligns to the number of GPUs or is dividable by the number of attention heads (40 and 56).
In contrast, by token-wise recomputation/swapping and memory planning, \name is able to train sequences over 1 million tokens in all scenarios.

\add{When training with relatively short sequences ranging from 4K to 32K, there's little memory pressure. 
In such cases, activation recomputation is often unnecessary, and different methods may use the same parallelism strategies.
Consequently, \name falls back to Megatron-LM, resulting in the same performance.
Nevertheless, \name significantly outperforms the baselines when the sequence length reaches 64K or even higher.}

Furthermore, when comparing \name to the baselines with aligned sequence lengths, \name achieves superior MFU, TGS, \add{and wall-clock time}.
Across all experimented workloads, \name achieves an average MFU of \add{51.04\%}. 
In contrast, Megatron-LM and DeepSpeed only achieve an average MFU of \add{35.01\%} and \add{30.74\%}, respectively. 
On average, \name achieves \add{$1.97\times$} and \add{$1.80\times$} MFU compared to Megatron-LM and DeepSpeed, respectively.
%

The deficiencies of baselines are not surprising --- due to the unsatisfactory memory management, 
%
\add{they are forced to utilize less-efficient configurations of parallelism strategies and rematerialization options.\footnote{\add{We have provided the detailed configurations in our supplementary materials.}} 
Firstly, the baselines need to employ a high model parallelism degree (i.e. large TP degree and/or large SP degree) to avoid OOM error. 
For instance, when training the 65B model with 256K sequence length, Megatron-LM has to use a TP degree of 16, while \name can support a TP degree of 8. 
Given the much lower inter-node bandwidth compared to intra-node, the TP communication cost of \name is 63\% lower than that of Megatron-LM (44.3 v.s. 120.4 seconds).
Secondly, the baselines require full activation recomputation to avoid the OOM error, leading to over 20\% redundant GPU calculation. 
In contrast, \name integrates fine-grained recomputation and swapping, incurring smaller overhead.
Last but not least, the lack of memory planning often triggers memory reorganization during training, which blocks GPU computation and significantly hampers training efficiency.}
When training the 7B model on 8 GPUs using Megatron-LM, the memory reorganization operation is triggered 6 times and 16 times per iteration for sequence lengths of 128K and 256K, respectively.
\add{A detailed ablation study is presented in Section~\ref{subsec:ablation}}.

As a result, \name consistently achieves an MFU of approximately 50\% across all model sizes and sequence lengths, enabling more efficient training of significantly long sequences compared to the baselines.

\begin{table*}[!t]
\scriptsize

\caption{MFU of different methods for ablation studies. The experiments are conducted by training the 7B model on 8 GPUs.}
\label{tab:expr:ablation}
\newcommand{\cwidth}[0]{0.7cm}
\begin{tabular}{c|p{\cwidth}<{\centering}p{\cwidth}<{\centering}p{\cwidth}<{\centering}p{\cwidth}<{\centering}p{\cwidth}<{\centering}p{\cwidth}<{\centering}p{\cwidth}<{\centering}p{\cwidth}<{\centering}}
\toprule[1pt]
  &  \multicolumn{8}{c}{\textbf{Sequence Length}}  \\
\multirow{-2}{*}{\textbf{Method}} & \textbf{64K} & \textbf{128K} & \textbf{256K} & \textbf{384K} & \textbf{512K} & \textbf{640K} & \textbf{768K} & \textbf{896K} \\
\midrule[0.8pt]
Full Recomputation &41.19\% &23.00\% &29.07\% &25.67\% &\XSolidBrush$_{oom}$ &\XSolidBrush$_{oom}$ &\XSolidBrush$_{oom}$ &\XSolidBrush$_{oom}$ \\ 
Full Recomputation + Memory Plan &42.91\% &43.17\% &42.05\% &42.49\% &41.90\% &42.15\% &\XSolidBrush$_{oom}$ &\XSolidBrush$_{oom}$ \\ 
Full Swapping + Memory Plan &37.40\% &46.33\% &53.62\% &\XSolidBrush$_{oohm}$ &\XSolidBrush$_{oohm}$ &\XSolidBrush$_{oohm}$ &\XSolidBrush$_{oohm}$ &\XSolidBrush$_{oohm}$ \\ 
\name 
(Fine-grained Management + Memory Plan) 
&\textbf{47.99\%} &\textbf{50.96\%} &\textbf{53.62\%} &\textbf{53.04\%} &\textbf{51.84\%} &\textbf{52.59\%} &\textbf{51.89\%} &\textbf{52.71\%} \\ 
\bottomrule[1pt]
\end{tabular}

\end{table*}

\begin{table*}[!t]
\scriptsize
\caption{\add{Impact of offloading fraction $\alpha$ on training efficiency.}}
\label{tab:expr:alpha}

\newcommand{\cwidth}[0]{0.8cm}
\begin{tabular}{c|p{\cwidth}<{\centering}p{\cwidth}<{\centering}p{\cwidth}<{\centering}p{\cwidth}<{\centering}p{\cwidth}<{\centering}p{\cwidth}<{\centering}p{\cwidth}<{\centering}p{\cwidth}<{\centering}p{\cwidth}<{\centering}}
\toprule[1pt]
  &  \multicolumn{8}{c}{\add{\textbf{Value of $\alpha$}}}  \\
\multirow{-2}{*}{\textbf{\add{Sequence Length}}} & \add{\textbf{0.000}} & \add{\textbf{0.125}} & \add{\textbf{0.250}} & \add{\textbf{0.375}} & \add{\textbf{0.500}} & \add{\textbf{0.625}} & \add{\textbf{0.750}} & \add{\textbf{0.875}} & \add{\textbf{1.000}} \\ 
\midrule[0.8pt]
\add{192K} &\add{51.04\%} & \add{51.86\%} &\add{52.14\%} &\add{52.30\%} &\add{52.72\%} &\add{52.78\%} &\add{53.11\%} &\add{51.74\%} &\add{49.69\%} \\ 
\add{256K} &\add{52.03\%} & \add{52.05\%} &\add{52.19\%} &\add{52.45\%} &\add{52.70\%} &\add{52.96\%} &\add{53.21\%} &\add{53.29\%} &\add{53.62\%} \\
\add{320K} &\add{52.15\%} &\add{52.36\%} &\add{52.52\%} &\add{52.66\%} &\add{52.92\%} &\add{53.10\%} &\add{53.35\%} &\add{\XSolidBrush$_{oohm}$} &\add{\XSolidBrush$_{oohm}$}  \\
\add{384K} &\add{52.09\%} &\add{52.31\%} &\add{52.45\%} &\add{52.58\%} &\add{53.04\%} &\add{\XSolidBrush$_{oohm}$} &\add{\XSolidBrush$_{oohm}$} &\add{\XSolidBrush$_{oohm}$} &\add{\XSolidBrush$_{oohm}$}  \\ 
\bottomrule[1pt]
\end{tabular}

\end{table*}

\subsection{Ablation Studies}
\label{subsec:ablation}
Next, we assess the effectiveness of the proposed techniques in \name. All experiments in ablation studies are conducted by training the 7B model on 8 GPUs, keeping the parallelism configuration fixed at a TP degree of 4 and a CP degree of 2.

\subsubsection{\textbf{Effectiveness of Memory Planning}}
To evaluate the effectiveness of memory planning, we evaluate two variants of \name with full recomputation, both with and without memory planning.
As shown in the first two rows of Table~\ref{tab:expr:ablation}, without memory planning, the longest sequence supported is only 384K, achieving an MFU of 25.67\%. 
After applying memory planning, the longest supported sequence length increases to 640K, with an MFU of 42.15\%. 
The results are reasonable since full recomputation without memory planning has severe memory fragmentation, resulting in OOM errors in large sequence length scenarios.

Additionally, the frequent GPU memory reorganization process further impairs training efficiency.
By employing memory planning, the fragmentation issue can be eliminated, providing more memory for longer context training.
Getting rid of GPU memory reorganization, memory planning brings an average of $1.51\times$ MFU when facing the same context length.

\subsubsection{\textbf{Effectiveness of Token-wise Recomputation}}
For token-wise recomputation and swapping, we compare \name and its variants, one with full recomputation and another with full swapping.
The results are shown in the last three rows of Table~\ref{tab:expr:ablation}. 
%
%
When training with appropriate sequence length, which is 256k in this scenario, the computation time of one transformer layer can fully overlap with the offloading time of a layer's activations. 
Therefore, full swapping with memory planning can achieve an MFU of 53.62\% under 256K sequence length, far exceeding the 42.05\% achieved by full recomputation with memory planning. 
However, for short sequence lengths, such as 64K, the offloading time of one layer's activations blocks the GPU computation,
%
resulting in a lower MFU than full recomputation. 
Full swapping presents another challenge as the sequence length grows longer: the host memory is rapidly depleted by offloaded activations, leading to OOHM errors.
\add{Full recomputation, which does not offload the input tensor of each layer, also encounters OOM errors at a sequence length of 768K.}

By employing token-wise recomputation together with swapping, \name consistently improves training efficiency for both short and long context lengths. 
For short sequence lengths like 64K, our tensor-level design only offloads the input tensor of the transformer layer and the FlashAttention output tensor to CPU memory, enabling efficient overlap of GPU computation and data transmission. 
For long context lengths, our rounding-buffer token-level management successfully avoids depleting the CPU memory, and incurs only minimal recomputation overhead. 
%
Among all the methods, \name supports the longest sequence length.
Considering MFU, \name achieves an average of $1.22\times$ MFU compared to full recomptutation with memory planning, and an average of $1.13\times$ MFU compared to full offloading with memory planning.

\subsubsection{\add{\textbf{Effect of the Offloading Fraction $\alpha$}}}
\add{To investigate the impact of the offloading fraction $\alpha$ on training efficiency, we train a 7B model on 8 GPUs, with TP$=4$, CP$=2$ and varying $\alpha$ values. 
The results are presented in Table~\ref{tab:expr:alpha}. 
For a sequence length of 192K, the MFU initially increases with $\alpha$, peaking at 53.11\% when $\alpha=0.75$, and then declines. 
This trend occurs because, for $\alpha < 0.75$, a higher $\alpha$ reduces recomputation overhead by increasing the proportion of skeletal activations that are swapped. 
Beyond this point, however, further increases in $\alpha$ lead to swapping overhead that exceeds the computation time of an individual transformer layer, stalling GPU computations until the swapping concludes and thus reducing the MFU.
At a sequence length of 256K, the computation time of a single transformer layer, due to the quadratic complexity of self-attention, surpasses the offloading time for all skeletal activations in this layer. 
Consequently, training efficiency consistently improves as $\alpha$ increases. 
For longer sequences such as 320K and 384K, offloading all skeletal activations becomes unviable due to CPU memory constraints. 
Considering the constraints, \name can effectively identify the optimal $\alpha$ to minimize recomputation overhead.}

\begin{figure}[tb]
    \centering
    \subfigure[Longest supported sequence length.]{
    \scalebox{0.23}{
    \includegraphics[width=\linewidth]{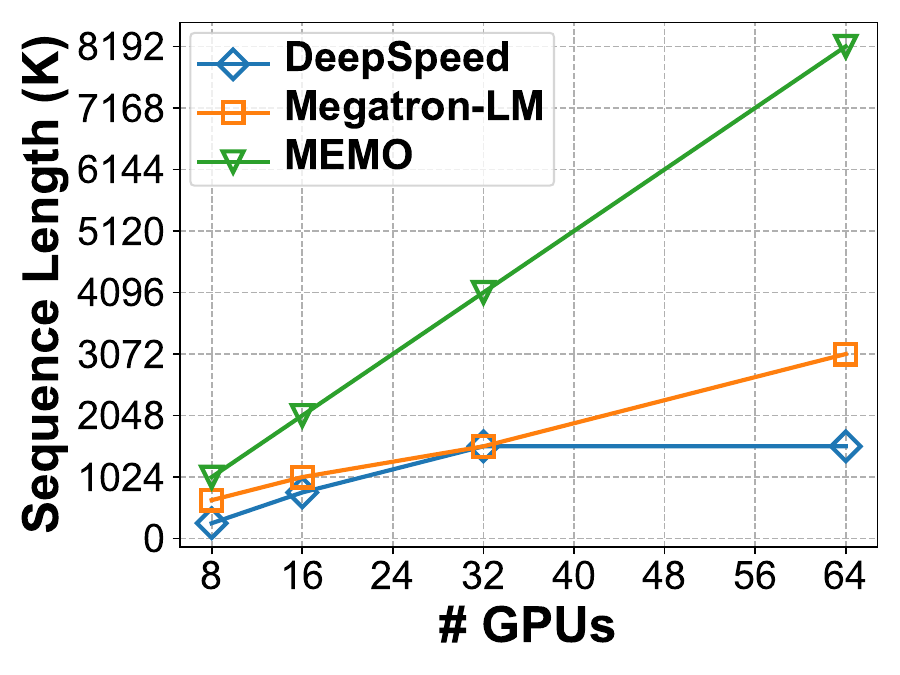}
    }\label{fig:expr:scalability:seqlen}
    }
    \subfigure[MFU at longest sequence length.]{
    \scalebox{0.23}{
    \includegraphics[width=\linewidth]{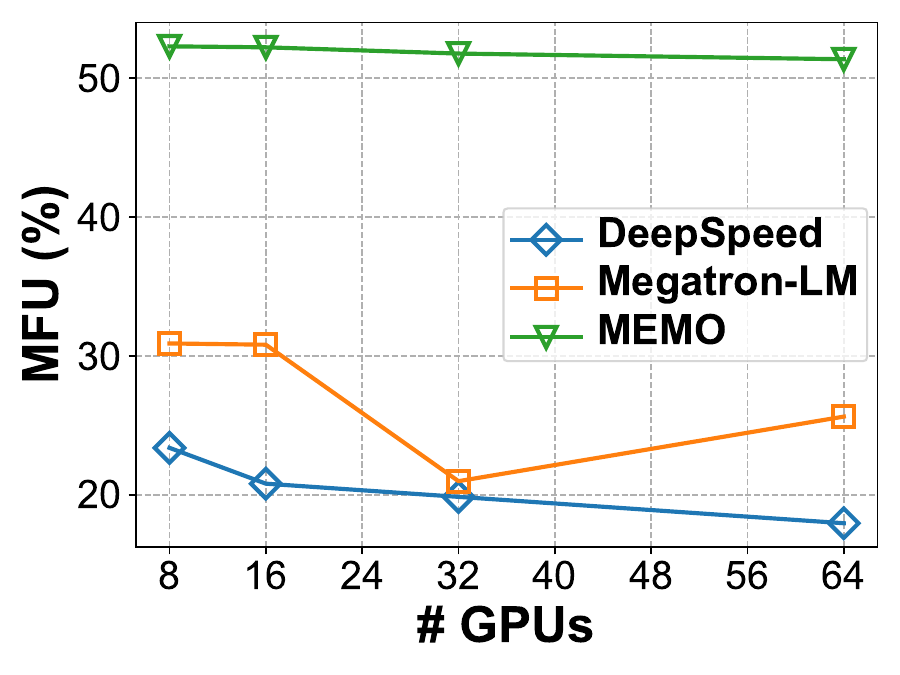}
    }\label{fig:expr:scalability:mfu}
    }
    \subfigure[MFU at varying sequence lengths.]{
    \scalebox{0.23}{
    \includegraphics[width=\linewidth]{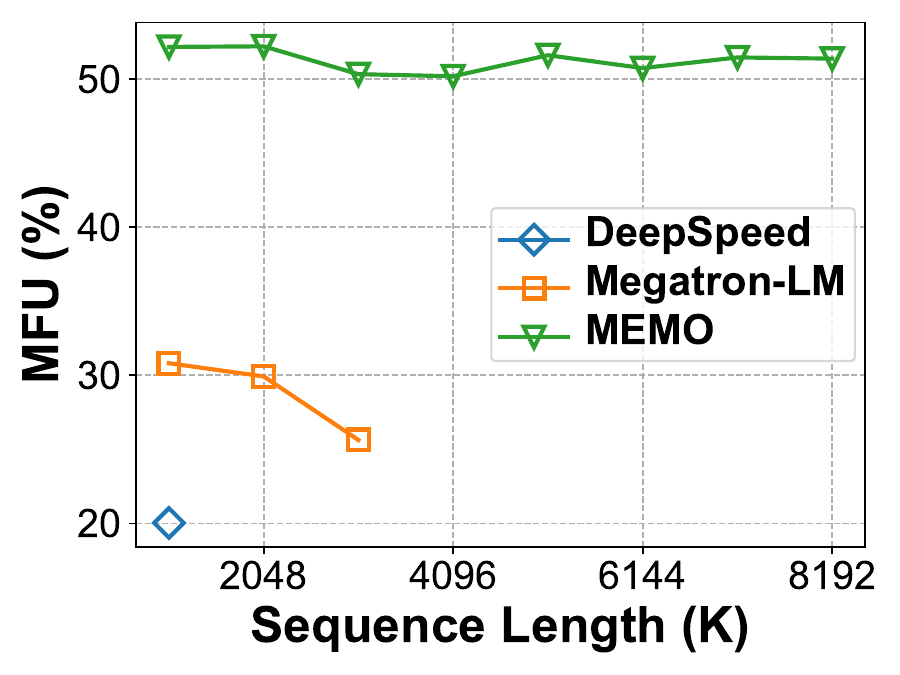}
    }\label{fig:exp:scale}
    }
    \subfigure[Training loss curves.]{
    \scalebox{0.23}{
    \includegraphics[width=\linewidth]{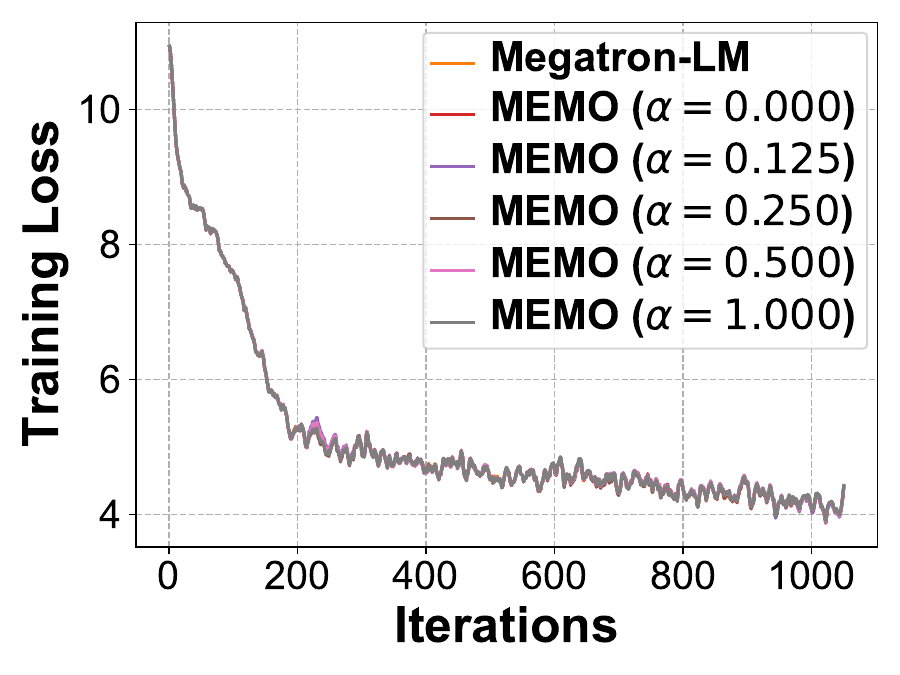}
    }\label{fig:exp:convergence}
    }
    \caption{(a), (b): The longest supported sequence length and corresponding MFU of DeepSpeed, Megatron-LM and \name when training the 7B model on various numbers of GPUs. (c): The MFU when training the 7B model on 64 GPUs with sequence length varying from 1024K to 8192K. (d): Training loss curves of Megatron-LM and \name with different $\alpha$.}
\end{figure}

\subsection{Scalability}

To show the scalability of \name, we train the 7B model on 8, 16, 32, and 64 GPUs respectively, and report the maximum supported sequence length.
As shown in Figure~\ref{fig:expr:scalability:seqlen}, when the number of GPUs increases, the maximum sequence length supported by \name increases linearly. 
When training on 8, 16, 32, and 64 GPUs, \name is capable of training 1, 2, 4, 8 million sequence lengths, respectively, which demonstrates ideal scalability. \name also consistently maintains an MFU of over 50\% across different numbers of GPUs, as shown in Figure~\ref{fig:expr:scalability:mfu}.

For DeepSpeed, as the number of GPUs increases, it can enlarge the SP degree, leading to longer supported sequence length. Note that since the 7B model has 32 attention heads, the maximum SP degree is 32. As a result, DeepSpeed achieves the same maximum sequence length of 1536K for both 32 and 64 GPUs. 
%
Megatron-LM supports context parallelism, which has better scalability. When the number of GPUs increases, the longest sequence length it can handle grows sublinearly.
Compared to the baselines, \name introduces fine-grained activation memory management, achieving not only ideal scalability but also better efficiency.

We also evaluate the MFU of the three systems when training the 7B model on 64 GPUs with sequence lengths varing from 1024K to 8192K. 
In Figure~\ref{fig:exp:scale}, as sequence length increases, the MFU of \name maintains above 50\%, surpassing the baselines.




\subsection{Convergence of \name}

To confirm the correctness of our system implementation, we conduct a convergence experiment. 
We train a 7B model with 128K sequence length on 8 GPUs and compare the convergence of \name and Megatron-LM for 1000 iterations. 
For both systems, we fix the parallelism strategy to TP$=4$ and CP$=2$. For \name, we enumerate the value of $\alpha$ in $\{0, 0.125, 0.25, 0.5 , 1\}$.
As shown in Figure~\ref{fig:exp:convergence}, the loss curves of \name with different $\alpha$ values all align with Megatron-LM, confirming the correctness of \name. 

\section{Related Work}

\noindent\textbf{\add{Data management and machine learning:}}
    \add{
        Recently, the intersection of data management and machine learning has gained popularity, with numerous studies emerging from the data management community~\cite{demystify, sdpipe, mics, fsdp, tfdata, fastflow, where, stall, het, orca, simple, galvatron,galvatron_2, tensorrelational, learningmodelsoverrelational, atensorcompiler, tensors, optimizetensorprogram, tod, dse1, dse2, jcst2, scis1, embedding_ea, cafe}. 
        %
        Some works focus on optimizing I/O data management to accelerate data pre-processing pipelines~\cite{tfdata, fastflow, where, stall}. 
        Another work proposes to use a fine-grained automatic parallelism strategy to speed up model training~\cite{galvatron}.
        \name also employs data management techniques and methodologies to enhance the efficiency of machine learning tasks, specifically addressing the challenges associated with long-sequence LLM training.
    }

\noindent\textbf{Parallelism strategies for long context training:}
To tackle the challenge of long context training, DeepSpeed-Ulysses~\cite{DBLP:journals/corr/abs-2309-14509} employs an novel AllToAll communication to facilitate the partition of input sequence among GPUs, achieving lower communication overhead compared with Megatron-LM sequence parallelism~\cite{DBLP:journals/corr/abs-2205-05198}. 
LightSeq~\cite{DBLP:journals/corr/abs-2310-03294}, Ring Attention~\cite{DBLP:journals/corr/abs-2310-01889}, and Megatron-LM context parallelism~\cite{contextparallel} propose to split the sequence within self-attention computation, achieving better scalability. 
Recent efforts in the realm of sequence and context parallelisms~\cite{DBLP:journals/corr/abs-2401-09149,DBLP:journals/corr/abs-2405-07719, hotspa} aim to integrate multiple strategies and enhance existing distributed settings.
It is worth noting that the fine-grained memory management of \name is orthogonal to these distributed parallelism strategies.

\noindent\textbf{Activation recomputation and swapping:}
Capuchin~\cite{peng2020capuchin} proposes to combine recomputation and swapping to reduce memory footprint during training. The swapping decision is made by considering the tensor access pattern.
In addition to tensor recomputation, MegTaichi~\cite{DBLP:conf/ics/HuXDLZZMST22} also proposes to co-optimize the tensor partition. 
Coop~\cite{DBLP:conf/nips/ZhangMLY23} notices that naive tensor recomputation leads to severe memory fragmentation, and proposes heuristics to reduce memory fragmentation during tensor recomputation. 
While these works offer solutions for common deep learning models, they do not take advantage of the specific characteristics of LLM training to achieve full overlapping and fragmentation minimization.

\noindent\textbf{Memory planning for deep learning models:} 
The memory allocation problem in deep learning models can be regarded as a DSA problem and solved by MIP~\cite{sekiyama2018profileguidedmemoryoptimizationdeep}.
OLLA~\cite{steiner2022ollaoptimizinglifetimelocation} proposes to optimize the lifetime and memory location of tensors during the training process by solving a joint ILP problem, reducing the peak memory during training. However, it does not exploit the repetitive substructure in LLMs and relies on heuristics to simplify the integer programming problem.

\section{Conclusion}

In this paper, we proposed \name to address the memory challenges in long context LLM training. 
We designed a fine-grained activation recomputation and swapping strategy to fully utilize the idle PCIe bandwidth during the GPU computation, thereby reducing the activation rematerialization cost in long context LLM training. 
We employed a bi-level MIP technique to first solve the problem of memory allocation within one transformer layer, and then reused the same memory space for each identical layer so as to eliminate memory fragmentation.
%
%
Through extensive experiments, we demonstrated that \name achieved an average of $1.97\times$ MFU compared to Megatron-LM.
By leveraging fine-grained tensor memory management, \name achieved 52.30\% MFU when training 7B LLM with 1 million sequence length on only 8 A800 GPUs.

%

\section*{Acknowledgement}

This work is supported by National Science and Technology Major Project (2022ZD0116315), National Natural Science Foundation of China (U22B2037, U23B2048, 62402011), Beijing Municipal Science and Technology Project (Z231100010323002), China National Postdoctoral Program for Innovative Talents (BX20230012), China Postdoctoral Science Foundation (2024M750103), Beijing Natural Science Foundation (4244080), research grant No. IPT-2024JK29, PKU-Tencent joint research Lab, and High-performance Computing Platform of Peking University. 
Fangcheng Fu and Bin Cui are the corresponding authors.

\bibliographystyle{ACM-Reference-Format}
\bibliography{sample-base}


\begin{thebibliography}{104}


\ifx \showCODEN    \undefined \def \showCODEN     #1{\unskip}     \fi
\ifx \showISBNx    \undefined \def \showISBNx     #1{\unskip}     \fi
\ifx \showISBNxiii \undefined \def \showISBNxiii  #1{\unskip}     \fi
\ifx \showISSN     \undefined \def \showISSN      #1{\unskip}     \fi
\ifx \showLCCN     \undefined \def \showLCCN      #1{\unskip}     \fi
\ifx \shownote     \undefined \def \shownote      #1{#1}          \fi
\ifx \showarticletitle \undefined \def \showarticletitle #1{#1}   \fi
\ifx \showURL      \undefined \def \showURL       {\relax}        \fi
\providecommand\bibfield[2]{#2}
\providecommand\bibinfo[2]{#2}
\providecommand\natexlab[1]{#1}
\providecommand\showeprint[2][]{arXiv:#2}

\bibitem[AI(2024)]%
        {kimichat}
\bibfield{author}{\bibinfo{person}{Moonshot AI}.} \bibinfo{year}{2024}\natexlab{}.
\newblock \bibinfo{title}{KimiChat}.
\newblock
\urldef\tempurl%
\url{https://kimi.moonshot.cn/}
\showURL{%
\tempurl}


\bibitem[AI4Science and Quantum(2023)]%
        {DBLP:journals/corr/abs-2311-07361}
\bibfield{author}{\bibinfo{person}{Microsoft~Research AI4Science} {and} \bibinfo{person}{Microsoft~Azure Quantum}.} \bibinfo{year}{2023}\natexlab{}.
\newblock \showarticletitle{The Impact of Large Language Models on Scientific Discovery: a Preliminary Study using {GPT-4}}.
\newblock \bibinfo{journal}{\emph{arxiv preprint, 2311.07361}} (\bibinfo{year}{2023}).
\newblock


\bibitem[Ansel et~al\mbox{.}(2024)]%
        {DBLP:conf/asplos/AnselYHGJVBBBBC24}
\bibfield{author}{\bibinfo{person}{Jason Ansel}, \bibinfo{person}{Edward~Z. Yang}, \bibinfo{person}{Horace He}, \bibinfo{person}{Natalia Gimelshein}, \bibinfo{person}{Animesh Jain}, \bibinfo{person}{Michael Voznesensky}, \bibinfo{person}{Bin Bao}, \bibinfo{person}{Peter Bell}, \bibinfo{person}{David Berard}, \bibinfo{person}{Evgeni Burovski}, \bibinfo{person}{Geeta Chauhan}, \bibinfo{person}{Anjali Chourdia}, \bibinfo{person}{Will Constable}, \bibinfo{person}{Alban Desmaison}, \bibinfo{person}{Zachary DeVito}, \bibinfo{person}{Elias Ellison}, \bibinfo{person}{Will Feng}, \bibinfo{person}{Jiong Gong}, \bibinfo{person}{Michael Gschwind}, \bibinfo{person}{Brian Hirsh}, \bibinfo{person}{Sherlock Huang}, \bibinfo{person}{Kshiteej Kalambarkar}, \bibinfo{person}{Laurent Kirsch}, \bibinfo{person}{Michael Lazos}, \bibinfo{person}{Mario Lezcano}, \bibinfo{person}{Yanbo Liang}, \bibinfo{person}{Jason Liang}, \bibinfo{person}{Yinghai Lu}, \bibinfo{person}{C.~K. Luk}, \bibinfo{person}{Bert Maher}, \bibinfo{person}{Yunjie
  Pan}, \bibinfo{person}{Christian Puhrsch}, \bibinfo{person}{Matthias Reso}, \bibinfo{person}{Mark Saroufim}, \bibinfo{person}{Marcos~Yukio Siraichi}, \bibinfo{person}{Helen Suk}, \bibinfo{person}{Shunting Zhang}, \bibinfo{person}{Michael Suo}, \bibinfo{person}{Phil Tillet}, \bibinfo{person}{Xu Zhao}, \bibinfo{person}{Eikan Wang}, \bibinfo{person}{Keren Zhou}, \bibinfo{person}{Richard Zou}, \bibinfo{person}{Xiaodong Wang}, \bibinfo{person}{Ajit Mathews}, \bibinfo{person}{William Wen}, \bibinfo{person}{Gregory Chanan}, \bibinfo{person}{Peng Wu}, {and} \bibinfo{person}{Soumith Chintala}.} \bibinfo{year}{2024}\natexlab{}.
\newblock \showarticletitle{PyTorch 2: Faster Machine Learning Through Dynamic Python Bytecode Transformation and Graph Compilation}. In \bibinfo{booktitle}{\emph{Proceedings of the International Conference on Architectural Support for Programming Languages and Operating Systems (ASPLOS)}}.
\newblock


\bibitem[Beltagy et~al\mbox{.}(2020)]%
        {beltagy2020longformerlongdocumenttransformer}
\bibfield{author}{\bibinfo{person}{Iz Beltagy}, \bibinfo{person}{Matthew~E. Peters}, {and} \bibinfo{person}{Arman Cohan}.} \bibinfo{year}{2020}\natexlab{}.
\newblock \showarticletitle{Longformer: The Long-Document Transformer}.
\newblock \bibinfo{journal}{\emph{arxiv preprint ,2004.05150}} (\bibinfo{year}{2020}).
\newblock


\bibitem[Bi et~al\mbox{.}(2023)]%
        {DBLP:journals/nature/BiXZCG023}
\bibfield{author}{\bibinfo{person}{Kaifeng Bi}, \bibinfo{person}{Lingxi Xie}, \bibinfo{person}{Hengheng Zhang}, \bibinfo{person}{Xin Chen}, \bibinfo{person}{Xiaotao Gu}, {and} \bibinfo{person}{Qi Tian}.} \bibinfo{year}{2023}\natexlab{}.
\newblock \showarticletitle{Accurate medium-range global weather forecasting with 3D neural networks}.
\newblock \bibinfo{journal}{\emph{Nat.}}  \bibinfo{volume}{619} (\bibinfo{year}{2023}).
\newblock


\bibitem[Chandra et~al\mbox{.}(2023)]%
        {chandra2023transformer}
\bibfield{author}{\bibinfo{person}{Abel Chandra}, \bibinfo{person}{Laura T{\"u}nnermann}, \bibinfo{person}{Tommy L{\"o}fstedt}, {and} \bibinfo{person}{Regina Gratz}.} \bibinfo{year}{2023}\natexlab{}.
\newblock \showarticletitle{Transformer-based deep learning for predicting protein properties in the life sciences}.
\newblock \bibinfo{journal}{\emph{Elife}}  \bibinfo{volume}{12} (\bibinfo{year}{2023}).
\newblock


\bibitem[Chen et~al\mbox{.}(2024)]%
        {DBLP:journals/corr/abs-2401-09149}
\bibfield{author}{\bibinfo{person}{Qiaoling Chen}, \bibinfo{person}{Diandian Gu}, \bibinfo{person}{Guoteng Wang}, \bibinfo{person}{Xun Chen}, \bibinfo{person}{YingTong Xiong}, \bibinfo{person}{Ting Huang}, \bibinfo{person}{Qinghao Hu}, \bibinfo{person}{Xin Jin}, \bibinfo{person}{Yonggang Wen}, \bibinfo{person}{Tianwei Zhang}, {and} \bibinfo{person}{Peng Sun}.} \bibinfo{year}{2024}\natexlab{}.
\newblock \showarticletitle{InternEvo: Efficient Long-sequence Large Language Model Training via Hybrid Parallelism and Redundant Sharding}.
\newblock \bibinfo{journal}{\emph{arxiv preprint, 2401.09149}} (\bibinfo{year}{2024}).
\newblock


\bibitem[Chen et~al\mbox{.}(2016)]%
        {DBLP:journals/corr/ChenXZG16}
\bibfield{author}{\bibinfo{person}{Tianqi Chen}, \bibinfo{person}{Bing Xu}, \bibinfo{person}{Chiyuan Zhang}, {and} \bibinfo{person}{Carlos Guestrin}.} \bibinfo{year}{2016}\natexlab{}.
\newblock \showarticletitle{Training Deep Nets with Sublinear Memory Cost}.
\newblock \bibinfo{journal}{\emph{arxiv preprint, 1604.06174}} (\bibinfo{year}{2016}).
\newblock


\bibitem[Cloud(2024)]%
        {tongyiqianwen}
\bibfield{author}{\bibinfo{person}{Alibaba Cloud}.} \bibinfo{year}{2024}\natexlab{}.
\newblock \bibinfo{title}{Tongyi Qianwen}.
\newblock
\urldef\tempurl%
\url{https://tongyi.aliyun.com/qianwen/}
\showURL{%
\tempurl}


\bibitem[Contributors(2023)]%
        {torch_mem_utils}
\bibfield{author}{\bibinfo{person}{PyTorch Contributors}.} \bibinfo{year}{2023}\natexlab{}.
\newblock \bibinfo{title}{Understanding CUDA Memory Usage}.
\newblock
\newblock
\shownote{\url{https://pytorch.org/docs/stable/torch_cuda_memory.html}}.


\bibitem[Dao(2024)]%
        {flash2}
\bibfield{author}{\bibinfo{person}{Tri Dao}.} \bibinfo{year}{2024}\natexlab{}.
\newblock \showarticletitle{FlashAttention-2: Faster Attention with Better Parallelism and Work Partitioning}. In \bibinfo{booktitle}{\emph{International Conference on Learning Representations (ICLR)}}.
\newblock


\bibitem[Dao et~al\mbox{.}(2022)]%
        {flash1}
\bibfield{author}{\bibinfo{person}{Tri Dao}, \bibinfo{person}{Daniel~Y. Fu}, \bibinfo{person}{Stefano Ermon}, \bibinfo{person}{Atri Rudra}, {and} \bibinfo{person}{Christopher R{\'{e}}}.} \bibinfo{year}{2022}\natexlab{}.
\newblock \showarticletitle{FlashAttention: Fast and Memory-Efficient Exact Attention with IO-Awareness}. In \bibinfo{booktitle}{\emph{Advances in Neural Information Processing Systems (NeurIPS)}}.
\newblock


\bibitem[Dean et~al\mbox{.}(2012)]%
        {DBLP:conf/nips/DeanCMCDLMRSTYN12}
\bibfield{author}{\bibinfo{person}{Jeffrey Dean}, \bibinfo{person}{Greg Corrado}, \bibinfo{person}{Rajat Monga}, \bibinfo{person}{Kai Chen}, \bibinfo{person}{Matthieu Devin}, \bibinfo{person}{Quoc~V. Le}, \bibinfo{person}{Mark~Z. Mao}, \bibinfo{person}{Marc'Aurelio Ranzato}, \bibinfo{person}{Andrew~W. Senior}, \bibinfo{person}{Paul~A. Tucker}, \bibinfo{person}{Ke Yang}, {and} \bibinfo{person}{Andrew~Y. Ng}.} \bibinfo{year}{2012}\natexlab{}.
\newblock \showarticletitle{Large Scale Distributed Deep Networks}. In \bibinfo{booktitle}{\emph{Advances in Neural Information Processing Systems (NeurIPS)}}.
\newblock


\bibitem[Ding et~al\mbox{.}(2024)]%
        {DBLP:journals/corr/abs-2402-13753}
\bibfield{author}{\bibinfo{person}{Yiran Ding}, \bibinfo{person}{Li~Lyna Zhang}, \bibinfo{person}{Chengruidong Zhang}, \bibinfo{person}{Yuanyuan Xu}, \bibinfo{person}{Ning Shang}, \bibinfo{person}{Jiahang Xu}, \bibinfo{person}{Fan Yang}, {and} \bibinfo{person}{Mao Yang}.} \bibinfo{year}{2024}\natexlab{}.
\newblock \showarticletitle{LongRoPE: Extending {LLM} Context Window Beyond 2 Million Tokens}. In \bibinfo{booktitle}{\emph{Proceedings of the International Conference on Machine Learning (ICML)}}.
\newblock


\bibitem[Dosovitskiy et~al\mbox{.}(2021)]%
        {DBLP:conf/iclr/DosovitskiyB0WZ21}
\bibfield{author}{\bibinfo{person}{Alexey Dosovitskiy}, \bibinfo{person}{Lucas Beyer}, \bibinfo{person}{Alexander Kolesnikov}, \bibinfo{person}{Dirk Weissenborn}, \bibinfo{person}{Xiaohua Zhai}, \bibinfo{person}{Thomas Unterthiner}, \bibinfo{person}{Mostafa Dehghani}, \bibinfo{person}{Matthias Minderer}, \bibinfo{person}{Georg Heigold}, \bibinfo{person}{Sylvain Gelly}, \bibinfo{person}{Jakob Uszkoreit}, {and} \bibinfo{person}{Neil Houlsby}.} \bibinfo{year}{2021}\natexlab{}.
\newblock \showarticletitle{An Image is Worth 16x16 Words: Transformers for Image Recognition at Scale}. In \bibinfo{booktitle}{\emph{International Conference on Learning Representations (ICLR)}}.
\newblock


\bibitem[Durner et~al\mbox{.}(2019)]%
        {defragment1}
\bibfield{author}{\bibinfo{person}{Dominik Durner}, \bibinfo{person}{Viktor Leis}, {and} \bibinfo{person}{Thomas Neumann}.} \bibinfo{year}{2019}\natexlab{}.
\newblock \showarticletitle{On the Impact of Memory Allocation on High-Performance Query Processing}. In \bibinfo{booktitle}{\emph{Proceedings of the 15th International Workshop on Data Management on New Hardware (DaMoN)}}.
\newblock


\bibitem[Fang and Zhao(2024)]%
        {DBLP:journals/corr/abs-2405-07719}
\bibfield{author}{\bibinfo{person}{Jiarui Fang} {and} \bibinfo{person}{Shangchun Zhao}.} \bibinfo{year}{2024}\natexlab{}.
\newblock \showarticletitle{{USP:} {A} Unified Sequence Parallelism Approach for Long Context Generative {AI}}.
\newblock \bibinfo{journal}{\emph{arxiv preprint, 2405.07719}} (\bibinfo{year}{2024}).
\newblock


\bibitem[Fu et~al\mbox{.}(2025)]%
        {DBLP:journals/corr/abs-2402-10171}
\bibfield{author}{\bibinfo{person}{Yao Fu}, \bibinfo{person}{Rameswar Panda}, \bibinfo{person}{Xinyao Niu}, \bibinfo{person}{Xiang Yue}, \bibinfo{person}{Hannaneh Hajishirzi}, \bibinfo{person}{Yoon Kim}, {and} \bibinfo{person}{Hao Peng}.} \bibinfo{year}{2025}\natexlab{}.
\newblock \showarticletitle{Data engineering for scaling language models to 128K context}. In \bibinfo{booktitle}{\emph{Proceedings of the International Conference on Machine Learning (ICML)}}.
\newblock


\bibitem[Gao et~al\mbox{.}(2024)]%
        {simple}
\bibfield{author}{\bibinfo{person}{Shihong Gao}, \bibinfo{person}{Yiming Li}, \bibinfo{person}{Xin Zhang}, \bibinfo{person}{Yanyan Shen}, \bibinfo{person}{Yingxia Shao}, {and} \bibinfo{person}{Lei Chen}.} \bibinfo{year}{2024}\natexlab{}.
\newblock \showarticletitle{{SIMPLE:} Efficient Temporal Graph Neural Network Training at Scale with Dynamic Data Placement}. In \bibinfo{booktitle}{\emph{Proceedings of the ACM on Management of Data (SIGMOD)}}.
\newblock


\bibitem[Ge et~al\mbox{.}(2024)]%
        {hotspa}
\bibfield{author}{\bibinfo{person}{Hao Ge}, \bibinfo{person}{Fangcheng Fu}, \bibinfo{person}{Haoyang Li}, \bibinfo{person}{Xuanyu Wang}, \bibinfo{person}{Sheng Lin}, \bibinfo{person}{Yujie Wang}, \bibinfo{person}{Xiaonan Nie}, \bibinfo{person}{Hailin Zhang}, \bibinfo{person}{Xupeng Miao}, {and} \bibinfo{person}{Bin Cui}.} \bibinfo{year}{2024}\natexlab{}.
\newblock \showarticletitle{Enabling Parallelism Hot Switching for Efficient Training of Large Language Models}. In \bibinfo{booktitle}{\emph{Proceedings of the Symposium on Operating Systems Principles (SOSP)}}.
\newblock


\bibitem[Guan et~al\mbox{.}(2024)]%
        {jcst1}
\bibfield{author}{\bibinfo{person}{Lei Guan}, \bibinfo{person}{Dong{-}Sheng Li}, \bibinfo{person}{Jiye Liang}, \bibinfo{person}{Wen{-}Jian Wang}, \bibinfo{person}{Ke{-}shi Ge}, {and} \bibinfo{person}{Xicheng Lu}.} \bibinfo{year}{2024}\natexlab{}.
\newblock \showarticletitle{Advances of Pipeline Model Parallelism for Deep Learning Training: An Overview}.
\newblock \bibinfo{journal}{\emph{J. Comput. Sci. Technol.}}  \bibinfo{volume}{39} (\bibinfo{year}{2024}).
\newblock


\bibitem[Guo et~al\mbox{.}(2024a)]%
        {DBLP:conf/asplos/0003ZXLLHGWZZZ24}
\bibfield{author}{\bibinfo{person}{Cong Guo}, \bibinfo{person}{Rui Zhang}, \bibinfo{person}{Jiale Xu}, \bibinfo{person}{Jingwen Leng}, \bibinfo{person}{Zihan Liu}, \bibinfo{person}{Ziyu Huang}, \bibinfo{person}{Minyi Guo}, \bibinfo{person}{Hao Wu}, \bibinfo{person}{Shouren Zhao}, \bibinfo{person}{Junping Zhao}, {and} \bibinfo{person}{Ke Zhang}.} \bibinfo{year}{2024}\natexlab{a}.
\newblock \showarticletitle{GMLake: Efficient and Transparent {GPU} Memory Defragmentation for Large-scale {DNN} Training with Virtual Memory Stitching}. In \bibinfo{booktitle}{\emph{Proceedings of the International Conference on Architectural Support for Programming Languages and Operating Systems (ASPLOS)}}.
\newblock


\bibitem[Guo et~al\mbox{.}(2024b)]%
        {DBLP:journals/corr/abs-2401-14196}
\bibfield{author}{\bibinfo{person}{Daya Guo}, \bibinfo{person}{Qihao Zhu}, \bibinfo{person}{Dejian Yang}, \bibinfo{person}{Zhenda Xie}, \bibinfo{person}{Kai Dong}, \bibinfo{person}{Wentao Zhang}, \bibinfo{person}{Guanting Chen}, \bibinfo{person}{Xiao Bi}, \bibinfo{person}{Y. Wu}, \bibinfo{person}{Y.~K. Li}, \bibinfo{person}{Fuli Luo}, \bibinfo{person}{Yingfei Xiong}, {and} \bibinfo{person}{Wenfeng Liang}.} \bibinfo{year}{2024}\natexlab{b}.
\newblock \showarticletitle{DeepSeek-Coder: When the Large Language Model Meets Programming - The Rise of Code Intelligence}.
\newblock \bibinfo{journal}{\emph{arxiv preprint, 2401.14196}} (\bibinfo{year}{2024}).
\newblock


\bibitem[{Gurobi Optimization, LLC}(2024)]%
        {gurobi}
\bibfield{author}{\bibinfo{person}{{Gurobi Optimization, LLC}}.} \bibinfo{year}{2024}\natexlab{}.
\newblock \bibinfo{title}{{Gurobi Optimizer Reference Manual}}.
\newblock


\bibitem[Hu et~al\mbox{.}(2022)]%
        {DBLP:conf/ics/HuXDLZZMST22}
\bibfield{author}{\bibinfo{person}{Zhongzhe Hu}, \bibinfo{person}{Junmin Xiao}, \bibinfo{person}{Zheye Deng}, \bibinfo{person}{Mingyi Li}, \bibinfo{person}{Kewei Zhang}, \bibinfo{person}{Xiaoyang Zhang}, \bibinfo{person}{Ke Meng}, \bibinfo{person}{Ninghui Sun}, {and} \bibinfo{person}{Guangming Tan}.} \bibinfo{year}{2022}\natexlab{}.
\newblock \showarticletitle{MegTaiChi: dynamic tensor-based memory management optimization for {DNN} training}. In \bibinfo{booktitle}{\emph{International Conference on Supercomputing (ICS)}}.
\newblock


\bibitem[Huang et~al\mbox{.}(2019)]%
        {DBLP:conf/nips/HuangCBFCCLNLWC19}
\bibfield{author}{\bibinfo{person}{Yanping Huang}, \bibinfo{person}{Youlong Cheng}, \bibinfo{person}{Ankur Bapna}, \bibinfo{person}{Orhan Firat}, \bibinfo{person}{Dehao Chen}, \bibinfo{person}{Mia~Xu Chen}, \bibinfo{person}{HyoukJoong Lee}, \bibinfo{person}{Jiquan Ngiam}, \bibinfo{person}{Quoc~V. Le}, \bibinfo{person}{Yonghui Wu}, {and} \bibinfo{person}{Zhifeng Chen}.} \bibinfo{year}{2019}\natexlab{}.
\newblock \showarticletitle{GPipe: Efficient Training of Giant Neural Networks using Pipeline Parallelism}. In \bibinfo{booktitle}{\emph{Advances in Neural Information Processing Systems (NeurIPS)}}.
\newblock


\bibitem[Isenko et~al\mbox{.}(2022)]%
        {where}
\bibfield{author}{\bibinfo{person}{Alexander Isenko}, \bibinfo{person}{Ruben Mayer}, \bibinfo{person}{Jeffrey Jedele}, {and} \bibinfo{person}{Hans{-}Arno Jacobsen}.} \bibinfo{year}{2022}\natexlab{}.
\newblock \showarticletitle{Where Is My Training Bottleneck? Hidden Trade-Offs in Deep Learning Preprocessing Pipelines}. In \bibinfo{booktitle}{\emph{Proceedings of the ACM on Management of Data (SIGMOD)}}.
\newblock


\bibitem[Jacobs et~al\mbox{.}(2023)]%
        {DBLP:journals/corr/abs-2309-14509}
\bibfield{author}{\bibinfo{person}{Sam~Ade Jacobs}, \bibinfo{person}{Masahiro Tanaka}, \bibinfo{person}{Chengming Zhang}, \bibinfo{person}{Minjia Zhang}, \bibinfo{person}{Shuaiwen~Leon Song}, \bibinfo{person}{Samyam Rajbhandari}, {and} \bibinfo{person}{Yuxiong He}.} \bibinfo{year}{2023}\natexlab{}.
\newblock \showarticletitle{DeepSpeed Ulysses: System Optimizations for Enabling Training of Extreme Long Sequence Transformer Models}.
\newblock \bibinfo{journal}{\emph{arxiv preprint, 2309.14509}} (\bibinfo{year}{2023}).
\newblock


\bibitem[Jiang et~al\mbox{.}(2023)]%
        {DBLP:journals/corr/abs-2310-06825}
\bibfield{author}{\bibinfo{person}{Albert~Q. Jiang}, \bibinfo{person}{Alexandre Sablayrolles}, \bibinfo{person}{Arthur Mensch}, \bibinfo{person}{Chris Bamford}, \bibinfo{person}{Devendra~Singh Chaplot}, \bibinfo{person}{Diego de Las~Casas}, \bibinfo{person}{Florian Bressand}, \bibinfo{person}{Gianna Lengyel}, \bibinfo{person}{Guillaume Lample}, \bibinfo{person}{Lucile Saulnier}, \bibinfo{person}{L{\'{e}}lio~Renard Lavaud}, \bibinfo{person}{Marie{-}Anne Lachaux}, \bibinfo{person}{Pierre Stock}, \bibinfo{person}{Teven~Le Scao}, \bibinfo{person}{Thibaut Lavril}, \bibinfo{person}{Thomas Wang}, \bibinfo{person}{Timoth{\'{e}}e Lacroix}, {and} \bibinfo{person}{William~El Sayed}.} \bibinfo{year}{2023}\natexlab{}.
\newblock \showarticletitle{Mistral 7B}.
\newblock \bibinfo{journal}{\emph{arxiv preprint, 2310.06825}} (\bibinfo{year}{2023}).
\newblock


\bibitem[Johnson et~al\mbox{.}(2021)]%
        {faiss}
\bibfield{author}{\bibinfo{person}{Jeff Johnson}, \bibinfo{person}{Matthijs Douze}, {and} \bibinfo{person}{Herv{\'{e}} J{\'{e}}gou}.} \bibinfo{year}{2021}\natexlab{}.
\newblock \showarticletitle{Billion-Scale Similarity Search with GPUs}.
\newblock \bibinfo{journal}{\emph{{IEEE} Trans. Big Data}}  \bibinfo{volume}{7} (\bibinfo{year}{2021}).
\newblock


\bibitem[Kalamkar et~al\mbox{.}(2019)]%
        {bf16_mixed_precition}
\bibfield{author}{\bibinfo{person}{Dhiraj~D. Kalamkar}, \bibinfo{person}{Dheevatsa Mudigere}, \bibinfo{person}{Naveen Mellempudi}, \bibinfo{person}{Dipankar Das}, \bibinfo{person}{Kunal Banerjee}, \bibinfo{person}{Sasikanth Avancha}, \bibinfo{person}{Dharma~Teja Vooturi}, \bibinfo{person}{Nataraj Jammalamadaka}, \bibinfo{person}{Jianyu Huang}, \bibinfo{person}{Hector Yuen}, \bibinfo{person}{Jiyan Yang}, \bibinfo{person}{Jongsoo Park}, \bibinfo{person}{Alexander Heinecke}, \bibinfo{person}{Evangelos Georganas}, \bibinfo{person}{Sudarshan Srinivasan}, \bibinfo{person}{Abhisek Kundu}, \bibinfo{person}{Misha Smelyanskiy}, \bibinfo{person}{Bharat Kaul}, {and} \bibinfo{person}{Pradeep Dubey}.} \bibinfo{year}{2019}\natexlab{}.
\newblock \showarticletitle{A Study of {BFLOAT16} for Deep Learning Training}.
\newblock \bibinfo{journal}{\emph{arxiv preprint, 1905.12322}} (\bibinfo{year}{2019}).
\newblock


\bibitem[Katharopoulos et~al\mbox{.}(2020)]%
        {transformersarernn}
\bibfield{author}{\bibinfo{person}{Angelos Katharopoulos}, \bibinfo{person}{Apoorv Vyas}, \bibinfo{person}{Nikolaos Pappas}, {and} \bibinfo{person}{Fran\c{c}ois Fleuret}.} \bibinfo{year}{2020}\natexlab{}.
\newblock \showarticletitle{Transformers are RNNs: fast autoregressive transformers with linear attention}. In \bibinfo{booktitle}{\emph{Proceedings of the International Conference on Machine Learning (ICML)}}.
\newblock


\bibitem[Khamis et~al\mbox{.}(2020)]%
        {learningmodelsoverrelational}
\bibfield{author}{\bibinfo{person}{Mahmoud~Abo Khamis}, \bibinfo{person}{Hung~Q. Ngo}, \bibinfo{person}{XuanLong Nguyen}, \bibinfo{person}{Dan Olteanu}, {and} \bibinfo{person}{Maximilian Schleich}.} \bibinfo{year}{2020}\natexlab{}.
\newblock \showarticletitle{Learning Models over Relational Data Using Sparse Tensors and Functional Dependencies}.
\newblock \bibinfo{journal}{\emph{{ACM} Trans. Database Syst.}}  \bibinfo{volume}{45} (\bibinfo{year}{2020}).
\newblock


\bibitem[Kirisame et~al\mbox{.}(2021)]%
        {DBLP:conf/iclr/KirisameLHBHRCT21}
\bibfield{author}{\bibinfo{person}{Marisa Kirisame}, \bibinfo{person}{Steven Lyubomirsky}, \bibinfo{person}{Altan Haan}, \bibinfo{person}{Jennifer Brennan}, \bibinfo{person}{Mike He}, \bibinfo{person}{Jared Roesch}, \bibinfo{person}{Tianqi Chen}, {and} \bibinfo{person}{Zachary Tatlock}.} \bibinfo{year}{2021}\natexlab{}.
\newblock \showarticletitle{Dynamic Tensor Rematerialization}. In \bibinfo{booktitle}{\emph{International Conference on Learning Representations (ICLR)}}.
\newblock


\bibitem[Korthikanti et~al\mbox{.}(2022)]%
        {DBLP:journals/corr/abs-2205-05198}
\bibfield{author}{\bibinfo{person}{Vijay Korthikanti}, \bibinfo{person}{Jared Casper}, \bibinfo{person}{Sangkug Lym}, \bibinfo{person}{Lawrence McAfee}, \bibinfo{person}{Michael Andersch}, \bibinfo{person}{Mohammad Shoeybi}, {and} \bibinfo{person}{Bryan Catanzaro}.} \bibinfo{year}{2022}\natexlab{}.
\newblock \showarticletitle{Reducing Activation Recomputation in Large Transformer Models}.
\newblock \bibinfo{journal}{\emph{arxiv preprint, 2205.05198}} (\bibinfo{year}{2022}).
\newblock


\bibitem[Koutsoukos et~al\mbox{.}(2021)]%
        {tensors}
\bibfield{author}{\bibinfo{person}{Dimitrios Koutsoukos}, \bibinfo{person}{Supun Nakandala}, \bibinfo{person}{Konstantinos Karanasos}, \bibinfo{person}{Karla Saur}, \bibinfo{person}{Gustavo Alonso}, {and} \bibinfo{person}{Matteo Interlandi}.} \bibinfo{year}{2021}\natexlab{}.
\newblock \showarticletitle{Tensors: An abstraction for general data processing}.
\newblock \bibinfo{journal}{\emph{Proceedings of the VLDB Endowment}}  \bibinfo{volume}{14} (\bibinfo{year}{2021}).
\newblock


\bibitem[Li et~al\mbox{.}(2023a)]%
        {DBLP:journals/corr/abs-2310-03294}
\bibfield{author}{\bibinfo{person}{Dacheng Li}, \bibinfo{person}{Rulin Shao}, \bibinfo{person}{Anze Xie}, \bibinfo{person}{Eric~P. Xing}, \bibinfo{person}{Joseph~E. Gonzalez}, \bibinfo{person}{Ion Stoica}, \bibinfo{person}{Xuezhe Ma}, {and} \bibinfo{person}{Hao Zhang}.} \bibinfo{year}{2023}\natexlab{a}.
\newblock \showarticletitle{LightSeq: Sequence Level Parallelism for Distributed Training of Long Context Transformers}.
\newblock \bibinfo{journal}{\emph{arxiv preprint, 2310.03294}} (\bibinfo{year}{2023}).
\newblock


\bibitem[Li et~al\mbox{.}(2014)]%
        {limu}
\bibfield{author}{\bibinfo{person}{Mu Li}, \bibinfo{person}{David~G. Anderson}, \bibinfo{person}{Jun~Woo Park}, \bibinfo{person}{Alexander~J. Smola}, \bibinfo{person}{Amr Ahmed}, \bibinfo{person}{Vanja Josifovski}, \bibinfo{person}{James Long}, \bibinfo{person}{Eugene~J. Shekita}, {and} \bibinfo{person}{Bor-Yiing Su}.} \bibinfo{year}{2014}\natexlab{}.
\newblock \showarticletitle{Scaling Distributed Machine Learning with the Parameter Server}. In \bibinfo{booktitle}{\emph{Proceedings of the USENIX Symposium on Operating Systems Design and Implementation (OSDI)}}.
\newblock


\bibitem[Li et~al\mbox{.}(2023b)]%
        {orca}
\bibfield{author}{\bibinfo{person}{Yiming Li}, \bibinfo{person}{Yanyan Shen}, \bibinfo{person}{Lei Chen}, {and} \bibinfo{person}{Mingxuan Yuan}.} \bibinfo{year}{2023}\natexlab{b}.
\newblock \showarticletitle{Orca: Scalable Temporal Graph Neural Network Training with Theoretical Guarantees}. In \bibinfo{booktitle}{\emph{Proceedings of the ACM on Management of Data (SIGMOD)}}.
\newblock


\bibitem[Li et~al\mbox{.}(2022)]%
        {DBLP:journals/corr/abs-2201-11838}
\bibfield{author}{\bibinfo{person}{Yikuan Li}, \bibinfo{person}{Ramsey~M. Wehbe}, \bibinfo{person}{Faraz~S. Ahmad}, \bibinfo{person}{Hanyin Wang}, {and} \bibinfo{person}{Yuan Luo}.} \bibinfo{year}{2022}\natexlab{}.
\newblock \showarticletitle{Clinical-Longformer and Clinical-BigBird: Transformers for long clinical sequences}.
\newblock \bibinfo{journal}{\emph{arxiv preprint, 2201.11838}} (\bibinfo{year}{2022}).
\newblock


\bibitem[Liu et~al\mbox{.}(2024b)]%
        {DBLP:journals/corr/abs-2402-08268}
\bibfield{author}{\bibinfo{person}{Hao Liu}, \bibinfo{person}{Wilson Yan}, \bibinfo{person}{Matei Zaharia}, {and} \bibinfo{person}{Pieter Abbeel}.} \bibinfo{year}{2024}\natexlab{b}.
\newblock \showarticletitle{World Model on Million-Length Video And Language With Blockwise RingAttention}.
\newblock \bibinfo{journal}{\emph{arxiv preprint, 2402.08268}} (\bibinfo{year}{2024}).
\newblock


\bibitem[Liu et~al\mbox{.}(2023)]%
        {DBLP:journals/corr/abs-2310-01889}
\bibfield{author}{\bibinfo{person}{Hao Liu}, \bibinfo{person}{Matei Zaharia}, {and} \bibinfo{person}{Pieter Abbeel}.} \bibinfo{year}{2023}\natexlab{}.
\newblock \showarticletitle{Ring Attention with Blockwise Transformers for Near-Infinite Context}.
\newblock \bibinfo{journal}{\emph{arxiv preprint, 2310.01889}} (\bibinfo{year}{2023}).
\newblock


\bibitem[Liu et~al\mbox{.}(2024a)]%
        {DBLP:journals/corr/abs-2310-05209}
\bibfield{author}{\bibinfo{person}{Xiaoran Liu}, \bibinfo{person}{Hang Yan}, \bibinfo{person}{Chenxin An}, \bibinfo{person}{Xipeng Qiu}, {and} \bibinfo{person}{Dahua Lin}.} \bibinfo{year}{2024}\natexlab{a}.
\newblock \showarticletitle{Scaling Laws of RoPE-based Extrapolation}. In \bibinfo{booktitle}{\emph{International Conference on Learning Representations (ICLR)}}.
\newblock


\bibitem[Miao et~al\mbox{.}(2024a)]%
        {demystify}
\bibfield{author}{\bibinfo{person}{Xupeng Miao}, \bibinfo{person}{Zhihao Jia}, {and} \bibinfo{person}{Bin Cui}.} \bibinfo{year}{2024}\natexlab{a}.
\newblock \showarticletitle{Demystifying Data Management for Large Language Models}. In \bibinfo{booktitle}{\emph{Proceedings of the ACM on Management of Data (SIGMOD)}}.
\newblock


\bibitem[Miao et~al\mbox{.}(2023)]%
        {sdpipe}
\bibfield{author}{\bibinfo{person}{Xupeng Miao}, \bibinfo{person}{Yining Shi}, \bibinfo{person}{Zhi Yang}, \bibinfo{person}{Bin Cui}, {and} \bibinfo{person}{Zhihao Jia}.} \bibinfo{year}{2023}\natexlab{}.
\newblock \showarticletitle{SDPipe: {A} Semi-Decentralized Framework for Heterogeneity-aware Pipeline-parallel Training}.
\newblock \bibinfo{journal}{\emph{Proceedings of the VLDB Endowment}}  \bibinfo{volume}{16} (\bibinfo{year}{2023}).
\newblock


\bibitem[Miao et~al\mbox{.}(2022)]%
        {galvatron}
\bibfield{author}{\bibinfo{person}{Xupeng Miao}, \bibinfo{person}{Yujie Wang}, \bibinfo{person}{Youhe Jiang}, \bibinfo{person}{Chunan Shi}, \bibinfo{person}{Xiaonan Nie}, \bibinfo{person}{Hailin Zhang}, {and} \bibinfo{person}{Bin Cui}.} \bibinfo{year}{2022}\natexlab{}.
\newblock \showarticletitle{Galvatron: Efficient Transformer Training over Multiple GPUs Using Automatic Parallelism}.
\newblock \bibinfo{journal}{\emph{Proceedings of the VLDB Endowment}}  \bibinfo{volume}{16} (\bibinfo{year}{2022}).
\newblock


\bibitem[Miao et~al\mbox{.}(2021)]%
        {het}
\bibfield{author}{\bibinfo{person}{Xupeng Miao}, \bibinfo{person}{Hailin Zhang}, \bibinfo{person}{Yining Shi}, \bibinfo{person}{Xiaonan Nie}, \bibinfo{person}{Zhi Yang}, \bibinfo{person}{Yangyu Tao}, {and} \bibinfo{person}{Bin Cui}.} \bibinfo{year}{2021}\natexlab{}.
\newblock \showarticletitle{{HET:} Scaling out Huge Embedding Model Training via Cache-enabled Distributed Framework}.
\newblock \bibinfo{journal}{\emph{Proceedings of the VLDB Endowment}}  \bibinfo{volume}{15} (\bibinfo{year}{2021}).
\newblock


\bibitem[Miao et~al\mbox{.}(2024b)]%
        {efficient_transformer}
\bibfield{author}{\bibinfo{person}{Xupeng Miao}, \bibinfo{person}{Shenhan Zhu}, \bibinfo{person}{Fangcheng Fu}, \bibinfo{person}{Ziyu Guo}, \bibinfo{person}{Zhi Yang}, \bibinfo{person}{Yaofeng Tu}, \bibinfo{person}{Zhihao Jia}, {and} \bibinfo{person}{Bin Cui}.} \bibinfo{year}{2024}\natexlab{b}.
\newblock \showarticletitle{X-former Elucidator: Reviving Efficient Attention for Long Context Language Modeling}. In \bibinfo{booktitle}{\emph{Proceedings of the International Joint Conference on Artificial Intelligence (IJCAI)}}.
\newblock


\bibitem[Micikevicius et~al\mbox{.}(2018)]%
        {mixed_precision}
\bibfield{author}{\bibinfo{person}{Paulius Micikevicius}, \bibinfo{person}{Sharan Narang}, \bibinfo{person}{Jonah Alben}, \bibinfo{person}{Gregory Diamos}, \bibinfo{person}{Erich Elsen}, \bibinfo{person}{David Garcia}, \bibinfo{person}{Boris Ginsburg}, \bibinfo{person}{Michael Houston}, \bibinfo{person}{Oleksii Kuchaiev}, \bibinfo{person}{Ganesh Venkatesh}, {and} \bibinfo{person}{Hao Wu}.} \bibinfo{year}{2018}\natexlab{}.
\newblock \showarticletitle{Mixed Precision Training}. In \bibinfo{booktitle}{\emph{International Conference on Learning Representations (ICLR)}}.
\newblock


\bibitem[Mohan et~al\mbox{.}(2021)]%
        {stall}
\bibfield{author}{\bibinfo{person}{Jayashree Mohan}, \bibinfo{person}{Amar Phanishayee}, \bibinfo{person}{Ashish Raniwala}, {and} \bibinfo{person}{Vijay Chidambaram}.} \bibinfo{year}{2021}\natexlab{}.
\newblock \showarticletitle{Analyzing and Mitigating Data Stalls in {DNN} Training}.
\newblock \bibinfo{journal}{\emph{Proceedings of the VLDB Endowment}}  \bibinfo{volume}{14} (\bibinfo{year}{2021}).
\newblock


\bibitem[Murray et~al\mbox{.}(2021)]%
        {tfdata}
\bibfield{author}{\bibinfo{person}{Derek~Gordon Murray}, \bibinfo{person}{Jiri Simsa}, \bibinfo{person}{Ana Klimovic}, {and} \bibinfo{person}{Ihor Indyk}.} \bibinfo{year}{2021}\natexlab{}.
\newblock \showarticletitle{tf.data: {A} Machine Learning Data Processing Framework}.
\newblock \bibinfo{journal}{\emph{Proceedings of the VLDB Endowment}}  \bibinfo{volume}{14} (\bibinfo{year}{2021}).
\newblock


\bibitem[Nakandala et~al\mbox{.}(2020)]%
        {atensorcompiler}
\bibfield{author}{\bibinfo{person}{Supun Nakandala}, \bibinfo{person}{Karla Saur}, \bibinfo{person}{Gyeong{-}In Yu}, \bibinfo{person}{Konstantinos Karanasos}, \bibinfo{person}{Carlo Curino}, \bibinfo{person}{Markus Weimer}, {and} \bibinfo{person}{Matteo Interlandi}.} \bibinfo{year}{2020}\natexlab{}.
\newblock \showarticletitle{A Tensor Compiler for Unified Machine Learning Prediction Serving}. In \bibinfo{booktitle}{\emph{Proceedings of the USENIX Symposium on Operating Systems Design and Implementation (OSDI)}}.
\newblock


\bibitem[Narayanan et~al\mbox{.}(2021)]%
        {DBLP:conf/icml/NarayananPSCZ21}
\bibfield{author}{\bibinfo{person}{Deepak Narayanan}, \bibinfo{person}{Amar Phanishayee}, \bibinfo{person}{Kaiyu Shi}, \bibinfo{person}{Xie Chen}, {and} \bibinfo{person}{Matei Zaharia}.} \bibinfo{year}{2021}\natexlab{}.
\newblock \showarticletitle{Memory-Efficient Pipeline-Parallel {DNN} Training}. In \bibinfo{booktitle}{\emph{Proceedings of the International Conference on Machine Learning (ICML)}}.
\newblock


\bibitem[Nguyen et~al\mbox{.}(2023)]%
        {DBLP:conf/icml/NguyenBKGG23}
\bibfield{author}{\bibinfo{person}{Tung Nguyen}, \bibinfo{person}{Johannes Brandstetter}, \bibinfo{person}{Ashish Kapoor}, \bibinfo{person}{Jayesh~K. Gupta}, {and} \bibinfo{person}{Aditya Grover}.} \bibinfo{year}{2023}\natexlab{}.
\newblock \showarticletitle{ClimaX: {A} foundation model for weather and climate}. In \bibinfo{booktitle}{\emph{Proceedings of the International Conference on Machine Learning (ICML)}}.
\newblock


\bibitem[NVIDIA(2024a)]%
        {contextparallel}
\bibfield{author}{\bibinfo{person}{NVIDIA}.} \bibinfo{year}{2024}\natexlab{a}.
\newblock \bibinfo{title}{Context Parallelism}.
\newblock
\newblock
\shownote{\url{https://docs.nvidia.com/megatron-core/developer-guide/latest/api-guide/context_parallel.html}}.


\bibitem[NVIDIA(2024b)]%
        {transformerengine}
\bibfield{author}{\bibinfo{person}{NVIDIA}.} \bibinfo{year}{2024}\natexlab{b}.
\newblock \bibinfo{title}{Transformer Engine}.
\newblock
\newblock
\shownote{\url{https://github.com/NVIDIA/TransformerEngine}}.


\bibitem[Ooi et~al\mbox{.}(2024)]%
        {scis1}
\bibfield{author}{\bibinfo{person}{Beng~Chin Ooi}, \bibinfo{person}{Shaofeng Cai}, \bibinfo{person}{Gang Chen}, \bibinfo{person}{Kian{-}Lee Tan}, \bibinfo{person}{Yuncheng Wu}, \bibinfo{person}{Xiaokui Xiao}, \bibinfo{person}{Naili Xing}, \bibinfo{person}{Cong Yue}, \bibinfo{person}{Lingze Zeng}, \bibinfo{person}{Meihui Zhang}, {and} \bibinfo{person}{Zhanhao Zhao}.} \bibinfo{year}{2024}\natexlab{}.
\newblock \showarticletitle{NeurDB: An AI-powered Autonomous Data System}.
\newblock \bibinfo{journal}{\emph{Science China Information Sciences}}  \bibinfo{volume}{67} (\bibinfo{year}{2024}).
\newblock


\bibitem[OpenAI(2023)]%
        {DBLP:journals/corr/abs-2303-08774}
\bibfield{author}{\bibinfo{person}{OpenAI}.} \bibinfo{year}{2023}\natexlab{}.
\newblock \showarticletitle{{GPT-4} Technical Report}.
\newblock  (\bibinfo{year}{2023}).
\newblock


\bibitem[Oukid et~al\mbox{.}(2017)]%
        {defragment2}
\bibfield{author}{\bibinfo{person}{Ismail Oukid}, \bibinfo{person}{Daniel Booss}, \bibinfo{person}{Adrien Lespinasse}, \bibinfo{person}{Wolfgang Lehner}, \bibinfo{person}{Thomas Willhalm}, {and} \bibinfo{person}{Gr\'{e}goire Gomes}.} \bibinfo{year}{2017}\natexlab{}.
\newblock \showarticletitle{Memory management techniques for large-scale persistent-main-memory systems}.
\newblock \bibinfo{journal}{\emph{Proceedings of the VLDB Endowment}}  \bibinfo{volume}{10} (\bibinfo{year}{2017}).
\newblock


\bibitem[Paszke et~al\mbox{.}(2019)]%
        {DBLP:conf/nips/PaszkeGMLBCKLGA19}
\bibfield{author}{\bibinfo{person}{Adam Paszke}, \bibinfo{person}{Sam Gross}, \bibinfo{person}{Francisco Massa}, \bibinfo{person}{Adam Lerer}, \bibinfo{person}{James Bradbury}, \bibinfo{person}{Gregory Chanan}, \bibinfo{person}{Trevor Killeen}, \bibinfo{person}{Zeming Lin}, \bibinfo{person}{Natalia Gimelshein}, \bibinfo{person}{Luca Antiga}, \bibinfo{person}{Alban Desmaison}, \bibinfo{person}{Andreas K{\"{o}}pf}, \bibinfo{person}{Edward~Z. Yang}, \bibinfo{person}{Zachary DeVito}, \bibinfo{person}{Martin Raison}, \bibinfo{person}{Alykhan Tejani}, \bibinfo{person}{Sasank Chilamkurthy}, \bibinfo{person}{Benoit Steiner}, \bibinfo{person}{Lu Fang}, \bibinfo{person}{Junjie Bai}, {and} \bibinfo{person}{Soumith Chintala}.} \bibinfo{year}{2019}\natexlab{}.
\newblock \showarticletitle{PyTorch: An Imperative Style, High-Performance Deep Learning Library}. In \bibinfo{booktitle}{\emph{Advances in Neural Information Processing Systems (NeurIPS)}}.
\newblock


\bibitem[Peebles and Xie(2023)]%
        {DBLP:conf/iccv/PeeblesX23}
\bibfield{author}{\bibinfo{person}{William Peebles} {and} \bibinfo{person}{Saining Xie}.} \bibinfo{year}{2023}\natexlab{}.
\newblock \showarticletitle{Scalable Diffusion Models with Transformers}. In \bibinfo{booktitle}{\emph{IEEE/CVF International Conference on Computer Vision (ICCV)}}.
\newblock


\bibitem[Peng et~al\mbox{.}(2024)]%
        {DBLP:journals/corr/abs-2309-00071}
\bibfield{author}{\bibinfo{person}{Bowen Peng}, \bibinfo{person}{Jeffrey Quesnelle}, \bibinfo{person}{Honglu Fan}, {and} \bibinfo{person}{Enrico Shippole}.} \bibinfo{year}{2024}\natexlab{}.
\newblock \showarticletitle{YaRN: Efficient Context Window Extension of Large Language Models}. In \bibinfo{booktitle}{\emph{International Conference on Learning Representations (ICLR)}}.
\newblock


\bibitem[Peng et~al\mbox{.}(2020)]%
        {peng2020capuchin}
\bibfield{author}{\bibinfo{person}{Quan Peng}, \bibinfo{person}{Xuanhua Shi}, \bibinfo{person}{Hulin Dai}, \bibinfo{person}{Hai Jin}, \bibinfo{person}{Weiliang Ma}, \bibinfo{person}{Qian Xiong}, \bibinfo{person}{Fan Yang}, {and} \bibinfo{person}{Xuehai Qian}.} \bibinfo{year}{2020}\natexlab{}.
\newblock \showarticletitle{Capuchin: Tensor-based GPU Memory Management for Deep Learning}. In \bibinfo{booktitle}{\emph{Proceedings of the International Conference on Architectural Support for Programming Languages and Operating Systems (ASPLOS)}}.
\newblock


\bibitem[Phan and Li(2008)]%
        {dynamic}
\bibfield{author}{\bibinfo{person}{Thomas Phan} {and} \bibinfo{person}{Wen-Syan Li}.} \bibinfo{year}{2008}\natexlab{}.
\newblock \showarticletitle{Dynamic Materialization of Query Views for Data Warehouse Workloads}. In \bibinfo{booktitle}{\emph{IEEE International Conference on Data Engineering (ICDE)}}.
\newblock


\bibitem[Pope et~al\mbox{.}(2023)]%
        {mfu}
\bibfield{author}{\bibinfo{person}{Reiner Pope}, \bibinfo{person}{Sholto Douglas}, \bibinfo{person}{Aakanksha Chowdhery}, \bibinfo{person}{Jacob Devlin}, \bibinfo{person}{James Bradbury}, \bibinfo{person}{Jonathan Heek}, \bibinfo{person}{Kefan Xiao}, \bibinfo{person}{Shivani Agrawal}, {and} \bibinfo{person}{Jeff Dean}.} \bibinfo{year}{2023}\natexlab{}.
\newblock \showarticletitle{Efficiently Scaling Transformer Inference}. In \bibinfo{booktitle}{\emph{Proceedings of the Conference on Machine Learning and Systems (MLSys)}}.
\newblock


\bibitem[Press et~al\mbox{.}(2022)]%
        {DBLP:conf/iclr/PressSL22}
\bibfield{author}{\bibinfo{person}{Ofir Press}, \bibinfo{person}{Noah~A. Smith}, {and} \bibinfo{person}{Mike Lewis}.} \bibinfo{year}{2022}\natexlab{}.
\newblock \showarticletitle{Train Short, Test Long: Attention with Linear Biases Enables Input Length Extrapolation}. In \bibinfo{booktitle}{\emph{International Conference on Learning Representations (ICLR)}}.
\newblock


\bibitem[Qin et~al\mbox{.}(2022)]%
        {qin2022cosformerrethinkingsoftmaxattention}
\bibfield{author}{\bibinfo{person}{Zhen Qin}, \bibinfo{person}{Weixuan Sun}, \bibinfo{person}{Hui Deng}, \bibinfo{person}{Dongxu Li}, \bibinfo{person}{Yunshen Wei}, \bibinfo{person}{Baohong Lv}, \bibinfo{person}{Junjie Yan}, \bibinfo{person}{Lingpeng Kong}, {and} \bibinfo{person}{Yiran Zhong}.} \bibinfo{year}{2022}\natexlab{}.
\newblock \showarticletitle{cosFormer: Rethinking Softmax In Attention}. In \bibinfo{booktitle}{\emph{International Conference on Learning Representations (ICLR)}}.
\newblock


\bibitem[Rajbhandari et~al\mbox{.}(2019)]%
        {zero}
\bibfield{author}{\bibinfo{person}{Samyam Rajbhandari}, \bibinfo{person}{Jeff Rasley}, \bibinfo{person}{Olatunji Ruwase}, {and} \bibinfo{person}{Yuxiong He}.} \bibinfo{year}{2019}\natexlab{}.
\newblock \showarticletitle{ZeRO: Memory Optimization Towards Training {A} Trillion Parameter Models}.
\newblock \bibinfo{journal}{\emph{arxiv preprint, 1910.02054}} (\bibinfo{year}{2019}).
\newblock


\bibitem[Rasley et~al\mbox{.}(2020)]%
        {DBLP:conf/kdd/RasleyRRH20}
\bibfield{author}{\bibinfo{person}{Jeff Rasley}, \bibinfo{person}{Samyam Rajbhandari}, \bibinfo{person}{Olatunji Ruwase}, {and} \bibinfo{person}{Yuxiong He}.} \bibinfo{year}{2020}\natexlab{}.
\newblock \showarticletitle{DeepSpeed: System Optimizations Enable Training Deep Learning Models with Over 100 Billion Parameters}. In \bibinfo{booktitle}{\emph{Conference on Knowledge Discovery and Data Mining (KDD)}}.
\newblock


\bibitem[Ren et~al\mbox{.}(2021)]%
        {zero-offload}
\bibfield{author}{\bibinfo{person}{Jie Ren}, \bibinfo{person}{Samyam Rajbhandari}, \bibinfo{person}{Reza~Yazdani Aminabadi}, \bibinfo{person}{Olatunji Ruwase}, \bibinfo{person}{Shuangyan Yang}, \bibinfo{person}{Minjia Zhang}, \bibinfo{person}{Dong Li}, {and} \bibinfo{person}{Yuxiong He}.} \bibinfo{year}{2021}\natexlab{}.
\newblock \showarticletitle{ZeRO-Offload: Democratizing Billion-Scale Model Training}.
\newblock \bibinfo{journal}{\emph{arxiv preprint, 2101.06840}} (\bibinfo{year}{2021}).
\newblock


\bibitem[Rhu et~al\mbox{.}(2016)]%
        {vdnn}
\bibfield{author}{\bibinfo{person}{Minsoo Rhu}, \bibinfo{person}{Natalia Gimelshein}, \bibinfo{person}{Jason Clemons}, \bibinfo{person}{Arslan Zulfiqar}, {and} \bibinfo{person}{Stephen~W. Keckler}.} \bibinfo{year}{2016}\natexlab{}.
\newblock \showarticletitle{vDNN: Virtualized deep neural networks for scalable, memory-efficient neural network design}. In \bibinfo{booktitle}{\emph{International Symposium on Microarchitecture (MICRO)}}.
\newblock


\bibitem[Rozi{\`{e}}re et~al\mbox{.}(2023)]%
        {DBLP:journals/corr/abs-2308-12950}
\bibfield{author}{\bibinfo{person}{Baptiste Rozi{\`{e}}re}, \bibinfo{person}{Jonas Gehring}, \bibinfo{person}{Fabian Gloeckle}, \bibinfo{person}{Sten Sootla}, \bibinfo{person}{Itai Gat}, \bibinfo{person}{Xiaoqing~Ellen Tan}, \bibinfo{person}{Yossi Adi}, \bibinfo{person}{Jingyu Liu}, \bibinfo{person}{Tal Remez}, \bibinfo{person}{J{\'{e}}r{\'{e}}my Rapin}, \bibinfo{person}{Artyom Kozhevnikov}, \bibinfo{person}{Ivan Evtimov}, \bibinfo{person}{Joanna Bitton}, \bibinfo{person}{Manish Bhatt}, \bibinfo{person}{Cristian Canton{-}Ferrer}, \bibinfo{person}{Aaron Grattafiori}, \bibinfo{person}{Wenhan Xiong}, \bibinfo{person}{Alexandre D{\'{e}}fossez}, \bibinfo{person}{Jade Copet}, \bibinfo{person}{Faisal Azhar}, \bibinfo{person}{Hugo Touvron}, \bibinfo{person}{Louis Martin}, \bibinfo{person}{Nicolas Usunier}, \bibinfo{person}{Thomas Scialom}, {and} \bibinfo{person}{Gabriel Synnaeve}.} \bibinfo{year}{2023}\natexlab{}.
\newblock \showarticletitle{Code Llama: Open Foundation Models for Code}.
\newblock \bibinfo{journal}{\emph{arxiv preprint, 2308.12950}} (\bibinfo{year}{2023}).
\newblock


\bibitem[Ruan and Jin(2022)]%
        {DBLP:journals/aiopen/RuanJ22}
\bibfield{author}{\bibinfo{person}{Ludan Ruan} {and} \bibinfo{person}{Qin Jin}.} \bibinfo{year}{2022}\natexlab{}.
\newblock \showarticletitle{Survey: Transformer based video-language pre-training}.
\newblock \bibinfo{journal}{\emph{{AI} Open}}  \bibinfo{volume}{3} (\bibinfo{year}{2022}).
\newblock


\bibitem[Rui et~al\mbox{.}(2020)]%
        {gpudatabaseoverlap1}
\bibfield{author}{\bibinfo{person}{Ran Rui}, \bibinfo{person}{Hao Li}, {and} \bibinfo{person}{Yi-Cheng Tu}.} \bibinfo{year}{2020}\natexlab{}.
\newblock \showarticletitle{Efficient join algorithms for large database tables in a multi-GPU environment}.
\newblock \bibinfo{journal}{\emph{Proceedings of the VLDB Endowment}}  \bibinfo{volume}{14} (\bibinfo{year}{2020}).
\newblock


\bibitem[Schleich et~al\mbox{.}(2023)]%
        {optimizetensorprogram}
\bibfield{author}{\bibinfo{person}{Maximilian Schleich}, \bibinfo{person}{Amir Shaikhha}, {and} \bibinfo{person}{Dan Suciu}.} \bibinfo{year}{2023}\natexlab{}.
\newblock \showarticletitle{Optimizing Tensor Programs on Flexible Storage}. In \bibinfo{booktitle}{\emph{Proceedings of the ACM on Management of Data (SIGMOD)}}.
\newblock


\bibitem[Sekiyama et~al\mbox{.}(2018)]%
        {sekiyama2018profileguidedmemoryoptimizationdeep}
\bibfield{author}{\bibinfo{person}{Taro Sekiyama}, \bibinfo{person}{Takashi Imamichi}, \bibinfo{person}{Haruki Imai}, {and} \bibinfo{person}{Rudy Raymond}.} \bibinfo{year}{2018}\natexlab{}.
\newblock \showarticletitle{Profile-guided memory optimization for deep neural networks}.
\newblock \bibinfo{journal}{\emph{arxiv preprint, 1804.10001}} (\bibinfo{year}{2018}).
\newblock


\bibitem[Shoeybi et~al\mbox{.}(2019)]%
        {DBLP:journals/corr/abs-1909-08053}
\bibfield{author}{\bibinfo{person}{Mohammad Shoeybi}, \bibinfo{person}{Mostofa Patwary}, \bibinfo{person}{Raul Puri}, \bibinfo{person}{Patrick LeGresley}, \bibinfo{person}{Jared Casper}, {and} \bibinfo{person}{Bryan Catanzaro}.} \bibinfo{year}{2019}\natexlab{}.
\newblock \showarticletitle{Megatron-LM: Training Multi-Billion Parameter Language Models Using Model Parallelism}.
\newblock \bibinfo{journal}{\emph{arxiv preprint, 1909.08053}} (\bibinfo{year}{2019}).
\newblock


\bibitem[Steiner et~al\mbox{.}(2022)]%
        {steiner2022ollaoptimizinglifetimelocation}
\bibfield{author}{\bibinfo{person}{Benoit Steiner}, \bibinfo{person}{Mostafa Elhoushi}, \bibinfo{person}{Jacob Kahn}, {and} \bibinfo{person}{James Hegarty}.} \bibinfo{year}{2022}\natexlab{}.
\newblock \showarticletitle{OLLA: Optimizing the Lifetime and Location of Arrays to Reduce the Memory Usage of Neural Networks}.
\newblock \bibinfo{journal}{\emph{arxiv preprint, 2210.12924}} (\bibinfo{year}{2022}).
\newblock


\bibitem[Taori et~al\mbox{.}(2023)]%
        {taori2023alpaca}
\bibfield{author}{\bibinfo{person}{Rohan Taori}, \bibinfo{person}{Ishaan Gulrajani}, \bibinfo{person}{Tianyi Zhang}, \bibinfo{person}{Yann Dubois}, \bibinfo{person}{Xuechen Li}, \bibinfo{person}{Carlos Guestrin}, \bibinfo{person}{Percy Liang}, {and} \bibinfo{person}{Tatsunori~B Hashimoto}.} \bibinfo{year}{2023}\natexlab{}.
\newblock \showarticletitle{Alpaca: A strong, replicable instruction-following model}.
\newblock \bibinfo{journal}{\emph{Stanford Center for Research on Foundation Models. https://crfm. stanford. edu/2023/03/13/alpaca. html}}  \bibinfo{volume}{3} (\bibinfo{year}{2023}).
\newblock


\bibitem[Together.ai(2023)]%
        {llama32k}
\bibfield{author}{\bibinfo{person}{Together.ai}.} \bibinfo{year}{2023}\natexlab{}.
\newblock \bibinfo{title}{LLaMA-2-7B-32K}.
\newblock
\urldef\tempurl%
\url{https://huggingface.co/togethercomputer/LLaMA-2-7B-32K}
\showURL{%
\tempurl}


\bibitem[Touvron et~al\mbox{.}(2023)]%
        {DBLP:journals/corr/abs-2307-09288}
\bibfield{author}{\bibinfo{person}{Hugo Touvron}, \bibinfo{person}{Louis Martin}, \bibinfo{person}{Kevin Stone}, \bibinfo{person}{Peter Albert}, \bibinfo{person}{Amjad Almahairi}, \bibinfo{person}{Yasmine Babaei}, \bibinfo{person}{Nikolay Bashlykov}, \bibinfo{person}{Soumya Batra}, \bibinfo{person}{Prajjwal Bhargava}, \bibinfo{person}{Shruti Bhosale}, \bibinfo{person}{Dan Bikel}, \bibinfo{person}{Lukas Blecher}, \bibinfo{person}{Cristian Canton{-}Ferrer}, \bibinfo{person}{Moya Chen}, \bibinfo{person}{Guillem Cucurull}, \bibinfo{person}{David Esiobu}, \bibinfo{person}{Jude Fernandes}, \bibinfo{person}{Jeremy Fu}, \bibinfo{person}{Wenyin Fu}, \bibinfo{person}{Brian Fuller}, \bibinfo{person}{Cynthia Gao}, \bibinfo{person}{Vedanuj Goswami}, \bibinfo{person}{Naman Goyal}, \bibinfo{person}{Anthony Hartshorn}, \bibinfo{person}{Saghar Hosseini}, \bibinfo{person}{Rui Hou}, \bibinfo{person}{Hakan Inan}, \bibinfo{person}{Marcin Kardas}, \bibinfo{person}{Viktor Kerkez}, \bibinfo{person}{Madian Khabsa},
  \bibinfo{person}{Isabel Kloumann}, \bibinfo{person}{Artem Korenev}, \bibinfo{person}{Punit~Singh Koura}, \bibinfo{person}{Marie{-}Anne Lachaux}, \bibinfo{person}{Thibaut Lavril}, \bibinfo{person}{Jenya Lee}, \bibinfo{person}{Diana Liskovich}, \bibinfo{person}{Yinghai Lu}, \bibinfo{person}{Yuning Mao}, \bibinfo{person}{Xavier Martinet}, \bibinfo{person}{Todor Mihaylov}, \bibinfo{person}{Pushkar Mishra}, \bibinfo{person}{Igor Molybog}, \bibinfo{person}{Yixin Nie}, \bibinfo{person}{Andrew Poulton}, \bibinfo{person}{Jeremy Reizenstein}, \bibinfo{person}{Rashi Rungta}, \bibinfo{person}{Kalyan Saladi}, \bibinfo{person}{Alan Schelten}, \bibinfo{person}{Ruan Silva}, \bibinfo{person}{Eric~Michael Smith}, \bibinfo{person}{Ranjan Subramanian}, \bibinfo{person}{Xiaoqing~Ellen Tan}, \bibinfo{person}{Binh Tang}, \bibinfo{person}{Ross Taylor}, \bibinfo{person}{Adina Williams}, \bibinfo{person}{Jian~Xiang Kuan}, \bibinfo{person}{Puxin Xu}, \bibinfo{person}{Zheng Yan}, \bibinfo{person}{Iliyan Zarov}, \bibinfo{person}{Yuchen
  Zhang}, \bibinfo{person}{Angela Fan}, \bibinfo{person}{Melanie Kambadur}, \bibinfo{person}{Sharan Narang}, \bibinfo{person}{Aur{\'{e}}lien Rodriguez}, \bibinfo{person}{Robert Stojnic}, \bibinfo{person}{Sergey Edunov}, {and} \bibinfo{person}{Thomas Scialom}.} \bibinfo{year}{2023}\natexlab{}.
\newblock \showarticletitle{Llama 2: Open Foundation and Fine-Tuned Chat Models}.
\newblock \bibinfo{journal}{\emph{arxiv preprint, 2307.09288}} (\bibinfo{year}{2023}).
\newblock


\bibitem[Um et~al\mbox{.}(2023)]%
        {fastflow}
\bibfield{author}{\bibinfo{person}{Taegeon Um}, \bibinfo{person}{Byungsoo Oh}, \bibinfo{person}{Byeongchan Seo}, \bibinfo{person}{Minhyeok Kweun}, \bibinfo{person}{Goeun Kim}, {and} \bibinfo{person}{Woo{-}Yeon Lee}.} \bibinfo{year}{2023}\natexlab{}.
\newblock \showarticletitle{FastFlow: Accelerating Deep Learning Model Training with Smart Offloading of Input Data Pipeline}.
\newblock \bibinfo{journal}{\emph{Proceedings of the VLDB Endowment}}  \bibinfo{volume}{16} (\bibinfo{year}{2023}).
\newblock


\bibitem[Wang et~al\mbox{.}(2023)]%
        {dse2}
\bibfield{author}{\bibinfo{person}{Guozheng Wang}, \bibinfo{person}{Yongmei Lei}, \bibinfo{person}{Zeyu Zhang}, {and} \bibinfo{person}{Cunlu Peng}.} \bibinfo{year}{2023}\natexlab{}.
\newblock \showarticletitle{A Communication Efficient ADMM-based Distributed Algorithm Using Two-Dimensional Torus Grouping AllReduce}.
\newblock \bibinfo{journal}{\emph{Data Sci. Eng.}}  \bibinfo{volume}{8} (\bibinfo{year}{2023}).
\newblock


\bibitem[Wang et~al\mbox{.}(2014)]%
        {gpudatabaseoverlap2}
\bibfield{author}{\bibinfo{person}{Kaibo Wang}, \bibinfo{person}{Kai Zhang}, \bibinfo{person}{Yuan Yuan}, \bibinfo{person}{Siyuan Ma}, \bibinfo{person}{Rubao Lee}, \bibinfo{person}{Xiaoning Ding}, {and} \bibinfo{person}{Xiaodong Zhang}.} \bibinfo{year}{2014}\natexlab{}.
\newblock \showarticletitle{Concurrent analytical query processing with GPUs}.
\newblock \bibinfo{journal}{\emph{Proceedings of the VLDB Endowment}}  \bibinfo{volume}{7} (\bibinfo{year}{2014}).
\newblock


\bibitem[Wang et~al\mbox{.}(2020)]%
        {wang2020linformerselfattentionlinearcomplexity}
\bibfield{author}{\bibinfo{person}{Sinong Wang}, \bibinfo{person}{Belinda~Z. Li}, \bibinfo{person}{Madian Khabsa}, \bibinfo{person}{Han Fang}, {and} \bibinfo{person}{Hao Ma}.} \bibinfo{year}{2020}\natexlab{}.
\newblock \showarticletitle{Linformer: Self-Attention with Linear Complexity}.
\newblock \bibinfo{journal}{\emph{arxiv preprint, 2006.04768}} (\bibinfo{year}{2020}).
\newblock


\bibitem[Wang et~al\mbox{.}(2024)]%
        {galvatron_2}
\bibfield{author}{\bibinfo{person}{Yujie Wang}, \bibinfo{person}{Youhe Jiang}, \bibinfo{person}{Xupeng Miao}, \bibinfo{person}{Fangcheng Fu}, \bibinfo{person}{Shenhan Zhu}, \bibinfo{person}{Xiaonan Nie}, \bibinfo{person}{Yaofeng Tu}, {and} \bibinfo{person}{Bin Cui}.} \bibinfo{year}{2024}\natexlab{}.
\newblock \showarticletitle{{ Improving Automatic Parallel Training via Balanced Memory Workload Optimization }}.
\newblock \bibinfo{journal}{\emph{IEEE Transactions on Knowledge and Data Engineering (TKDE)}}  \bibinfo{volume}{36} (\bibinfo{year}{2024}).
\newblock


\bibitem[Xiong et~al\mbox{.}(2021)]%
        {xiong2021nystromformernystrombasedalgorithmapproximating}
\bibfield{author}{\bibinfo{person}{Yunyang Xiong}, \bibinfo{person}{Zhanpeng Zeng}, \bibinfo{person}{Rudrasis Chakraborty}, \bibinfo{person}{Mingxing Tan}, \bibinfo{person}{Glenn Fung}, \bibinfo{person}{Yin Li}, {and} \bibinfo{person}{Vikas Singh}.} \bibinfo{year}{2021}\natexlab{}.
\newblock \showarticletitle{Nystr{\"{o}}mformer: {A} Nystr{\"{o}}m-based Algorithm for Approximating Self-Attention}. In \bibinfo{booktitle}{\emph{The AAAI Conference on Artificial Intelligence (AAAI)}}.
\newblock


\bibitem[Yu et~al\mbox{.}(2023)]%
        {jcst2}
\bibfield{author}{\bibinfo{person}{Feng Yu}, \bibinfo{person}{Jiacheng Zhao}, \bibinfo{person}{Hui{-}Min Cui}, \bibinfo{person}{Xiaobing Feng}, {and} \bibinfo{person}{Jingling Xue}.} \bibinfo{year}{2023}\natexlab{}.
\newblock \showarticletitle{VTensor: Using Virtual Tensors to Build a Layout-Oblivious {AI} Programming Framework}.
\newblock \bibinfo{journal}{\emph{J. Comput. Sci. Technol.}}  \bibinfo{volume}{38} (\bibinfo{year}{2023}).
\newblock


\bibitem[Yuan et~al\mbox{.}(2021)]%
        {tensorrelational}
\bibfield{author}{\bibinfo{person}{Binhang Yuan}, \bibinfo{person}{Dimitrije Jankov}, \bibinfo{person}{Jia Zou}, \bibinfo{person}{Yuxin Tang}, \bibinfo{person}{Daniel Bourgeois}, {and} \bibinfo{person}{Chris Jermaine}.} \bibinfo{year}{2021}\natexlab{}.
\newblock \showarticletitle{Tensor Relational Algebra for Distributed Machine Learning System Design}.
\newblock \bibinfo{journal}{\emph{Proceedings of the VLDB Endowment}}  \bibinfo{volume}{14} (\bibinfo{year}{2021}).
\newblock


\bibitem[Zaharia et~al\mbox{.}(2012)]%
        {rdd}
\bibfield{author}{\bibinfo{person}{Matei Zaharia}, \bibinfo{person}{Mosharaf Chowdhury}, \bibinfo{person}{Tathagata Das}, \bibinfo{person}{Ankur Dave}, \bibinfo{person}{Justin Ma}, \bibinfo{person}{Murphy McCauley}, \bibinfo{person}{Michael~J. Franklin}, \bibinfo{person}{Scott Shenker}, {and} \bibinfo{person}{Ion Stoica}.} \bibinfo{year}{2012}\natexlab{}.
\newblock \showarticletitle{Resilient distributed datasets: a fault-tolerant abstraction for in-memory cluster computing}. In \bibinfo{booktitle}{\emph{Proceedings of the USENIX Conference on Networked Systems Design and Implementation (NSDI)}}.
\newblock


\bibitem[Zaharia et~al\mbox{.}(2016)]%
        {spark}
\bibfield{author}{\bibinfo{person}{Matei Zaharia}, \bibinfo{person}{Reynold~S. Xin}, \bibinfo{person}{Patrick Wendell}, \bibinfo{person}{Tathagata Das}, \bibinfo{person}{Michael Armbrust}, \bibinfo{person}{Ankur Dave}, \bibinfo{person}{Xiangrui Meng}, \bibinfo{person}{Josh Rosen}, \bibinfo{person}{Shivaram Venkataraman}, \bibinfo{person}{Michael~J. Franklin}, \bibinfo{person}{Ali Ghodsi}, \bibinfo{person}{Joseph Gonzalez}, \bibinfo{person}{Scott Shenker}, {and} \bibinfo{person}{Ion Stoica}.} \bibinfo{year}{2016}\natexlab{}.
\newblock \showarticletitle{Apache Spark: a unified engine for big data processing}.
\newblock \bibinfo{journal}{\emph{Commun. {ACM}}}  \bibinfo{volume}{59} (\bibinfo{year}{2016}).
\newblock


\bibitem[Zaheer et~al\mbox{.}(2020)]%
        {zaheer2021bigbirdtransformerslonger}
\bibfield{author}{\bibinfo{person}{Manzil Zaheer}, \bibinfo{person}{Guru Guruganesh}, \bibinfo{person}{Kumar~Avinava Dubey}, \bibinfo{person}{Joshua Ainslie}, \bibinfo{person}{Chris Alberti}, \bibinfo{person}{Santiago Onta{\~{n}}{\'{o}}n}, \bibinfo{person}{Philip Pham}, \bibinfo{person}{Anirudh Ravula}, \bibinfo{person}{Qifan Wang}, \bibinfo{person}{Li Yang}, {and} \bibinfo{person}{Amr Ahmed}.} \bibinfo{year}{2020}\natexlab{}.
\newblock \showarticletitle{Big Bird: Transformers for Longer Sequences}. In \bibinfo{booktitle}{\emph{Advances in Neural Information Processing Systems (NeurIPS)}}.
\newblock


\bibitem[Zhang et~al\mbox{.}(2024a)]%
        {cafe}
\bibfield{author}{\bibinfo{person}{Hailin Zhang}, \bibinfo{person}{Zirui Liu}, \bibinfo{person}{Boxuan Chen}, \bibinfo{person}{Yikai Zhao}, \bibinfo{person}{Tong Zhao}, \bibinfo{person}{Tong Yang}, {and} \bibinfo{person}{Bin Cui}.} \bibinfo{year}{2024}\natexlab{a}.
\newblock \showarticletitle{{CAFE:} Towards Compact, Adaptive, and Fast Embedding for Large-scale Recommendation Models}.
\newblock \bibinfo{journal}{\emph{Proceedings of the ACM on Management of Data (SIGMOD)}}.
\newblock


\bibitem[Zhang et~al\mbox{.}(2024b)]%
        {scis2}
\bibfield{author}{\bibinfo{person}{Huangzhao Zhang}, \bibinfo{person}{Kechi Zhang}, \bibinfo{person}{Zhuo Li}, \bibinfo{person}{Jia Li}, \bibinfo{person}{Jia Li}, \bibinfo{person}{Yongmin Li}, \bibinfo{person}{Yunfei Zhao}, \bibinfo{person}{Yuqi Zhu}, \bibinfo{person}{Fang Liu}, \bibinfo{person}{Ge Li}, {and} \bibinfo{person}{Zhi Jin}.} \bibinfo{year}{2024}\natexlab{b}.
\newblock \showarticletitle{Deep learning for code generation: a survey}.
\newblock \bibinfo{journal}{\emph{Science China Information Sciences}}  \bibinfo{volume}{67} (\bibinfo{year}{2024}).
\newblock


\bibitem[Zhang et~al\mbox{.}(2023b)]%
        {embedding_ea}
\bibfield{author}{\bibinfo{person}{Hailin Zhang}, \bibinfo{person}{Penghao Zhao}, \bibinfo{person}{Xupeng Miao}, \bibinfo{person}{Yingxia Shao}, \bibinfo{person}{Zirui Liu}, \bibinfo{person}{Tong Yang}, {and} \bibinfo{person}{Bin Cui}.} \bibinfo{year}{2023}\natexlab{b}.
\newblock \showarticletitle{Experimental Analysis of Large-scale Learnable Vector Storage Compression}.
\newblock \bibinfo{journal}{\emph{Proceedings of the VLDB Endowment}}  \bibinfo{volume}{17} (\bibinfo{year}{2023}).
\newblock


\bibitem[Zhang et~al\mbox{.}(2023a)]%
        {DBLP:conf/nips/ZhangMLY23}
\bibfield{author}{\bibinfo{person}{Jianhao Zhang}, \bibinfo{person}{Shihan Ma}, \bibinfo{person}{Peihong Liu}, {and} \bibinfo{person}{Jinhui Yuan}.} \bibinfo{year}{2023}\natexlab{a}.
\newblock \showarticletitle{Coop: Memory is not a Commodity}. In \bibinfo{booktitle}{\emph{Advances in Neural Information Processing Systems (NeurIPS)}}.
\newblock


\bibitem[Zhang et~al\mbox{.}(2022)]%
        {mics}
\bibfield{author}{\bibinfo{person}{Zhen Zhang}, \bibinfo{person}{Shuai Zheng}, \bibinfo{person}{Yida Wang}, \bibinfo{person}{Justin Chiu}, \bibinfo{person}{George Karypis}, \bibinfo{person}{Trishul Chilimbi}, \bibinfo{person}{Mu Li}, {and} \bibinfo{person}{Xin Jin}.} \bibinfo{year}{2022}\natexlab{}.
\newblock \showarticletitle{MiCS: Near-linear Scaling for Training Gigantic Model on Public Cloud}.
\newblock \bibinfo{journal}{\emph{Proceedings of the VLDB Endowment}}  \bibinfo{volume}{16} (\bibinfo{year}{2022}).
\newblock


\bibitem[Zhao et~al\mbox{.}(2024)]%
        {rag_survey}
\bibfield{author}{\bibinfo{person}{Penghao Zhao}, \bibinfo{person}{Hailin Zhang}, \bibinfo{person}{Qinhan Yu}, \bibinfo{person}{Zhengren Wang}, \bibinfo{person}{Yunteng Geng}, \bibinfo{person}{Fangcheng Fu}, \bibinfo{person}{Ling Yang}, \bibinfo{person}{Wentao Zhang}, \bibinfo{person}{Jie Jiang}, {and} \bibinfo{person}{Bin Cui}.} \bibinfo{year}{2024}\natexlab{}.
\newblock \showarticletitle{Retrieval-Augmented Generation for AI-Generated Content: A Survey}.
\newblock \bibinfo{journal}{\emph{arxiv preprint, 2402.19473}} (\bibinfo{year}{2024}).
\newblock


\bibitem[Zhao et~al\mbox{.}(2022)]%
        {tod}
\bibfield{author}{\bibinfo{person}{Yue Zhao}, \bibinfo{person}{George~H. Chen}, {and} \bibinfo{person}{Zhihao Jia}.} \bibinfo{year}{2022}\natexlab{}.
\newblock \showarticletitle{{TOD:} GPU-accelerated Outlier Detection via Tensor Operations}.
\newblock \bibinfo{journal}{\emph{Proceedings of the VLDB Endowment}}  \bibinfo{volume}{16} (\bibinfo{year}{2022}).
\newblock


\bibitem[Zhao et~al\mbox{.}(2023)]%
        {fsdp}
\bibfield{author}{\bibinfo{person}{Yanli Zhao}, \bibinfo{person}{Andrew Gu}, \bibinfo{person}{Rohan Varma}, \bibinfo{person}{Liang Luo}, \bibinfo{person}{Chien{-}Chin Huang}, \bibinfo{person}{Min Xu}, \bibinfo{person}{Less Wright}, \bibinfo{person}{Hamid Shojanazeri}, \bibinfo{person}{Myle Ott}, \bibinfo{person}{Sam Shleifer}, \bibinfo{person}{Alban Desmaison}, \bibinfo{person}{Can Balioglu}, \bibinfo{person}{Pritam Damania}, \bibinfo{person}{Bernard Nguyen}, \bibinfo{person}{Geeta Chauhan}, \bibinfo{person}{Yuchen Hao}, \bibinfo{person}{Ajit Mathews}, {and} \bibinfo{person}{Shen Li}.} \bibinfo{year}{2023}\natexlab{}.
\newblock \showarticletitle{PyTorch {FSDP:} Experiences on Scaling Fully Sharded Data Parallel}.
\newblock \bibinfo{journal}{\emph{Proceedings of the VLDB Endowment}}  \bibinfo{volume}{16} (\bibinfo{year}{2023}).
\newblock


\bibitem[Zheng et~al\mbox{.}(2024)]%
        {opensora}
\bibfield{author}{\bibinfo{person}{Zangwei Zheng}, \bibinfo{person}{Xiangyu Peng}, \bibinfo{person}{Tianji Yang}, \bibinfo{person}{Chenhui Shen}, \bibinfo{person}{Shenggui Li}, \bibinfo{person}{Hongxin Liu}, \bibinfo{person}{Yukun Zhou}, \bibinfo{person}{Tianyi Li}, {and} \bibinfo{person}{Yang You}.} \bibinfo{year}{2024}\natexlab{}.
\newblock \bibinfo{booktitle}{\emph{Open-Sora: Democratizing Efficient Video Production for All}}.
\newblock
\urldef\tempurl%
\url{https://github.com/hpcaitech/Open-Sora}
\showURL{%
\tempurl}


\bibitem[Zhou et~al\mbox{.}(2024)]%
        {dse1}
\bibfield{author}{\bibinfo{person}{Xuanhe Zhou}, \bibinfo{person}{Zhaoyan Sun}, {and} \bibinfo{person}{Guoliang Li}.} \bibinfo{year}{2024}\natexlab{}.
\newblock \showarticletitle{{DB-GPT:} Large Language Model Meets Database}.
\newblock \bibinfo{journal}{\emph{Data Sci. Eng.}}  \bibinfo{volume}{9} (\bibinfo{year}{2024}).
\newblock


\bibitem[Zhu et~al\mbox{.}(2024)]%
        {DBLP:journals/corr/abs-2304-04675}
\bibfield{author}{\bibinfo{person}{Wenhao Zhu}, \bibinfo{person}{Hongyi Liu}, \bibinfo{person}{Qingxiu Dong}, \bibinfo{person}{Jingjing Xu}, \bibinfo{person}{Shujian Huang}, \bibinfo{person}{Lingpeng Kong}, \bibinfo{person}{Jiajun Chen}, {and} \bibinfo{person}{Lei Li}.} \bibinfo{year}{2024}\natexlab{}.
\newblock \showarticletitle{Multilingual Machine Translation with Large Language Models: Empirical Results and Analysis}. In \bibinfo{booktitle}{\emph{Findings of the Association for Computational Linguistics (NAACL)}}.
\newblock


\bibitem[Zinkevich et~al\mbox{.}(2010)]%
        {DBLP:conf/nips/ZinkevichWSL10}
\bibfield{author}{\bibinfo{person}{Martin Zinkevich}, \bibinfo{person}{Markus Weimer}, \bibinfo{person}{Alexander~J. Smola}, {and} \bibinfo{person}{Lihong Li}.} \bibinfo{year}{2010}\natexlab{}.
\newblock \showarticletitle{Parallelized Stochastic Gradient Descent}. In \bibinfo{booktitle}{\emph{Advances in Neural Information Processing Systems (NeurIPS)}}.
\newblock


\end{thebibliography}

\end{document}